\documentclass{article}
\pdfoutput=1

\usepackage[final]{neurips_2022}

\usepackage[utf8]{inputenc} 
\usepackage[T1]{fontenc}    
\usepackage{hyperref}       
\usepackage{url}            
\usepackage{booktabs}       
\usepackage{amsfonts}       
\usepackage{nicefrac}       
\usepackage{microtype}      
\usepackage{xcolor}         

\usepackage{amsmath}
\usepackage{amssymb}
\usepackage{mathtools}
\usepackage{amsthm}

\usepackage{xspace}
\usepackage{enumerate, paralist}
\usepackage{natbib}
\usepackage{comment}
\usepackage{multirow}
\usepackage{xspace}
\usepackage{rotating}
\usepackage{graphicx}
\usepackage{booktabs} 
\usepackage{tabularx}
\usepackage{enumitem}
\usepackage{algorithm, algpseudocode} 
\usepackage{bbm}
\usepackage{subfigure,xcolor}

\newcommand{\bbx}{\mathbf{X}}

\newcommand{\bx}{\mathbf{x}}

\newcommand{\by}{\mathbf{y}}
\newcommand{\bbr}{\mathbb{R}}
\newcommand{\cald}{\mathcal{D}}

\newcommand{\calh}{\mathcal{H}}
\newcommand{\calt}{\mathcal{T}}
\newcommand{\calb}{\mathcal{B}}

\newcommand{\bbe}{\mathbb{E}}
\newcommand{\call}{\mathcal{L}}
\newcommand{\btheta}{\boldsymbol \theta}
\newcommand{\bbeta}{\boldsymbol \beta}
\newcommand{\tri}{\triangledown}
\newcommand{\calx}{\mathcal{X}}
\newcommand{\caly}{\mathcal{Y}}

\newcommand{\bbp}{\mathbb{P}}
\newcommand{\hbtheta}{\widehat{\btheta}}
\newcommand{\calo}{\mathcal{O}}
\newcommand{\bw}{\mathbf{W}}

\newcommand{\hi}{\widehat{i}}
\newcommand{\ci}{i^\circ}
\newcommand{\br}{\mathbf{R}}

\usepackage{tikz}

\newcommand{\sysn}{\text{I-NeurAL}\xspace}
\newcommand{\para}[1]{\noindent \textbf{#1}\xspace}

\newtheorem{theorem}{Theorem}[section]

\newtheorem{definition}{Definition}[section]
\newtheorem{lemma}{Lemma}[section]

\theoremstyle{definition} 
\newtheorem{remark}{Remark}[section]
\numberwithin{equation}{section}
\newtheorem{assumption}{Assumption}[section]

\usepackage{color-edits}
\usepackage{authblk}
\usepackage{lipsum}
\newcommand\blfootnote[1]{%
  \begingroup
  \renewcommand\thefootnote{}\footnote{#1}%
  \addtocounter{footnote}{-1}%
  \endgroup
}

\addauthor{ab}{blue}

\title{ Improved Algorithms for  Neural Active Learning}
\author[ ]{Yikun Ban$^*$\hspace{-0.5ex}}
\author[ ]{Yuheng Zhang$^*$\hspace{-0.5ex}}
\author[ ]{Hanghang Tong\hspace{-0.5ex}}
\author[ ]{Arindam Banerjee\hspace{-0.5ex}}
\author[ ]{Jingrui He\hspace{-0.5ex}}
\affil[ ]{University of Illinois Urbana-Champaign}
\affil[ ]{ \{yikunb2, yuhengz2, htong, arindamb, jingrui\}@illinois.edu }

\begin{document}

\maketitle
\vspace{-1cm}
\blfootnote{$*$ Both authors contribute equally.}
\begin{abstract}
 We improve the theoretical and empirical performance of neural-network(NN)-based active learning algorithms for the non-parametric streaming setting. In particular, we introduce two regret metrics by minimizing the population loss that are more suitable in active learning than the one used in state-of-the-art (SOTA) related work.  Then, the proposed algorithm leverages the powerful representation of NNs for both exploitation and  exploration, has the query decision-maker tailored for $k$-class classification problems with the performance guarantee, utilizes the full feedback, and updates parameters in a more practical and efficient manner. These careful designs lead to an instance-dependent regret upper bound, roughly improving by a multiplicative factor $O(\log T)$ and removing the curse of input dimensionality. Furthermore, we show that the algorithm can achieve the same performance as the Bayes-optimal classifier in the long run under the hard-margin setting in classification problems. In the end, we use extensive experiments to evaluate the proposed algorithm and SOTA baselines, to show the improved empirical performance.
 

\end{abstract}

\section{Introduction}

The Neural Network (NN) is one of the indispensable paradigms in machine learning and is widely used in multifarious supervised-learning tasks \cite{goodfellow2016deep}. As more and more complicated NNs are developed, the requirement of the training procedure on the labeled data grows, incurring significant cost of label annotation. Active learning investigates effective techniques on a much smaller labeled data set while attaining the comparable generalization performance to passive learning \cite{cohn1996active}.
In this paper, we focus on the classification problem in the streaming setting of active learning with NN models. At every round, the learner receives an instance and is compelled to decide on-the-fly whether or not to observe the label associated with this instance. This problem seeks to maximize the generalization capability of learned NNs in a sequence of rounds, such that the model has robust performance on the unseen data from the same distribution \cite{ren2021survey}.

In active learning, given access to the i.i.d. generated instances from a distribution $\cald$, suppose there exist a class of functions $\mathcal{F}$ that formulate the mapping from instances to theirs labels. In the parametric setting, i.e., $\mathcal{F}$ has finite VC-dimension \cite {hanneke2014theory}, existing works \cite{hanneke2007bound, beygelzimer2009importance, balcan2009agnostic} have shown that the active learning algorithms can achieve the convergence rate of $\widetilde{\calo}(1/\sqrt{N})$ to the best population loss in $\mathcal{F}$, where $N$ is the number of label queries. In the non-parametric setting, recent works \cite{locatelli2017adaptivity, minsker2012plug} provide the similar convergence results while suffering from the curse of input dimensionality. 
Unfortunately, most of NN-based approaches to active learning do not come with the performance guarantee, despite having powerful empirical results. 

The first performance guarantee for neural active learning has been established in a recent work by \cite{wang2021neural}, and the analysis is for over-parameterized neural networks with the assistance of Neural Tangent Kernel (NTK).
We carefully investigate the limitations of \cite{wang2021neural}, which turn into the main motivations of our paper. 
First, \cite{wang2021neural} transforms the classification problem into a multi-armed bandit problem \cite{zhou2020neural}, to minimize a pseudo regret metric. 
Yet, on the grounds that they seek to minimize the \textit{conditional} population loss on a sequence of given data, it is dubious that the pseudo regret used in \cite{wang2021neural} can explicitly measure the generalization capability of given algorithms (see Remark \ref{remark:r}). 
Second, the training process for NN models is not efficient, as~\cite{wang2021neural} uses vanilla gradient descent and starts from randomly initialized parameters in every round. Third, although \cite{wang2021neural} removes the curse of input dimensionality $d$, the performance guarantee strongly suffers from another introduced term, the effective dimensionality $\widetilde{d}$, which can be thought of as the non-linear dimensionalities of Hilbert space spanned by NTK. In the worse case, the magnitude of $\widetilde{d}$ can be an unacceptably large number and thus the performance guarantee collapses. 

\subsection{Main contributions}

In this paper, we propose a novel algorithm, \sysn (\textbf{I}mproved Algorithms for \textbf{Neur}al \textbf{A}ctive \textbf{L}earning),  to tackle the above limitations. Our contributions can be summarized as follows:
(1) We consider the $k$-class classification problem, and we introduce two new regret metrics to minimize the population loss, which can directly reflect the generalization capability of NN-based algorithms.
(2) \sysn has a neural exploration strategy with a novel component to decide whether or not to query the label, coming with the performance guarantee. 
 \sysn exploits the full feedback in active learning which is a subtle but effective idea. (3) \sysn is designed to support mini-batch Stochastic Gradient Descent (SGD). In particular, at every round, \sysn does mini-batch SGD starting with the parameters of the last round, i.e., with warm start, which is more efficient and practical compared to \cite{wang2021neural}.
(4) Without any noise assumption on the data distribution, we provide an instance-dependent performance guarantee of \sysn for over-parameterized neural networks. Compared to \cite{wang2021neural}, we remove the curse of both the input dimensionality $d$ and the effective dimensionality $\widetilde{d}$; Moreover, we roughly improve the regret by a multiplicative factor $\log (T)$, where $T$ is the number of rounds.
(5) under a hard-margin assumption on the data distribution,  we provide that NN models can achieve the same generalization capability as Bayes-optimal classifier after  $\calo(\log T)$ number of label queries;
(6) we conduct extensive experiments on real-world data sets to demonstrate the improved performance of \sysn over state-of-the-art baselines including the closest work \cite{wang2021neural} which has not provided empirical validation of their proposed algorithms.

\subsection{Related Work} \label{sec:relate}

Active learning has been extensively studied and applied to many essential applications \cite{settles2009active}.
Bayesian active learning methods typically use a probabilistic regression model to estimate the improvement of each query \cite{kapoor2007active, roy2001toward}. In spite of effectiveness on the small or moderate data sets, the Bayesian-based approaches are difficult to scale to large-scale data sets because of the batch sampling \cite{sener2017active}. Another important class, 
margin algorithms or uncertainty sampling \cite{lewis1994sequential}, obtains considerate performance improvement over passive learning and is further developed by many practitioners \cite{culotta2005reducing, joshi2009multi,mussmann2018uncertainty,brinker2003incorporating}. Margin algorithms are flexible and can be adapted to both streaming and pool settings.
In the pool setting, a line of works utilize the neural networks in active learning to improve the empirical performance \cite{moon2020confidence,schroder2020survey, ash2019deep, citovsky2021batch, kim2021lada, tan2021diversity, wang2021deep, Zhang_Tong_Xia_Zhu_Chi_Ying_2022, ash2021gone}.
However, they do not provide performance guarantee for NN-based active learning algorithms.
From the theoretical perspective, \cite{zhang2018efficient, dasgupta2005analysis, awasthi2014power,balcan2007margin,zhang2020efficient}
provide the performance guarantee with the specific classes of functions and
\cite{hanneke2019surrogate,desalvo2021online} present the theoretical analysis of active learning algorithms with the surrogate loss functions for binary classification. However, their performance guarantee is restricted within hypothesis classes, i.e, the parametric setting.
In contrast, our goal is to derive an NN-based algorithm in the non-parametric setting that performs well both empirically and theoretically. 
Neural contextual bandits\cite{zhou2020neural, zhang2020neural, ban2021ee, ban2021multi, ban2022neural, yunzheneural} provide the principled method to balance between the exploitation and exploration \cite{ban2020generic, ban2021local}.  
\cite{wang2021neural} transforms active learning into neural contextual bandit problem and obtains a performance guarantee, of which limitations are discussed above.

As \cite{wang2021neural} is the closest related work to our paper, we emphasize the differences of our techniques from  \cite{wang2021neural} throughout the paper.  We  introduce the problem definition and proposed algorithms in Section \ref{sec:prob} and Section \ref{sec:alg} respectively. Then, we provide performance guarantees in Section \ref{sec:reg} and empirical results in Section \ref{sec:exp}, ending with the conclusion in Section \ref{sec:con}.

\section{Problem Definition} \label{sec:prob}

In this paper, we study the streaming setting of active learning in  the $k$-class classification problem.
Let $\calx$ denote the input space over $\bbr^d$, $\caly = \{1, 2, \dots, k\}$ represent the label space, and $\cald$ be some unknown distribution over $\calx \times \caly$. 
At round $t \in [T] = \{1, 2, \dots, T\}$, an instance $\bx_t$ is drawn from the marginal distribution $\cald_{\calx}$ and accordingly $y_t$ is drawn from the conditional distribution  $\cald_{\caly|\bx_t}$. Here, $y_t$ can be thought of as the index of the class that $\bx_t$ belongs to.
Inspired by \cite{wang2021neural},  we first transform $\bx_t$ into $k$ context vectors representing the $k$ classes respectively:
$\bx_{t, 1} = (\bx_t^\top, \mathbf{0}^\top, \dots, \mathbf{0}^\top)^\top, \bx_{t, 2} = (\mathbf{0}^\top, \bx_t^\top, \dots, \mathbf{0}^\top)^\top, \dots, \bx_{t, k} = ( \mathbf{0}^\top,  \mathbf{0}^\top, \dots, \bx_t^\top)^\top$ and  $\bx_{t, i} \in \bbr^{dk}, \forall i \in [k]$. 
In accordance with context vectors, we construct the $k$ label vectors representing the $k$ possible prediction: $ \by_{t,1} =(1, 0, \dots, 0)^\top, \by_{t,2} = (0, 1, \dots, 0)^\top, \dots, \by_{t, k} = (0, 0, \dots, 1)^\top$ and $ \by_{t, i}  \in \bbr^k, \forall i \in [k]$. Thus, $\by_{t, y_t}$ is the ground-truth label vector for $\bx_{t}$.

Under the  non-parametric setting of active learning, we define an unknown function $h$ to formulate the conditional distribution $\cald_{\caly| \bx_t} $: $\calx^k \rightarrow [0, 1]$, such that
\begin{equation} \label{eq:h}
\forall i \in [k],  \bbp(\by_{t, y_t} = \by_{t, i} | \bx_{t}) = h(\bx_{t,i})~,
\end{equation}
which is subject to $\sum_{i = 1}^k h(\bx_{t,i}) = 1$. 
For simplicity, we consider the $k$-class classification problem with 0-1 loss. Given $\bx_t$, i.e., $\bx_{t,i}, i \in [k] $, let $\hi$ be the index of the class predicted by some hypothesis $f$ and thus  $\by_{t, \hi}$ is the prediction. Then, we have the following loss:
\begin{equation}  \label{eq:loss}
L(\by_{t, \hi}, \by_{t, y_t})  = \mathbbm{1}\{ \by_{t, \hi} \neq  \by_{t, y_t} \} \in \{0, 1\}~.
\end{equation}
where $\mathbbm{1}$ is the indicator function. 

Given the number of rounds $T$, at each round $t \in [T]$, the learner receives an instance $\bx_t$ drawn i.i.d. from $\cald_\calx$. Then, the learner needs to make a prediction $\by_{t, \hi}$,  and at the same time, decide on-the-fly whether or not to query the label $\by_{t, y_t}$ where $y_t$ is drawn i.i.d. from $\cald_{\caly|\bx_t}$.  
As the goal of active learning tasks is often to minimize the population loss \cite{ren2021survey}, we introduce the following two regret metrics.

\begin{definition}[Latest Population Regret]
Given the data distribution $\cald$, the number of rounds $T$, the Latest Population Regret is defined as
\begin{equation}
R_T = \underset{\bx_T \sim D_{\calx}}{\bbe} \left[\underset{ y_T \sim \cald_{\caly|\bx_T}}{\bbe} [  L(\by_{T, \hi}, \by_{T, y_T}) \mid \bx_T ] \right]  -  \underset{\bx_T \sim D_{\calx}}{\bbe} \left[  \underset{ y_T \sim \cald_{\caly|\bx_T}}{\bbe}[L(\by_{T, i^\ast}, \by_{T, y_T} )  \mid \bx_T] \right]
\end{equation}
where $\by_{T,i^\ast}$ is the prediction the Bayes-optimal classifier would make on instance $\bx_T$, i.e., 
$i^\ast = \arg \max_{i \in[k]} h(\bx_{T,i})$ for $\by_{T,i^\ast}$.
\end{definition}

\begin{definition}[Cumulative Population Regret]
Given the data distribution $\cald$, the number of rounds $T$, the Cumulative Population Regret is defined as:
\begin{equation} \label{eq:regret}
\mathbf{R}_T = \sum_{t =1 }^T\left(  \underset{\bx_t \sim D_{\calx}}{\bbe} \left[\underset{ y_t \sim \cald_{\caly|\bx_t}}{\bbe} [  L(\by_{t, \hi}, \by_{t, y_t}) \mid \bx_t ] \right]  -  \underset{\bx_t \sim D_{\calx}}{\bbe} \left[  \underset{ y_t \sim \cald_{\caly|\bx_t}}{\bbe}[L(\by_{t, i^\ast}, \by_{t, y_t} )  \mid \bx_t] \right] \right) 
\end{equation}
where $\by_{t, i^\ast}$ is the prediction the Bayes-optimal classifier would make on instance $\bx_t$, i.e., $i^\ast = \arg \max_{i \in[k]} h(\bx_{t,i})$ for $\by_{t, i^\ast}$.
\end{definition}

$R_T$ measures the performance at the last round $T$ only, and $\mathbf{R}_T$ measures the overall performance in $T$ rounds combined.
Therefore, the goal of this problem is to minimize $R_T$  or $\mathbf{R}_T$, or both.
At the same time, we also aim to minimize the following expected query cost:
\begin{equation} \label{eq:qureycost}
\mathbf{N}_T = \sum_{t =1 }^T \underset{ \bx_t \sim \cald_\calx}{\bbe} [\mathbf{I}_t  \mid \bx_t],
\end{equation}
where  $\mathbf{I}_t$ is the indicator of the query decision in round $t$ such that $\mathbf{I}_t = 1$ if $y_t$ is observed; $\mathbf{I}_t = 0$, otherwise.

\begin{remark} \label{remark:r}
Minimizing $R_T$ or $\mathbf{R}_T$ shows the generalization capability of the learned hypothesis on the distribution $\cald$. However, the problem defined in \cite{wang2021neural} is to minimize the cumulative \textit{conditional} population regret as follows:
\begin{equation}
\widetilde{\mathbf{R}}_T =  \sum_{t =1 }^T\left( \underset{ y_t \sim \cald_{\caly|\bx_t}}{\bbe}[L(\by_{t, \hi}, \by_{t, y_t}) | \bx_t ]    -  \underset{ y_t \sim \cald_{\caly|\bx_t}}{\bbe}[L(\by_{t, i^\ast}, \by_{t, y_t}) | \bx_t ] \right).
\end{equation}
As ${\bbe}_{y_t \sim \cald_{\caly|\bx_t}}[L(\by_{t, \hi}, \by_{t, y_t}) | \bx_t ]$ is the population loss conditioned on $\bx_t$, unfortunately, $\widetilde{\mathbf{R}}_T$ only measures the performance of the learned hypothesis on the collected data $\{\bx_t \}_{t=1}^T$, and $\widetilde{\mathbf{R}}_T$ cannot directly measure the accuracy of the hypothesis on unseen data instances.
Although $\widetilde{\mathbf{R}}_T$ follows the regret definition in multi-armed bandits \citep{zhou2020neural}, 
it is fair to say that $\widetilde{\mathbf{R}}_T $ may not be a good metric in active learning.
\end{remark}



\section{Proposed Algorithms} \label{sec:alg}

In this section, we elaborate on the proposed algorithm \sysn (Algorithm \ref{alg:main}).
In contrast to the directly comparable work \citep{wang2021neural}, \sysn has the following novel and advantageous aspects: (1) \sysn incorporates a neural-based exploration strategy (Line 6) inspired by recent advances in bandits \citep{ban2021ee} to solve the exploitation-exploration dilemma in the decision for whether or not to query labels; (2) \sysn includes a novel component (Line 11) to decide  whether or not to query labels in the $k$-class classification problem;
(3) \sysn infers and exploits the feedback of all the contexts (Lines 12-17), instead of only utilizing the feedback of the chosen context in \cite{wang2021neural}; 
(4) \sysn conducts mini-batch SGD based on the parameters of the last round (Algorithm \ref{alg:BGD}), which is more practical, as opposed to conducting vanilla gradient descent from the initialization at every round in \citep{wang2021neural}. Next, we will present the details of \sysn.

\textit{Exploitation Network $f_1$.}
Given $\bx_{t,i}, i \in [k]$, to learn the unknown function $h$ (Eq. (\ref{eq:h})), we use a fully-connected neural network $f_1$ with $L$-depth and $m$-width:
\begin{equation}
f_1 (\bx_{t,i}; \btheta^1) = \bw^1_L\sigma(\bw^1_{L-1}\sigma(\bw^1_{L-2} \dots \sigma(\bw^1_1 \bx_{t,i}))),
\end{equation}
where $\bw_1^1 \in \bbr^{m \times kd}, \bw_{l}^1  \in \bbr^{m \times m}$, for $2 \leq l \leq L-1$, $\bw_L^1 \in \bbr^{1 \times m}$, $\btheta^1 = [\text{vec}(\bw^1_1)^\top, \dots, \text{vec}(\bw^1_L)^\top]^\top \in \bbr^{p_1}$, and $\sigma$ is the ReLU activation function $\sigma(\bx) =  \max\{0, \bx\}$.
In round $t$, given $\bx_{t,i}, i \in [k]$,   $f_1 (\bx_{t,i}; \btheta^1_{t-1})$ is assigned to learn $h(\bx_{t,i})$. Based on the fact $h(\bx_{t,i}) = \underset{y_t \sim \cald_{\caly|\bx_t}}{\bbe}[1 - L(\by_{t, i}, \by_{t, y_t})]$, it is natural to regard  $1 - L(\by_{t, i}, \by_{t, y_t})$ as the label for training $f_1$.
Note that we take the basic fully-connected network as an example for the sake of analysis in over-parameterized networks and $f_1$ can be easily replaced with more complicated models depending on the tasks.

\textit{Exploration Network $f_2$.} In addition to the  network $f_1$, we assign another network $f_2$ to explore uncertain information contained in incoming instances. 
First, we carefully design the input of $f_2$ to incorporate the context vectors of the instance and the discrimination-ability of $f_1$, to learn the error between the Bayes-optimal probability $h(\bx_{t,i})$ and the prediction $f_1(\bx_{t,i}; \btheta^1)$.

\begin{definition}[Derivative-Context (DC) Embedding] \label{defembding}
Given the exploitation network $f_1(\cdot; \btheta_{t-1}^1)$ and an input context $\bx_{t,i}$, its DC embeding is defined as
\begin{equation}
\phi(\bx_{t,i}) = \left(\frac{ \text{vec}\left(\tri_{\bx_{t,i}}f_1( \bx_{t,i};\btheta_{t-1}^1)\right)^\top }{ \sqrt{2} \|\tri_{\bx_{t,i}}f_1( \bx_{t,i};\btheta_{t-1}^1) \|_2  },  \frac{ {\bx_{t,i}}^\top }{\sqrt{2}} \right) \in \bbr^{2dk},
\end{equation}
where $\tri_{\bx_{t,i}}f_1$ is the partial derivative of $f_1(\bx_{t,i}; \btheta_{t-1}^1)$ with respect to $\bx_{t,i}$. 
\end{definition}
 $\phi(\bx_{t,i})$ is normalized so that $\|\phi(\bx_{t,i}) \|_2 = 1$.  Note that the input for $f_2$ in \cite{ban2021ee} is the gradient with respect to $\theta_1$, denoted by $ \tri_{\theta_1}f_1( \bx_{t,i};\btheta_{t-1}^1) \in \bbr^{p_1}$. Its dimensionality is much larger than $\tri_{\bx_{t,i}}f_1( \bx_{t,i};\btheta_{t-1}^1)$ in Definition \ref{defembding},  may causing signiﬁcant computation cost. 

Given the input $\phi(\bx_{t,i})$, similarly, we choose the fully-connected network to build $f_2$: 
\begin{equation}
f_2 (\phi(\bx_{t,i}); \btheta^2) = \bw^2_L\sigma(\bw^2_{L-1}\sigma(\bw^2_{L-2} \dots \sigma(\bw^2_1 \phi( \bx_{t,i})))),
\end{equation}
where $\bw_1^2 \in \bbr^{m \times 2kd}, \bw_{l}^2  \in \bbr^{m \times m}$, for $2 \leq l \leq L-1$, $\bw_L^2 = \bbr^{1 \times m}$ and $\btheta^2 = [\text{vec}(\bw^2_1)^\top, \dots, \text{vec}(\bw^2_L)^\top]^\top \in \bbr^{p_2}$. In round $t$, given $\bx_{t,i}, \forall i \in [k]$, $f_2$ is to predict $h(\bx_{t,i}) - f_1(\bx_{t,i}; \btheta^1_{t-1})$ for exploration. 
Because $h(\bx_{t,i}) - f_1(\bx_{t,i}; \btheta^1_{t-1}) = \underset{y_t \sim \cald_{\caly|\bx_t}}{\bbe}[ 1 - L(\by_{t, i}, \by_{t, y_t}) -   f_1(\bx_{t,i}; \btheta^1_{t-1})  ]$, we regard $ 1 - L(\by_{t, i}, \by_{t, y_t}) -   f_1(\bx_{t,i}; \btheta^1_{t-1})$ as the label for training $f_2$.

To sum up, in round $t$, given $\bx_{t,i}, \forall i \in [k]$, the prediction $\hi$ ($\by_{t, \hi}$) is made based on the sum of exploitation and exploration scores, i.e., $ f_1(\bx_{t,i}; \btheta^1_{t-1}) +  f_2(\phi(\bx_{t,i}); \btheta^2_{t-1})$ (Lines 5-10).

\begin{algorithm}[t]
\begin{algorithmic}[1] 
\renewcommand{\algorithmicrequire}{\textbf{Input:}}
\renewcommand{\algorithmicensure}{\textbf{Output:}}
\caption{ \sysn}
\label{alg:main}
\Require $T$ (number of rounds) $f_1, f_2$ (neural networks),  $\eta_1, \eta_2$ (learning rate), $\gamma$ (exploration parameter), $b$ (batch size), $\delta$ (confidence level) 
\State Initialize $\btheta^1_0, \btheta^2_0$; $ \hbtheta^1_0 =\btheta^1_0 ; \hbtheta_0^2 = \btheta^2_0 $
\State $\mathcal{H}_0^1 = \emptyset; \mathcal{H}_0^2 = \emptyset$
\For{ $t = 1 , 2, \dots, T$}
\State Observe instance $\bx_t \in \bbr^d$ and build $\bx_{t, i}, \forall i \in [k]$
\For{each $i \in [k]$}
\State $f(\bx_{t,i}; \btheta_{t-1}) = \Big ( \underset{\text{Exploitation Score}}{ f_1(\bx_{t,i}; \btheta_{t-1}^1) }  + \underset{\text{Exploration Score}}{f_2( \phi(\bx_{t,i}); \btheta_{t-1}^2)} \Big) $
\EndFor
\State $\hi = \arg \max_{i \in [k]} f(\bx_{t,i}; \btheta_{t-1})$
\State $i^\circ = \arg \max_{i \in ([k] \setminus \{ \ \hi \ \})} f(\bx_{t,i}; \btheta_{t-1})   $
\State Predict $\by_{t,\hi}$
\State $\mathbf{I}_t = \mathbbm{1}\{|f(\bx_{t,\hi}; \btheta_{t-1}) -  f(\bx_{t,i^\circ}; \btheta_{t-1})| < 2 \gamma \bbeta_t \} \in \{0, 1\}$; $\bbeta_t = \sqrt{\frac{  2 c_1}{t}} + \left(\frac{c_2 3L}{\sqrt{2t}} \right) + \sqrt{ \frac{2 \log ( c_3 Tk)/\delta)}{t}} $ 
\If{$\mathbf{I}_t =1$}
\State Query $\bx_{t}$ and observe $y_t$ 
\For{$i \in [k]$}
\State $r_{t,i}^1 = 1 - L(\by_{t,i}, \by_{t, y_t})$ (defined in E.q. (\ref{eq:loss}))
\State $r_{t,i}^2 = r^1_{t,i} - f_1(\bx_{t,i}; \btheta^1_{t-1}) $
\EndFor
\Else
\For{$i \in [k]$}
\State $r_{t,i}^1 = 1 - L(\by_{t,i}, \by_{t, \hi })$
\State $r_{t,i}^2 = r^1_{t,i} - f_1(\bx_{t,i}; \btheta^1_{t-1}) $
\EndFor
\EndIf
\State  $\mathcal{H}_t^1 = \mathcal{H}_{t-1}^1 \cup \{(\bx_{t,i}, r^1_{t,i}), i \in [k]\}$
\State   $\mathcal{H}_t^2 = \mathcal{H}_{t-1}^2 \cup \{(\bx_{t,i}, r^2_{t,i}), i \in [k]\}$
\State  $\btheta_{t}^1, \btheta_{t}^2$  = Mini-Batch-SGD-Warm-Start ( $f_1$, $f_2$, $\mathcal{H}_t^1, \mathcal{H}_t^2$, $b$)
\EndFor
\State \textbf{Return} $(\btheta^1, \btheta^2)$ uniformly from  $((\btheta_{0}^1, \btheta_{0}^2), \dots, \btheta_{T-1}^1, \btheta_{T-1}^2  ) $
\end{algorithmic}
\end{algorithm}

\textit{Query Decision-maker (Line 11)}. A label query is made when \sysn is not confident enough to discriminate the Bayes-optimal class from other classes. $2\gamma\bbeta_t$ ($\bbeta_t$ is also defined in Lemma \ref{lemma:thetajbound}) can be thought of as a confidence interval for the distance between the optimal class and second optimal class, where $\gamma$ is the hyper-parameter to tune the sensitivity of the decision-maker in practice. Given any $\gamma \geq 1, \delta \in (0,1)$, with probability at least $1 - \delta$, based on our analysis (Lemma \ref{lemma:correctlabel}),
 $ \bbe_{(\bx_t, y_t) \sim \cald}[L(\by_{t, \hi}, \by_{t,y_t})] =\bbe_{(\bx_t, y_t) \sim \cald}[L(\by_{t, i^\ast}, \by_{t,y_t})] $
when $\mathbf{I}_t = 0$, i.e., \sysn suffers no regret. Thus, we use $ \by_{t,\hi}$ as the pseudo-label in this case and we have the following update rules.


\textit{Utilize Full Feedback (Lines 14-25)}.
Different from the bandit setting where the learner can only observe the reward of the selected context, we can infer the rewards of all contexts in active learning, as we know the specific class of the current instance. Thus, for each $\bx_{t,i}, i \in [k]$, $r^1_{t,i} = 1 - \call(\by_{t,i}, \by_{t, y_t})$ is regarded as the "reward" of $\bx_{t,i}$, predicted by $f_1$, and $r^2_{t,i} = r_{t,i}^1 - f_1(\bx_{t,i}; \btheta^1)$ is regarded as the "residual reward" of $\bx_{t,i}$, predicted by $f_2$. 
In summary, in round $t$, when $\mathbf{I}_t = 1$,  $\by_{t, y_t}$ is observed to update $r^1_{t,i}$ and $r^2_{t,i}$;  when $\mathbf{I}_t = 0$,  $\by_{t, \hi}$ is regard as the pseudo-label  to obtain $r^1_{t,i}$ and $r^2_{t,i}$, $\forall i \in [k]$.
Therefore, we have the training data $\mathcal{H}^1_t$ for $f_1$ and $\mathcal{H}^2_t$ for $f_2$.

\textit{Mini-Batch SGD with Warm-Start (Algorithm \ref{alg:BGD2})}. 
Unlike \cite{wang2021neural} that uses vanilla gradient descent from randomly initialized parameters in each round, causing unnecessarily expensive computation, we extend the training procedure to mini-batch SGD with warm start, i.e., we incrementally train the parameters $\btheta_t$ starting from the parameters of the last round $\btheta_{t-1}$ in each round $t$.

Algorithm \ref{alg:main} depicts the workflow of \sysn. Lines 1-2 initialize the parameters where each entry of $\bw_{l}$ is drawn from the normal distribution $\mathcal{N}(0, 2/m)$ and each entry of $\bw_L$
is drawn from $\mathcal{N}(0, 1/m)$ for both $f_1$ and $f_2$. $\mathcal{H}_0^1, \mathcal{H}_0^2$ store the historical data for $f_1$ and $f_2$ respectively. In each round $t$, Line 4 builds the $k$ contexts for the observed instance $\bx_t$, and Lines 5-7 calculate the exploitation-exploration score for each context. $\hi$ (Line 8) is the index of the optimal-predicted class and thus $\by_{t, \hi}$ is the prediction. $\ci$ (Line 9) is the index of the second optimal-predicted class, which is used to decide whether to make a query. Line 11 is our decision component. When $\mathbf{I_t} = 1$, it shows that we are not confident enough about our prediction, so that we make a query for $\bx_t$ and observe the rewards for each context (Lines 12-17).    When $\mathbf{I_t} = 0$, based on our analysis, with high confidence, the prediction $\by_{t, \hi}$ matches the one predicted by the Bayes-optimal classifier. Hence, we consider $\by_{t, \hi}$ as the label and observe the reward for all contexts (Lines 18-23). In the end, we update the networks $f_1$ and $f_2$, based on the collected data (Lines 24-26).

\begin{algorithm}[t]
\renewcommand{\algorithmicrequire}{\textbf{Input:}}
\renewcommand{\algorithmicensure}{\textbf{Output:}}
\caption{Mini-Batch-SGD-Warm-Start ( $f_1$, $f_2$, $ \mathcal{H}_t^1, \mathcal{H}_t^2$, $b$) }\label{alg:BGD}
\begin{algorithmic}[1] 
\State Define $\call_1[(\bx, r^1); \btheta^1] =  (r^1 - f_1(\bx; \btheta^1))^2/2$
\State Uniformly draw a set $\widehat{\mathcal{H}}^1_t  \subset \mathcal{H}_t^1, s.t.,  |\widehat{\mathcal{H}}^1_t| = b$
\State   $\hbtheta^1_t = \hbtheta^1_{t-1} -  \frac{\eta_1}{b} \underset{(\bx, r^1)\in  \widehat{\mathcal{H}}^1_t}{\sum}  \tri_{\btheta^1}  \call_1 [(\bx, r^1); \hbtheta_{t-1}^1]$
\State  Define $\call_2[(\phi(\bx), r^2); \btheta^2] =  (r^2 - f_2(\phi(\bx); \btheta^2))^2/2$
\State Uniformly draw a set $\widehat{\mathcal{H}}^2_t  \subset \mathcal{H}_t^2, s.t., |\widehat{\mathcal{H}}^2_t| = b$
\State $\hbtheta^2_t = \hbtheta^2_{t-1} -  \frac{\eta_2}{b} \underset{(\phi(\bx), r^1)\in  \widehat{\mathcal{H}}^2_t}{\sum}  \tri_{\btheta^2}  \call_2 [(\phi(\bx), r^2); \hbtheta_{t-1}^2]$
\State $\Omega_{t} = \Omega_{t-1} \cup \{(\hbtheta_{t}^1, \hbtheta_{t}^2) \}$
\State \textbf{Return} $(\btheta_{t}^1, \btheta_{t}^2)$ uniformly from  $\Omega_t$
\end{algorithmic}
\end{algorithm}
\section{Regret Analysis} \label{sec:reg}

In this section, we provide the regret analysis of \sysn in the over-parameterized neural networks. 
First, we need the standard normalization restricted to the input instances. 

\begin{assumption}
For any $t \in [T]$, $\|\bx_t \|_2 = 1$.
\end{assumption}

Inspired by \cite{cao2019generalization}, we define the following function class.
Given a constant $\nu > 0$, we define the following $\nu$-ball of $\btheta^2$ around the random initialization: $\calb(\btheta^2_0, \nu) = \{\widetilde{\btheta}^2: \|\widetilde{\btheta}^2 -  \btheta^2_0 \|_2 \leq \calo(\frac{\nu}{\sqrt{m}})\}$. 
Recall that $r^2_{t, \hi} = r^1_{t,  \hi} - f_1(\bx_{t, \hi}; \hbtheta_{t-1}^1)$. Let $\widehat{\btheta}^{1,\ast}_{t-1}$ represent the parameters trained on $ \calh_{t-1}^{1, \ast}$ using Algorithm \ref{alg:BGD} with the Bayes-optimal classifier, where $\calh_{t-1}^{1, \ast} = \{\bx_{\tau, i^\ast}, r^1_{\tau, i^\ast} \}_{\tau = 1}^{t-1}$ are the historical Bayes-optimal pairs. We define $r^{2, \ast}_{t, i^\ast} = r^1_{t,  i^\ast } - f_1(\bx_{t, i^\ast}; \hbtheta_{t-1}^{1, \ast})$.
Then, we provide the following regret bound that depends on the classification ability of exploration network class induced by $\calb(\btheta^2_0, \nu)$. 

\begin{theorem}\label{theo1}
Given the number of rounds $T$, for any $\delta \in (0, 1), \gamma >1,$ $\nu > 0,$ suppose $m \geq  \widetilde{ \Omega}(\text{poly} (T, k, L, \nu)), \eta_1 = \eta_2 = \Theta(\frac{ \kappa \nu}{\sqrt{T} m})$, $ \underset{\widetilde{\btheta}^2 \in \calb(\btheta^2_0, \nu) }{\inf}  \sum_{t=1}^T 
\left(      
f_2 ( \phi(\bx_{t, \hi}) ; \widetilde{\btheta}^2 ) -  r^2_{t, \hi}
\right)^2 \&   \underset{\widetilde{\btheta}^2 \in \calb(\btheta^2_0, \nu) }{\inf}  \sum_{t=1}^T 
\left(      
f_2 ( \phi(\bx_{t, i^\ast}) ; \widetilde{\btheta}^2 ) - r^{2, \ast}_{t, i^\ast}
\right)^2     \leq \mu$. Then,
with probability at least $1 -\delta$ over the initialization of $\btheta^1_0, \btheta^2_0$,
there exist a small enough constant $\kappa$, such that
 Algorithm \ref{alg:main} achieves the following regret bound:
\begin{equation}
\mathbf{R}_T \leq   \calo\left(2\sqrt{T} -1 \right) \left[\frac{6L\nu + 4 \sqrt{\mu}}{\sqrt{2}} + 2\sqrt{2 \log ( \calo(Tk)/\delta)} +  \calo (1)\right] 
\end{equation}
and at the same time $\mathbf{N}_T \leq \calo(T)$.
Suppose the Bayes-optimal classifier has zero classification errors, i.e., $L(\by_{t, i^\ast}, \by_{t, y_t} )  = 0, t \in [T] $. It holds that
\begin{equation}
 R_T \leq \calo \left(\frac{6L\nu + 4 \sqrt{\mu}}{\sqrt{2T}} \right) + 2 \sqrt{ \frac{2 \log ( \calo(Tk)/\delta) }{T}}.
\end{equation}
\end{theorem}

Theorem \ref{theo1} provides the regret bound of \sysn for $R_T$ and $\mathbf{R}_T$ respectively, $R_T \leq \calo( \frac{\sqrt{\log T }}{\sqrt{T}})$ and $\mathbf{R}_T \leq  \calo(\sqrt{T \log T})$.
As \cite{wang2021neural} only provides the regret bound for $\widetilde{R}_T$, to show the advantages of \sysn, we also provide the following lemma for fair comparison.

\begin{lemma} \label{lemma:1}
Given the number of rounds $T$, for any $\delta \in (0, 1),  \gamma >1$, $\nu > 0$, suppose $m \geq  \widetilde{\Omega}(\text{poly} (T, k, L, \nu)), \eta_1 = \eta_2 = \Theta(\frac{\kappa \nu}{\sqrt{T} m})$,  and $\mu$ satisfies the conditions in Theorem \ref{theo1}. Then, with probability at least $1 -\delta$ over the initialization of $\btheta^1_0, \btheta^2_0$, 
these exists a small enough constant $\kappa$, such that
Algorithm \ref{alg:main} can achieve the following regret bound:
\begin{equation} \label{eq:lemma1}
\widetilde{\mathbf{R}}_T \leq   \calo \left( \frac{6L\nu + 4 \sqrt{\mu}}{\sqrt{2}}  \right)  \sqrt{T} + 2\sqrt{2 T \log ( \calo(T)/\delta)}+  \calo(1) 
\end{equation}
and at the same time $\mathbf{N}_T \leq \calo(T)$.
\end{lemma}

\para{Comparison with \cite{wang2021neural}}. 
Lemma \ref{lemma:1} shows that \sysn can achieve the regret bound of same complexity for $\widetilde{\mathbf{R}}_T$ as $\br_T$. 
Under the same assumption in the over-parameterized neural networks, without any assumption on $\cald$, 
Theorem 1 in \cite{wang2021neural} (i.e., the lower-noise condition with exponent $\alpha = 0$, and $k=2$ is ignored in the binary classification) achieves the following regret bound:
$\widetilde{\mathbf{R}}_T \leq \calo(\log \det (I + \mathbf{H}) \sqrt{ T ( \log \det (I + \mathbf{H})  + S^2 )})$ where $\mathbf{H}$ is the NTK matrix \cite{ntk2018neural,arora2019exact} formed by received instances of all $T$ rounds, 
$S = \sqrt{\mathbf{h}^\top \mathbf{H}^{-1} \mathbf{h} }$ is a complexity term, and $\mathbf{h} = (h(\bx_{1, \hi}), \dots, h(\bx_{T, \hi}))^\top \in \bbr^T$.
Note that \sysn and \cite{wang2021neural} have the same trivial label complexity $\calo(T)$ in this difficult case.
According to the definition of effective dimension $\widetilde{d}$ in \cite{zhou2020neural}, the above regret bound obtained by \cite{wang2021neural} can be represented by:
\begin{equation} \label{eq:baseline}
\widetilde{\mathbf{R}}_T \leq \calo(\widetilde{d} \log (1 + T) ) \sqrt{ T ( \widetilde{d} \log (1 + T)  + S^2 )}  \ \ \  \text{and} \ \ \  \widetilde{d} = \frac{\log \det (I + \mathbf{H})}{\log (1 + T)} 
\end{equation}

\begin{remark} \label{remark:s}
The instance-dependent complexity term $\mu$ reflects the possible minimal regression error on the data instances caused by the functions induced by  $\calb(\btheta^2_0, \nu)$ controlled by $\nu$. Such complexity term is first introduced in \cite{cao2019generalization}. When $\nu$ is small, the corresponding ball $\calb(\btheta^2_0, \nu)$ is small, so $\mu$ tends to be large; Otherwise, when $\nu$ is large, $\mu$ tends to be small. In particular, when setting $\nu = \calo(1)$, Theorem \ref{theo1} and Lemma \ref{lemma:1}  suggests that if the data can be learned by a function in the function class formed by  $\calb(\btheta^2_0, \calo(1))$ with the small training error, then \sysn will have the regret with order $\widetilde{\calo}(\sqrt{T})$. Note that \cite{wang2021neural} has the complexity term $S$ as well, to reflect the boundary of optimal parameters specific to the data. 
\end{remark}


\begin{remark}
Theorem \ref{theo1} and Lemma \ref{lemma:1} do not depend on $\widetilde{d}$. The effective dimension $\widetilde{d}$ was first introduced in \cite{valko2013finite} and then used in \cite{zhou2020neural}, which can be thought of as the non-linear dimensionalities in the NTK kernel space. However, $\widetilde{d}$ can be $p= m + mkd+m^2(L-1)$ in the worst case, i.e., $\widetilde{d} \gg T$ (see details in Appendix \ref{sub:effective}).  Eq.\eqref{eq:baseline} has the term $\calo(\widetilde{d})$ and  thus the regret bound obtained by \cite{wang2021neural} can explode due to $\widetilde{d}$. This is because the analysis of \cite{wang2021neural} closely depends on NTK, i.e., to apply Confidence Ellipsoid bound (Theorem 2 in \cite{2011improved}) to the NTK approximation. This procedure inevitably bind their regret bound to the determinant of NTK that can have a very large magnitude. In contrast,  
Eq.\eqref{eq:lemma1}  does not have the term $\widetilde{d}$, because our analysis does not depend on the NTK approximation and \sysn directly utilizes the property of over-parameterized neural networks, i.e., the convergence error $\mu$ and the generalization concentration bound (Lemma \ref{lemma:thetabound}). These two terms are independent of $\widetilde{d}$, which paves the way for \sysn to remove the curse of $\widetilde{d}$. 
\end{remark}

\begin{remark}
Theorem \ref{theo1} and Lemma \ref{lemma:1} improve the regret by a multiplicative factor $\calo(\log T)$ over \cite{wang2021neural}. 
Note that the analysis of \cite{wang2021neural} is built for binary classification and thus $k=2$ in Theorem \ref{theo1} and Lemma \ref{lemma:1}.
This improvement stems from the different analysis workflow of \sysn from \cite{wang2021neural}. Again, our analysis does not rely on NTK approximation and it is built on the convergence and generalization bound of wide neural networks.
\end{remark}

\begin{remark}
Our proof workflow of Theorem \ref{theo1} and Lemma \ref{lemma:1} is inspired by \cite{ban2021ee}. Compared to \cite{ban2021ee}, we provide the first regret bound supporting  mini-batch SGD with warm-start and a  more generic generalization bound (Lemma 7.3) that holds for every arm (class). Moreover, we carry out the performance analysis of query decision-maker (Lemma 7.5), which is a new addition.
\end{remark}

For the label complexity, $\mathbf{N}_T$ has the trivial $\calo(T)$ complexity which is the same as Theorem 1 in \cite{wang2021neural} (with the exponent $\alpha = 0$). Because we have to consider the worst case where the unique Bayes-optimal class does not exist, i.e., given $\bx_{t,i}, i \in[k]$, there does not exist $i^\ast$ such that $h(\bx_{t,i^\ast}) > h(\bx_{t,i}), \forall i \in [k] \setminus \{i^\ast\}$.
Therefore, we provide the following analysis and show that $\mathbf{R}_T$ and $\mathbf{N}_T$ can be upper bounded by constants as long as there exists a unique Bayes-optimal class for the input instances, described by the following mild margin assumption.

\begin{assumption}[$\epsilon$-margin] \label{assum:margin}
In round $t \in [T]$, given an instance $\bx_t$ and the label $y_t$, then $\bx_t$ has the $\epsilon$-Unique optimal class if there exists $\epsilon > 0$ such that
\begin{equation}
 \bbp(\by_{t, y_t} = \by_{t, i^\ast} | \bx_{t}) -  \bbp(\by_{t, y_t} = \by_{t, i^\circ} | \bx_{t}) \geq \epsilon,
\end{equation}
where $i^\ast = \arg \max_{i \in [k]}h(\bx_{t,i}) $ is the Bayes-optimal class and $i^\circ = \arg \max_{i \in ([k] \setminus  \{ i^\ast\})} h(\bx_{t,i}) $ is the second  Bayes-optimal class.
\end{assumption}

Given any $i \in [k]$, let $i$ be a fixed index, i.e., suppose there exist a policy $\pi_i$ which always select the $i$-th context $(\bx_{t,i}, r^1_{t,i})$ for every round $t \in [T]$. Then, in round $t$, we have the collected data by $\Omega_i$: $\mathcal{H}_{t-1}^{1,i} = \{ \bx_{\tau,i}, r^1_{\tau,i}\}_{\tau= 1}^{t-1}$. Then, let $\widehat{\btheta}^{1, i}_{t - 1}$ represent the parameters trained only on $\mathcal{H}_{t-1}^{1,i}$ using Algorithm \ref{alg:BGD} with $\pi_i$ and $r^{2,i}_{t, i} = r^1_{t,i} -  f_1(\bx_{t, i}; \hbtheta_{t-1}^{1,i})$.

\begin{theorem} \label{theo2}
Suppose the instances that are drawn from $\cald$ satisfy Assumption \ref{assum:margin}. Then, given the number of rounds $T$, for any $\delta \in (0, 1),  \gamma >1, \epsilon \in (0 , 1)$, $\nu > 0$, suppose $m \geq \widetilde{\Omega}(\text{poly} (T, k, L, \nu)), \eta_1 = \eta_2 = \Theta(\frac{ \nu \kappa }{\sqrt{T} m})$, and $\mu$ satisfies the conditions in Theorem \ref{theo1} and $ \underset{i \in [k]}{\max} \left\{ \underset{\widetilde{\btheta}^2 \in \calb(\btheta^2_0, \nu) }{\inf}  \sum_{t=1}^T 
\left(      
f_2 \left( \bx_{t, \hi} ; \widetilde{\btheta}^2 \right) - r_{t,i}^{2,i} 
\right)^2  \right \} \leq \mu$. Then, with probability at least $1 -\delta$ over the initialization of $\btheta^1_0, \btheta^2_0$, 
there exists a small enough constant $\kappa$, such that Algorithm \ref{alg:main} achieves the following regret bound:
\begin{equation}
\mathbf{R}_T \leq  (2\sqrt{\bar{\calt}} -1) \left[ \calo \left(  \frac{6L\nu + 4\sqrt{\mu}}{\sqrt{2}} \right) + \sqrt{2 \log ( \calo(Tk)/\delta)} \right]  \ \ \  \mathbf{N}_T \leq   \bar{\calt}
\end{equation}
where 
$   \bar{\calt} =  \frac{12 (\gamma +1)^2 \cdot \left[  2 \mu + 9 L^2 \nu^2 C_1^2 + 2\log(C_2 T k / \delta) \right] }{\epsilon^2}. 
$
Suppose the Bayes-optimal classifier has zero classification errors, i.e., $L(\by_{t, i^\ast}, \by_{t, y_t} )  = 0, t \in [T] $. It holds that
\begin{equation}
\begin{cases}
R_T \leq  \calo \left(\frac{6L\nu + 4\sqrt{\mu} }{\sqrt{2T}} \right) + 2 \sqrt{ \frac{2 \log ( \calo(Tk)/\delta) }{T}}, \ \ \text{if} \ \ T \leq   \bar{\calt} ;   \\
R_T = 0,  \text{else} . 
\end{cases}
\end{equation}
\end{theorem}

\begin{remark}
Theorem \ref{theo2} provides the upper bound for $\mathbf{R}_T$ with order of $\calo(\log T)$. When other parameters are fixed,  this indicates $\mathbf{R}_T$ is upper bounded by $\calo(\log T)$. Moreover, the analysis of $R_T$ indicates that \sysn can achieve the same performance as Bayes-optimal classifier with high confidence after $\calo(\log T)$ number of rounds (i.e. $ T > \bar{\calt}$). In Theorem 1 of \cite{wang2021neural} ( with the exponent $\alpha \rightarrow +\infty$ equivalent to Assumption \ref{assum:margin}), $\widetilde{\mathbf{R}}_T \leq \calo(\widetilde{d} \log (1 + T) ) \sqrt{ ( \widetilde{d} \log (1 + T)  + S^2 )} $ that still is dependent on $\widetilde{d}$ because NTK depends on $\widetilde{d}$.
\end{remark}

\section{Experiments} \label{sec:exp}

\begin{figure*}[th]
\centering
\subfigure[Phishing]{
\includegraphics[width=0.32\textwidth]{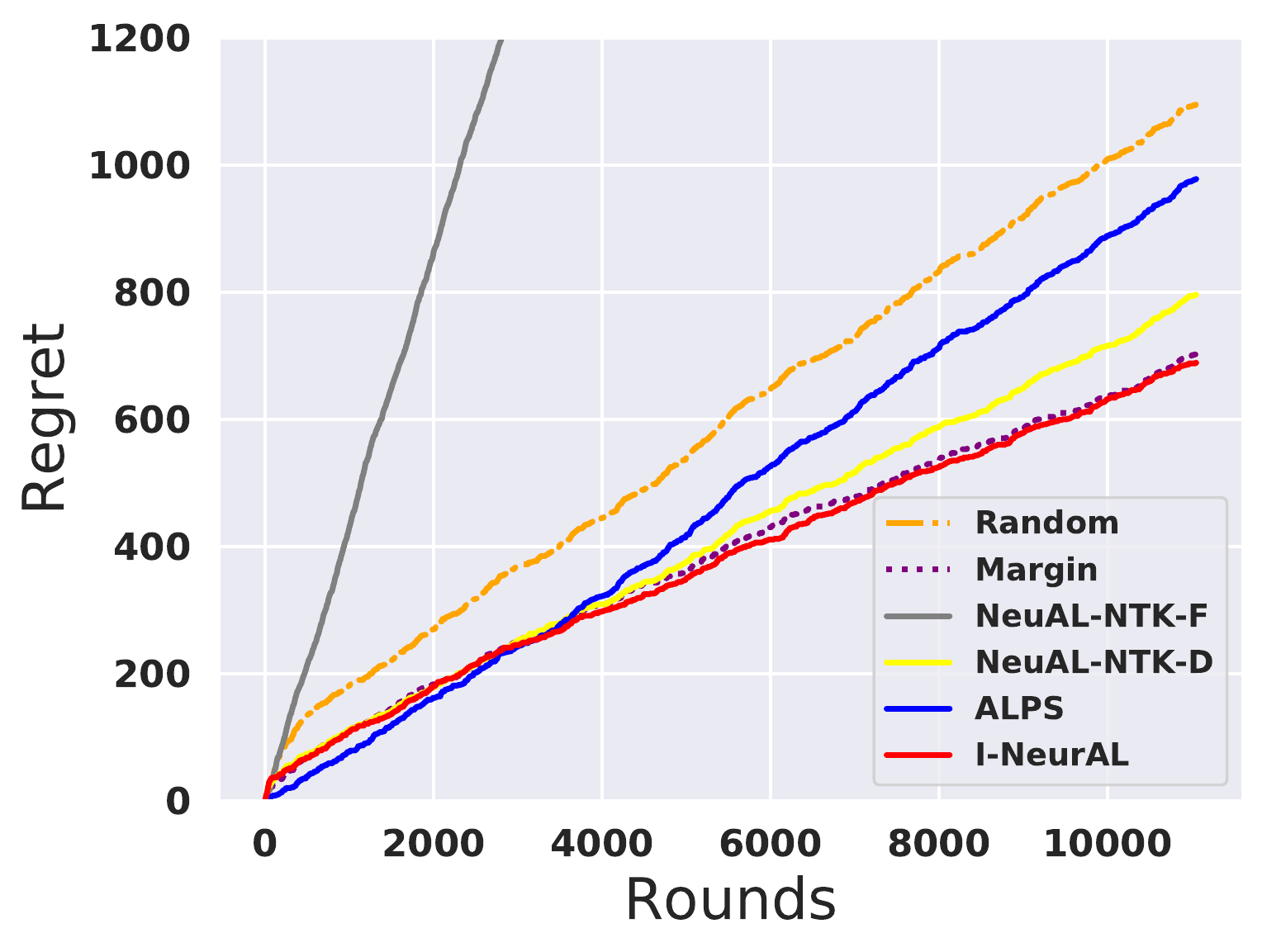}}
\subfigure[IJCNN]{
\includegraphics[width=0.32\textwidth]{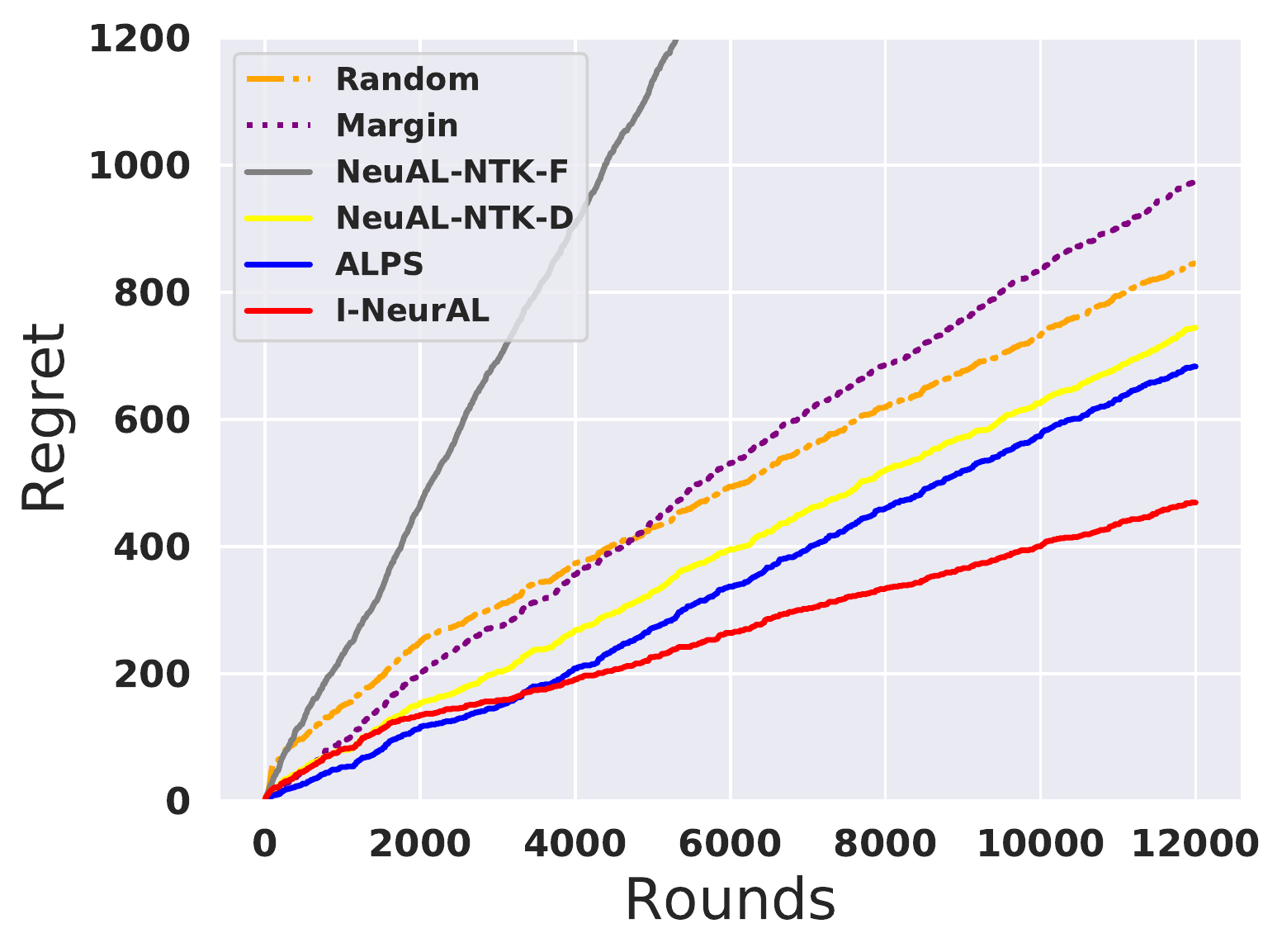}}
\subfigure[Letter]{
\includegraphics[width=0.32\textwidth]{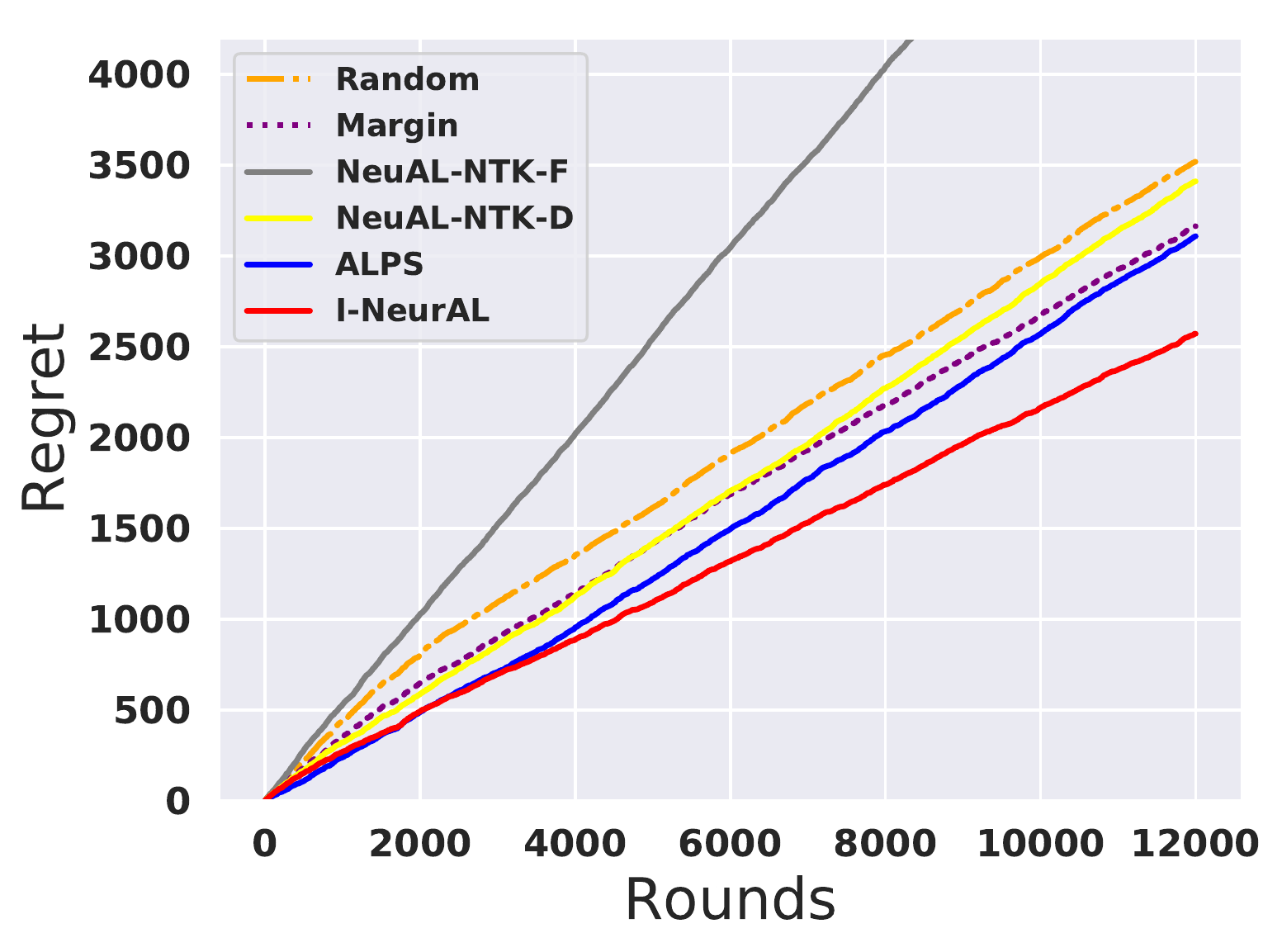}}
\subfigure[Fashion]{
\includegraphics[width=0.32\textwidth]{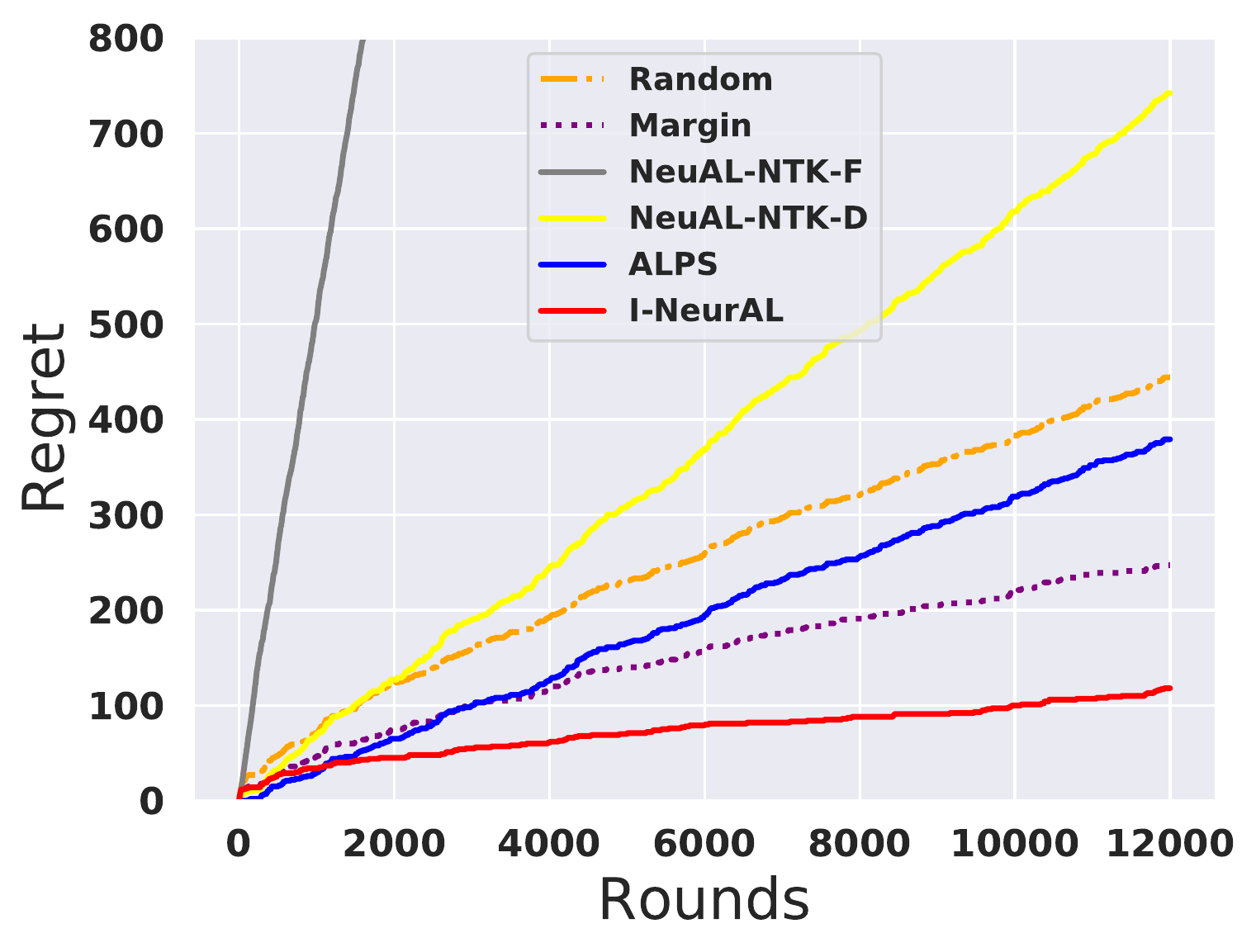}}
\subfigure[MNIST]{
\includegraphics[width=0.32\textwidth]{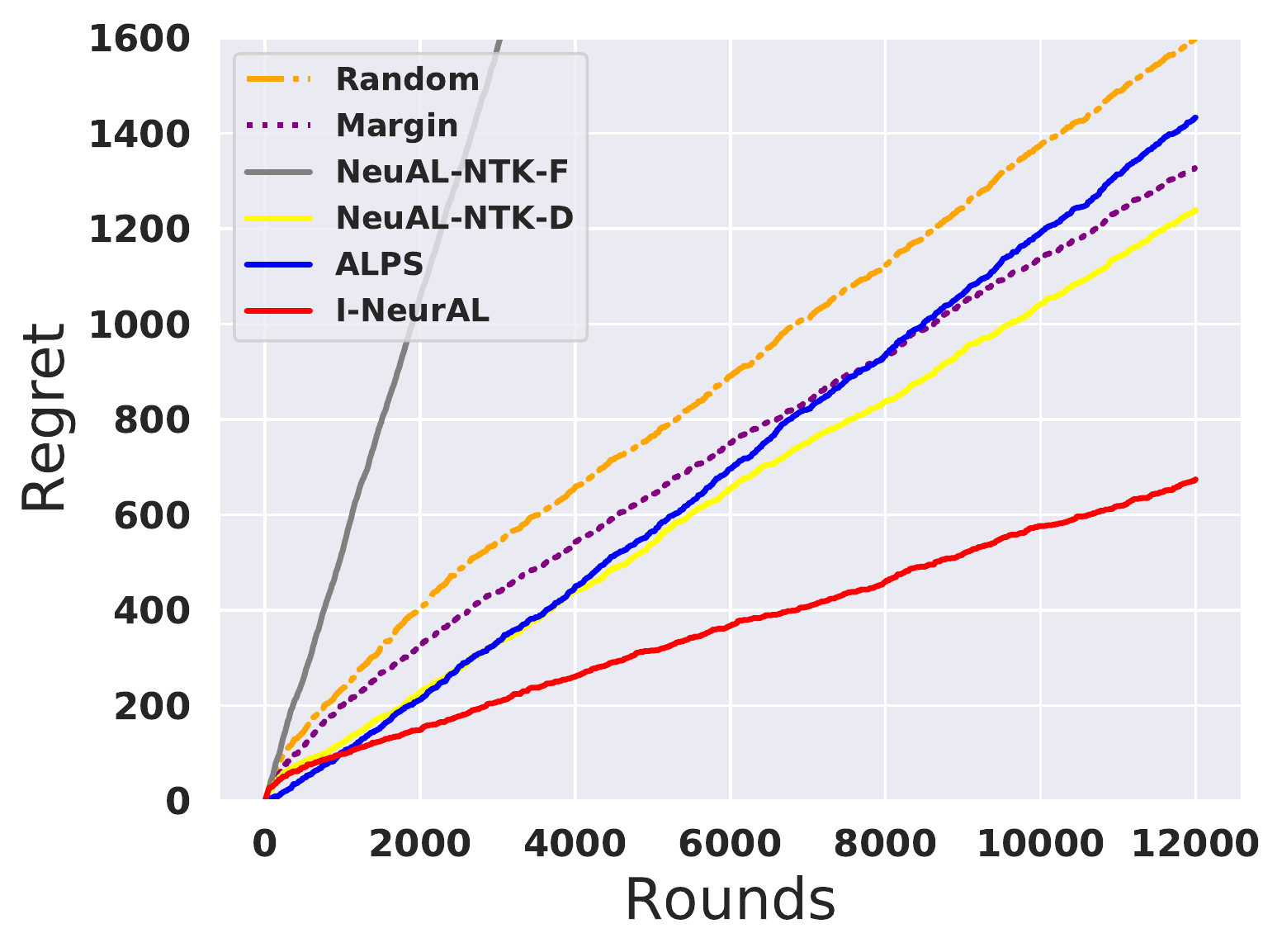}}
\subfigure[CIFAR-10]{
\includegraphics[width=0.32\textwidth]{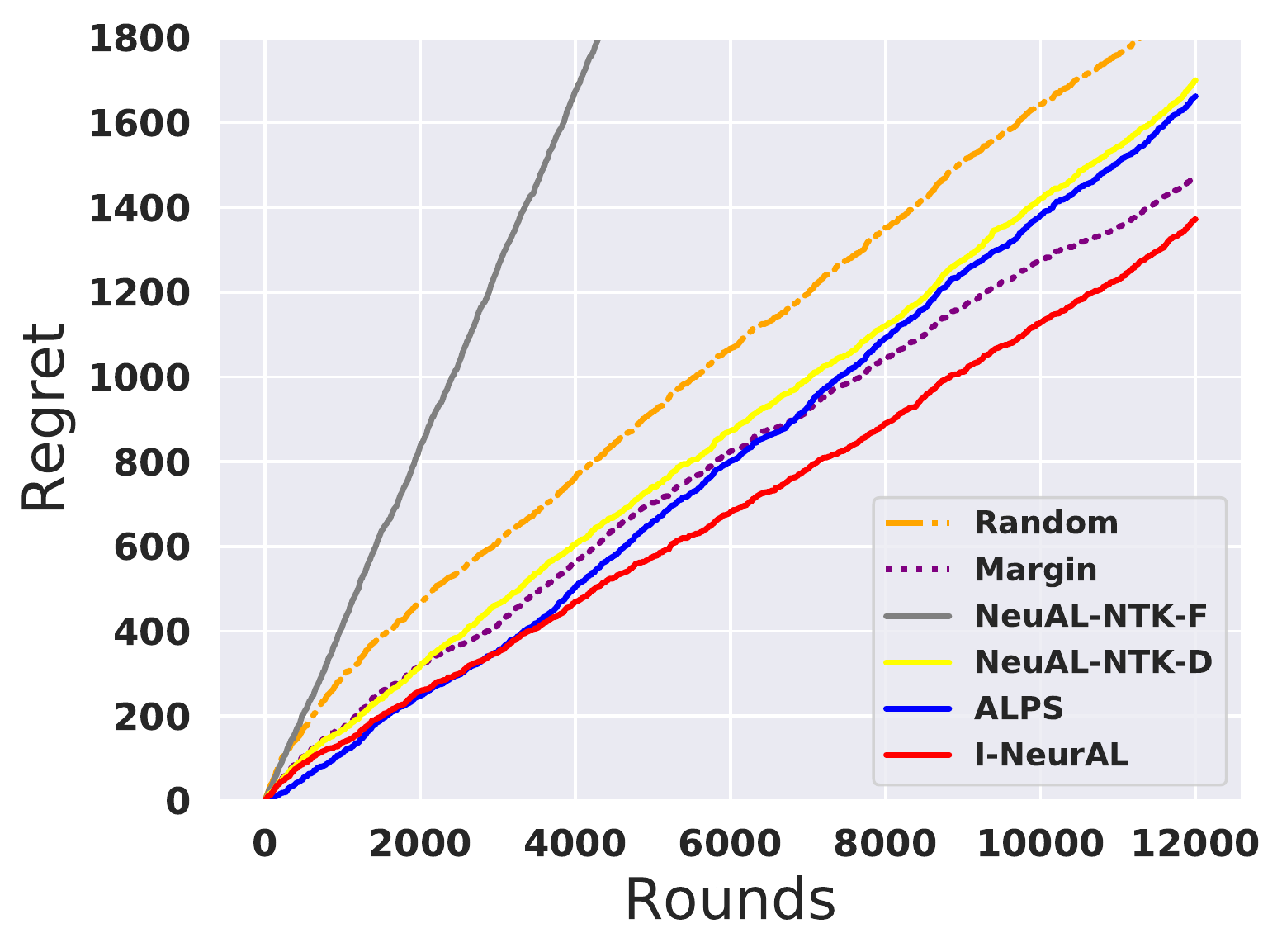}}
\caption{Regret comparison on six data sets. \sysn outperforms all baselines.}
\vspace{-1em}
\label{overall}
\end{figure*}

In this section, we evaluate \sysn on public classification data sets compared with state-of-the-art (SOTA) baselines. Due to the space limit, we only report the main results here and leave the implementation details and parameter sensitivity in the Appendix \ref{append:exp}. Codes are available\footnote{\url{https://github.com/matouk98/I-NeurAL}}.

We report the experimental results on the following six data sets: Phishing\footnote{\url{https://www.csie.ntu.edu.tw/~cjlin/libsvmtools/datasets/binary.html}}, IJCNN~\cite{prokhorov2001ijcnn}, Letter~\cite{cohen2017emnist}, Fashion~\cite{xiao2017/online}, MNIST~\cite{lecun1998gradient} and CIFAR-10~\cite{krizhevsky2009learning}. 
In each round, one instance is randomly drawn from the data set and the algorithm is compelled to make prediction on it. Then, the regret is $1$ if the prediction does not match the label; the regret is $0$, otherwise. At the same time, if the algorithm decides to observe the label, it costs one query budget. As the algorithm may abusively make label queries, we restrict the query budget to $3 \%$ of the total number of instances in the data set for fair comparison.

The compared baselines are described as follows. (1) \textbf{Random}: The NN classifier queries the label with a fixed probability $p$ until the query budget is exhausted; (2) \textbf{Margin}: The NN classifier queries the label when the predicted probability is lower than a threshold. These two baselines are used in \cite{desalvo2021online}. (3) \textbf{NeuAL-NTK-F (Algorithm 1 in ~\cite{wang2021neural}}: This model makes predictions based on the frozen NTK approximation coming with an Upper-Confidence-Bound(UCB)-based exploration strategy.  (4) \textbf{NeuAL-NTK-D (Algorithm 3 in \cite{wang2021neural})}: The prediction is made based on the NN classifier with a UCB while the NTK is updated accordingly. (5) \textbf{ALPS~\cite{desalvo2021online}}: Given a class of pre-trained hypotheses, the hypothesis minimizing the logistic loss of labeled and pseudo-labeled data is chosen to make predictions and the label query is based on the disagreement of different hypotheses.

\para{Results}.
The regret comparison on six data sets is shown in Table~\ref{totreg} and Figure~\ref{overall}. \sysn consistently outperforms all baselines across all data sets. In particular, \sysn surpasses the best baseline by 31.3\%, 45.6\%, 52.2\% on IJCNN, MNIST, Fashion respectively. 
Since NeuAL-NTK-F uses frozen NTK approximation, the new knowledge of each round is barely utilized by the neural network and thus it turns into the worst baseline. 
NeuAL-NTK-D updates the network parameters with gradient descent and queries the label based on the uncertainty estimation. However, its upper confidence bound is still based on the confidence ellipsoid. 
Instead, \sysn leverages the representation power of neural networks for both exploitation and exploration.
ALPS maintains a class of pre-trained hypotheses and tries to make the best decisions based on these hypotheses.
Nevertheless, the model parameters are fixed before the online active learning process. Hence, ALPS is not able to take the new knowledge obtained by queries into account and its performance is highly restricted by the hypothesis class.
Although Margin algorithm is simple and straightforward, it exhibits great empirical performance in practice. This observation is consistent with other studies~\cite{yang2018benchmark}~\cite{desalvo2021online}.
However, Margin algorithm does not incorporate the exploitation portion and the query criterion is not adaptive to difference instances, thus still outperformed by \sysn.

\begin{table}[t]
\centering
\scalebox{0.8}{
\begin{tabular}{c|cccccc}
\hline
            & Phishing     & IJCNN        & Letter          & Fashion       & MNIST        & CIFAR-10      \\ \hline
Random        & 1095         & 845          & 3519          & 444          & 1599         & 1910          \\
Margin      & \underline{704}          & 974          &      3164     & \underline{247}         & 1327         & \underline{1474}          \\
NeuAL-NTK-F & 4898         & 2684         & 6066          & 6001         & 6192         & 5007          \\
NeuAL-NTK-D & 796          & 744          & 3410          & 742          & \underline{1239}         & 1700          \\
ALPS        & 978          & \underline{683}          & \underline{3108}          & 379          & 1433         & 1662          \\ \hline
\sysn    & $\mathbf{689 (\uparrow 2.1\%)}$ & $\mathbf{469 (\uparrow 31.3\%)}$ & $\mathbf{2571 (\uparrow 17.3\%)}$ & $\mathbf{118 (\uparrow 52.2\%)}$ & $\mathbf{674 (\uparrow 45.6\%)}$ & $\mathbf{1372 (\uparrow 6.9\%)}$ \\ \hline
\end{tabular}}
\caption{Total regret comparison. }
\label{totreg}
\end{table}
\vspace{-1em}

\section{Conclusion} \label{sec:con}
In this paper, we introduce two regret metrics and propose a novel neural-based algorithm (\sysn) tailored for the streaming setting of non-parametric active learning. We carefully design its exploration strategy, query decision-maker, update rules, and training procedure, which lead to both the theoretical and empirical improvement compared to SOTA \cite{wang2021neural}. In the regret analysis, we provide an instance-dependent performance guarantee. On the other hand, we empirically show that \sysn consistently achieves better accuracy under the same query budget than the strong baselines including the SOTA work \cite{wang2021neural} and \cite{desalvo2021online}.

\section*{Acknowledgements}
This work is supported by NSF (IIS-1947203, IIS-2117902, IIS-2137468, IIS-2002540, DMS-2134079, IIS-2131335, OAC-2130835, and DBI-2021898), DARPA (HR001121C0165), ARO (W911NF2110088), and C3.ai. The views and conclusions are those of the authors and should not be interpreted as representing the official policies of the funding agencies or the government.

\setcitestyle{numbers}
\bibliographystyle{abbrv}
\bibliography{ref.bib}

\begin{thebibliography}{10}

\bibitem{2011improved}
Y.~Abbasi-Yadkori, D.~P{\'a}l, and C.~Szepesv{\'a}ri.
\newblock Improved algorithms for linear stochastic bandits.
\newblock In {\em Advances in Neural Information Processing Systems}, pages
  2312--2320, 2011.

\bibitem{allen2019convergence}
Z.~Allen-Zhu, Y.~Li, and Z.~Song.
\newblock A convergence theory for deep learning via over-parameterization.
\newblock In {\em International Conference on Machine Learning}, pages
  242--252. PMLR, 2019.

\bibitem{arora2019exact}
S.~Arora, S.~S. Du, W.~Hu, Z.~Li, R.~R. Salakhutdinov, and R.~Wang.
\newblock On exact computation with an infinitely wide neural net.
\newblock In {\em Advances in Neural Information Processing Systems}, pages
  8141--8150, 2019.

\bibitem{ash2021gone}
J.~Ash, S.~Goel, A.~Krishnamurthy, and S.~Kakade.
\newblock Gone fishing: Neural active learning with fisher embeddings.
\newblock {\em Advances in Neural Information Processing Systems}, 34, 2021.

\bibitem{ash2019deep}
J.~T. Ash, C.~Zhang, A.~Krishnamurthy, J.~Langford, and A.~Agarwal.
\newblock Deep batch active learning by diverse, uncertain gradient lower
  bounds.
\newblock {\em arXiv preprint arXiv:1906.03671}, 2019.

\bibitem{awasthi2014power}
P.~Awasthi, M.~F. Balcan, and P.~M. Long.
\newblock The power of localization for efficiently learning linear separators
  with noise.
\newblock In {\em Proceedings of the forty-sixth annual ACM symposium on Theory
  of computing}, pages 449--458, 2014.

\bibitem{balcan2009agnostic}
M.-F. Balcan, A.~Beygelzimer, and J.~Langford.
\newblock Agnostic active learning.
\newblock {\em Journal of Computer and System Sciences}, 75(1):78--89, 2009.

\bibitem{balcan2007margin}
M.-F. Balcan, A.~Broder, and T.~Zhang.
\newblock Margin based active learning.
\newblock In {\em International Conference on Computational Learning Theory},
  pages 35--50. Springer, 2007.

\bibitem{ban2020generic}
Y.~Ban and J.~He.
\newblock Generic outlier detection in multi-armed bandit.
\newblock In {\em Proceedings of the 26th ACM SIGKDD International Conference
  on Knowledge Discovery \& Data Mining}, pages 913--923, 2020.

\bibitem{ban2021local}
Y.~Ban and J.~He.
\newblock Local clustering in contextual multi-armed bandits.
\newblock In {\em Proceedings of the Web Conference 2021}, pages 2335--2346,
  2021.

\bibitem{ban2021multi}
Y.~Ban, J.~He, and C.~B. Cook.
\newblock Multi-facet contextual bandits: A neural network perspective.
\newblock In {\em The 27th {ACM} {SIGKDD} Conference on Knowledge Discovery and
  Data Mining, Virtual Event, Singapore, August 14-18, 2021}, pages 35--45,
  2021.

\bibitem{ban2022neural}
Y.~Ban, Y.~Qi, T.~Wei, and J.~He.
\newblock Neural collaborative filtering bandits via meta learning.
\newblock {\em ArXiv abs/2201.13395}, 2022.

\bibitem{ban2021ee}
Y.~Ban, Y.~Yan, A.~Banerjee, and J.~He.
\newblock {EE}-net: Exploitation-exploration neural networks in contextual
  bandits.
\newblock In {\em International Conference on Learning Representations}, 2022.

\bibitem{beygelzimer2009importance}
A.~Beygelzimer, S.~Dasgupta, and J.~Langford.
\newblock Importance weighted active learning.
\newblock In {\em Proceedings of the 26th annual international conference on
  machine learning}, pages 49--56, 2009.

\bibitem{brinker2003incorporating}
K.~Brinker.
\newblock Incorporating diversity in active learning with support vector
  machines.
\newblock In {\em Proceedings of the 20th international conference on machine
  learning (ICML-03)}, pages 59--66, 2003.

\bibitem{cao2019generalization}
Y.~Cao and Q.~Gu.
\newblock Generalization bounds of stochastic gradient descent for wide and
  deep neural networks.
\newblock {\em Advances in Neural Information Processing Systems},
  32:10836--10846, 2019.

\bibitem{citovsky2021batch}
G.~Citovsky, G.~DeSalvo, C.~Gentile, L.~Karydas, A.~Rajagopalan,
  A.~Rostamizadeh, and S.~Kumar.
\newblock Batch active learning at scale.
\newblock {\em Advances in Neural Information Processing Systems}, 34, 2021.

\bibitem{cohen2017emnist}
G.~Cohen, S.~Afshar, J.~Tapson, and A.~Van~Schaik.
\newblock Emnist: Extending mnist to handwritten letters.
\newblock In {\em 2017 international joint conference on neural networks
  (IJCNN)}, pages 2921--2926. IEEE, 2017.

\bibitem{cohn1996active}
D.~A. Cohn, Z.~Ghahramani, and M.~I. Jordan.
\newblock Active learning with statistical models.
\newblock {\em Journal of artificial intelligence research}, 4:129--145, 1996.

\bibitem{culotta2005reducing}
A.~Culotta and A.~McCallum.
\newblock Reducing labeling effort for structured prediction tasks.
\newblock In {\em AAAI}, volume~5, pages 746--751, 2005.

\bibitem{dasgupta2005analysis}
S.~Dasgupta, A.~T. Kalai, and C.~Monteleoni.
\newblock Analysis of perceptron-based active learning.
\newblock In {\em International conference on computational learning theory},
  pages 249--263. Springer, 2005.

\bibitem{desalvo2021online}
G.~DeSalvo, C.~Gentile, and T.~S. Thune.
\newblock Online active learning with surrogate loss functions.
\newblock {\em Advances in Neural Information Processing Systems}, 34, 2021.

\bibitem{goodfellow2016deep}
I.~Goodfellow, Y.~Bengio, and A.~Courville.
\newblock {\em Deep learning}.
\newblock MIT press, 2016.

\bibitem{hanneke2007bound}
S.~Hanneke.
\newblock A bound on the label complexity of agnostic active learning.
\newblock In {\em Proceedings of the 24th international conference on Machine
  learning}, pages 353--360, 2007.

\bibitem{hanneke2014theory}
S.~Hanneke et~al.
\newblock Theory of disagreement-based active learning.
\newblock {\em Foundations and Trends{\textregistered} in Machine Learning},
  7(2-3):131--309, 2014.

\bibitem{hanneke2019surrogate}
S.~Hanneke and L.~Yang.
\newblock Surrogate losses in passive and active learning.
\newblock {\em Electronic Journal of Statistics}, 13(2):4646--4708, 2019.

\bibitem{ntk2018neural}
A.~Jacot, F.~Gabriel, and C.~Hongler.
\newblock Neural tangent kernel: Convergence and generalization in neural
  networks.
\newblock In {\em Advances in neural information processing systems}, pages
  8571--8580, 2018.

\bibitem{joshi2009multi}
A.~J. Joshi, F.~Porikli, and N.~Papanikolopoulos.
\newblock Multi-class active learning for image classification.
\newblock In {\em 2009 ieee conference on computer vision and pattern
  recognition}, pages 2372--2379. IEEE, 2009.

\bibitem{kapoor2007active}
A.~Kapoor, K.~Grauman, R.~Urtasun, and T.~Darrell.
\newblock Active learning with gaussian processes for object categorization.
\newblock In {\em 2007 IEEE 11th international conference on computer vision},
  pages 1--8. IEEE, 2007.

\bibitem{kim2021lada}
Y.-Y. Kim, K.~Song, J.~Jang, and I.-c. Moon.
\newblock Lada: Look-ahead data acquisition via augmentation for deep active
  learning.
\newblock {\em Advances in Neural Information Processing Systems}, 34, 2021.

\bibitem{krizhevsky2009learning}
A.~Krizhevsky, G.~Hinton, et~al.
\newblock Learning multiple layers of features from tiny images.
\newblock 2009.

\bibitem{lecun1998gradient}
Y.~LeCun, L.~Bottou, Y.~Bengio, and P.~Haffner.
\newblock Gradient-based learning applied to document recognition.
\newblock {\em Proceedings of the IEEE}, 86(11):2278--2324, 1998.

\bibitem{lewis1994sequential}
D.~D. Lewis and W.~A. Gale.
\newblock A sequential algorithm for training text classifiers.
\newblock In {\em SIGIR’94}, pages 3--12. Springer, 1994.

\bibitem{locatelli2017adaptivity}
A.~Locatelli, A.~Carpentier, and S.~Kpotufe.
\newblock Adaptivity to noise parameters in nonparametric active learning.
\newblock In {\em Proceedings of the 2017 Conference on Learning Theory, PMLR},
  2017.

\bibitem{minsker2012plug}
S.~Minsker.
\newblock Plug-in approach to active learning.
\newblock {\em Journal of Machine Learning Research}, 13(1), 2012.

\bibitem{moon2020confidence}
J.~Moon, J.~Kim, Y.~Shin, and S.~Hwang.
\newblock Confidence-aware learning for deep neural networks.
\newblock In {\em international conference on machine learning}, pages
  7034--7044. PMLR, 2020.

\bibitem{mussmann2018uncertainty}
S.~Mussmann and P.~S. Liang.
\newblock Uncertainty sampling is preconditioned stochastic gradient descent on
  zero-one loss.
\newblock {\em Advances in Neural Information Processing Systems}, 31, 2018.

\bibitem{prokhorov2001ijcnn}
D.~Prokhorov.
\newblock Ijcnn 2001 neural network competition.
\newblock {\em Slide presentation in IJCNN}, 1(97):38, 2001.

\bibitem{yunzheneural}
Y.~Qi, Y.~Ban, and J.~He.
\newblock Neural bandit with arm group graph.
\newblock In {\em Proceedings of the 28th ACM SIGKDD Conference on Knowledge
  Discovery and Data Mining}, KDD '22, page 1379–1389, New York, NY, USA,
  2022. Association for Computing Machinery.

\bibitem{ren2021survey}
P.~Ren, Y.~Xiao, X.~Chang, P.-Y. Huang, Z.~Li, B.~B. Gupta, X.~Chen, and
  X.~Wang.
\newblock A survey of deep active learning.
\newblock {\em ACM Computing Surveys (CSUR)}, 54(9):1--40, 2021.

\bibitem{roy2001toward}
N.~Roy and A.~McCallum.
\newblock Toward optimal active learning through monte carlo estimation of
  error reduction.
\newblock {\em ICML, Williamstown}, 2:441--448, 2001.

\bibitem{schroder2020survey}
C.~Schr{\"o}der and A.~Niekler.
\newblock A survey of active learning for text classification using deep neural
  networks.
\newblock {\em arXiv preprint arXiv:2008.07267}, 2020.

\bibitem{sener2017active}
O.~Sener and S.~Savarese.
\newblock Active learning for convolutional neural networks: A core-set
  approach.
\newblock {\em arXiv preprint arXiv:1708.00489}, 2017.

\bibitem{settles2009active}
B.~Settles.
\newblock Active learning literature survey.
\newblock 2009.

\bibitem{tan2021diversity}
W.~Tan, L.~Du, and W.~Buntine.
\newblock Diversity enhanced active learning with strictly proper scoring
  rules.
\newblock {\em Advances in Neural Information Processing Systems}, 34, 2021.

\bibitem{valko2013finite}
M.~Valko, N.~Korda, R.~Munos, I.~Flaounas, and N.~Cristianini.
\newblock Finite-time analysis of kernelised contextual bandits.
\newblock {\em arXiv preprint arXiv:1309.6869}, 2013.

\bibitem{wang2021deep}
H.~Wang, W.~Huang, A.~Margenot, H.~Tong, and J.~He.
\newblock Deep active learning by leveraging training dynamics.
\newblock {\em arXiv preprint arXiv:2110.08611}, 2021.

\bibitem{wang2021neural}
Z.~Wang, P.~Awasthi, C.~Dann, A.~Sekhari, and C.~Gentile.
\newblock Neural active learning with performance guarantees.
\newblock {\em Advances in Neural Information Processing Systems}, 34, 2021.

\bibitem{xiao2017/online}
H.~Xiao, K.~Rasul, and R.~Vollgraf.
\newblock Fashion-mnist: a novel image dataset for benchmarking machine
  learning algorithms, 2017.

\bibitem{yang2018benchmark}
Y.~Yang and M.~Loog.
\newblock A benchmark and comparison of active learning for logistic
  regression.
\newblock {\em Pattern Recognition}, 83:401--415, 2018.

\bibitem{zhang2018efficient}
C.~Zhang.
\newblock Efficient active learning of sparse halfspaces.
\newblock In {\em Conference on Learning Theory}, pages 1856--1880. PMLR, 2018.

\bibitem{zhang2020efficient}
C.~Zhang, J.~Shen, and P.~Awasthi.
\newblock Efficient active learning of sparse halfspaces with arbitrary bounded
  noise.
\newblock {\em Advances in Neural Information Processing Systems},
  33:7184--7197, 2020.

\bibitem{zhang2020neural}
W.~Zhang, D.~Zhou, L.~Li, and Q.~Gu.
\newblock Neural thompson sampling.
\newblock In {\em International Conference on Learning Representations}, 2021.

\bibitem{Zhang_Tong_Xia_Zhu_Chi_Ying_2022}
Y.~Zhang, H.~Tong, Y.~Xia, Y.~Zhu, Y.~Chi, and L.~Ying.
\newblock Batch active learning with graph neural networks via multi-agent deep
  reinforcement learning.
\newblock {\em Proceedings of the AAAI Conference on Artificial Intelligence},
  36:9118--9126, 2022.

\bibitem{zhou2020neural}
D.~Zhou, L.~Li, and Q.~Gu.
\newblock Neural contextual bandits with ucb-based exploration.
\newblock In {\em International Conference on Machine Learning}, pages
  11492--11502. PMLR, 2020.

\end{thebibliography}

\begin{enumerate}

\item For all authors...
\begin{enumerate}
  \item Do the main claims made in the abstract and introduction accurately reflect the paper's contributions and scope?
    \answerYes{}
  \item Did you describe the limitations of your work?
    \answerNo{}
  \item Did you discuss any potential negative societal impacts of your work?
    \answerNA{}
  \item Have you read the ethics review guidelines and ensured that your paper conforms to them?
    \answerYes{}
\end{enumerate}

\item If you are including theoretical results...
\begin{enumerate}
  \item Did you state the full set of assumptions of all theoretical results?
    \answerYes{}
	\item Did you include complete proofs of all theoretical results?
   \answerYes{}
\end{enumerate}

\item If you ran experiments...
\begin{enumerate}
  \item Did you include the code, data, and instructions needed to reproduce the main experimental results (either in the supplemental material or as a URL)?
    \answerYes{}
  \item Did you specify all the training details (e.g., data splits, hyperparameters, how they were chosen)?
    \answerYes{}
	\item Did you report error bars (e.g., with respect to the random seed after running experiments multiple times)?
    \answerYes{} The random seed is fixed to 42 in all the experiments.
	\item Did you include the total amount of compute and the type of resources used (e.g., type of GPUs, internal cluster, or cloud provider)?
    \answerYes{}
\end{enumerate}

\item If you are using existing assets (e.g., code, data, models) or curating/releasing new assets...
\begin{enumerate}
  \item If your work uses existing assets, did you cite the creators?
    \answerYes{}
  \item Did you mention the license of the assets?
    \answerYes{}
  \item Did you include any new assets either in the supplemental material or as a URL?
    \answerYes{}
  \item Did you discuss whether and how consent was obtained from people whose data you're using/curating?
    \answerYes{}
  \item Did you discuss whether the data you are using/curating contains personally identifiable information or offensive content?
    \answerYes{}
\end{enumerate}

\item If you used crowdsourcing or conducted research with human subjects...
\begin{enumerate}
  \item Did you include the full text of instructions given to participants and screenshots, if applicable?
    \answerNA{}
  \item Did you describe any potential participant risks, with links to Institutional Review Board (IRB) approvals, if applicable?
    \answerNA{}
  \item Did you include the estimated hourly wage paid to participants and the total amount spent on participant compensation?
    \answerNA{}
\end{enumerate}

\end{enumerate}
\clearpage

\clearpage

In this \textbf{Appendix}, we first present the proof of Theorem \ref{theo1} and \ref{theo2} in Section \ref{sec:profthe1}; second, show the proof of Lemma \ref{lemma:1} in Section \ref{sec:prooflemma1}; third, provide an upper bound for the effective dimension $\widetilde{d}$ in Section \ref{sub:effective}; in the end, present the more experiment details in Section \ref{append:exp}.

\section{Proofs of Theorem \ref{theo1} and \ref{theo2}} \label{sec:profthe1}

\subsection{Proof of Theorem \ref{theo1}}

\begin{proof}
Let $f(\bx; \btheta) = f_1(\bx; \btheta^1) + f_2(\phi(\bx); \btheta^2)$ and we use $\bbe_{\bx_t, y_t}$ to denote $\bbe_{(\bx_t, y_t) \sim \cald}$ for brevity.
For any $t \in [T] \wedge (\mathbf{I}_t = 1)$, we have
\begin{equation}
\begin{aligned}
R_t| (\mathbf{I}_t = 1)  = & \underset{\bx_t, y_t }{\bbe} [  L(\by_{t, \hi}, \by_{t, y_t}) -  L(\by_{t, i^\ast}, \by_{t, y_t})]  \\
= &  \underset{\bx_t, y_t }{\bbe} \left[ 1 -  L(\by_{t, i^\ast}, \by_{t, y_t}) - \left ( 1 -  L(\by_{t, \hi}, \by_{t, y_t}) \right) \right] \\
= &   \underset{\bx_t, y_t }{\bbe}[ r_{t, i^\ast}^1 - r_{t, \hi}^1]\\
  \overset{E_1}{=}   &   \underset{\bx_t, y_t }{\bbe} [ \min\{ r_{t, i^\ast}^1 - r_{t, \hi}^1 , 1\}] \\
= &   \underset{\bx_t, y_t }{\bbe}\left[   \min \{r_{t, i^\ast}^1- f(\bx_{t, i_t}; \btheta_{t-1} )  + f(\bx_{t, i_t}; \btheta_{t-1})   - r_{t, \hi}^1, 1\}   \right] \\
 \overset{E_2}{\leq} & \underset{\bx_t, y_t }{\bbe} \left[  \min \{ r_{t, i^\ast}^1 - f(\bx_{t, i^\ast}; \btheta_{t-1})  + f(\bx_{t, i_t}; \btheta_{t-1})   - r_{t, \hi}^1, 1\}  \right]\\
\overset{E_3}{=} &  \underset{\bx_t, y_t }{\bbe} \Big[ \min \{ r_{t, i^\ast}^1 -  f(\bx_{t, i^\ast}; \btheta_{t-1}^\ast) +  f(\bx_{t, i^\ast}; \btheta_{t-1}^\ast) - f(\bx_{t, i^\ast}; \btheta_{t-1})   \\
 & + f(\bx_{t, i_t}; \btheta_{t-1})   - r_{t, \hi}^1, 1\} \Big] \\
 \leq & \underset{\bx_t, y_t }{\bbe} \left[ \min \{  r_{t, i^\ast}^1 -  f(\bx_{t, i^\ast}; \btheta_{t-1}^\ast), 1\} \right]   +  \underset{\bx_t}{\bbe} [  \min \left \{f(\bx_{t, i^\ast}; \btheta_{t-1}^\ast) - f(\bx_{t, i^\ast}; \btheta_{t-1}), 1 \right\} ]  \\
& +\underset{\bx_t, y_t }{\bbe} \left [ \min \{ f(\bx_{t, i_t}; \btheta_{t-1})   - r_{t, \hi}^1, 1\} \right] \\
 \leq & \underset{\bx_t, y_t }{\bbe} \left[ \min \{ | r_{t, i^\ast}^1 -  f(\bx_{t, i^\ast}; \btheta_{t-1}^\ast)|, 1\} \right]   +    \underset{\bx_t}{\bbe} [ \min \left \{|f(\bx_{t, i^\ast}; \btheta_{t-1}^\ast) - f(\bx_{t, i^\ast}; \btheta_{t-1})|, 1 \right\}  ]\\
& +\underset{\bx_t, y_t }{\bbe} \left [ \min \{ |f(\bx_{t, i_t}; \btheta_{t-1})   - r_{t, \hi}^1|, 1\} \right] \\
\end{aligned}
\end{equation}
where $E_1$ is based on the fact $  r_{t, i} \in [0,1], \forall i \in [k]$, $E_2$ is due to $f(\bx_{t, i^\ast}; \btheta_{t-1}) \leq f(\bx_{t, i_t}; \btheta_{t-1})$ according to our selection criterion, and $\btheta^\ast_{t-1}$ in $E_3$ are intermediate parameters to bound errors.

For any $t \in [T] \wedge  (\mathbf{I}_t = 0)$, we have $R_t| (\mathbf{I}_t = 0)  =  \underset{\bx_t, y_t }{\bbe} [  L(\by_{t, \hi}, \by_{t, y_t}) -  L(\by_{t, i^\ast}, \by_{t, y_t})] = 0$ based on Lemma \ref{lemma:correctlabel}. 

Therefore, for any $t \in [T]$, we have
\begin{equation}
\begin{aligned}
R_t \leq  & \underset{\bx_t, y_t }{\bbe} \left[ \min \{ | r_{t, i^\ast}^1 -  f(\bx_{t, i^\ast}; \btheta_{t-1}^\ast)|, 1\} \right]   +    \underset{\bx_t}{\bbe} [ \min \left \{|f(\bx_{t, i^\ast}; \btheta_{t-1}^\ast) - f(\bx_{t, i^\ast}; \btheta_{t-1})|, 1 \right\}]  \\
& +\underset{\bx_t, y_t }{\bbe} \left [ \min \{ |f(\bx_{t, i_t}; \btheta_{t-1})   - r_{t, \hi}^1|, 1\} \right]
\end{aligned}
\end{equation}

Based on Lemma \ref{lemma:thetabound}, Lemma \ref{lemma:thetaastbound}, and Lemma \ref{lemma:difference}, with probability at least $1 - \delta$, we have
\begin{equation}
R_t \leq 2 \left(  \calo \left(\frac{3L\nu +2 \sqrt{\mu}}{\sqrt{2t}} \right) + \calo \left( \sqrt{ \frac{2 \log ( \calo(k)/\delta) }{t}} \right) +  \xi_1 \right), 
\end{equation}
where $\xi_1 = \calo \left(\frac{\nu L}{\sqrt{m}} \right) +  \mathcal{O} \left(  \frac{ L^2  \sqrt{\log m }  \nu^{4/3}}{  m^{1/6}}\right)$.

Applying the union bound over all the rounds,  with probability at least $1 - \delta$, we have
\begin{equation} \label{eq:regretoft}
\forall t \in [T], \ \ \  R_t \leq  2 \left( \calo \left(\frac{3L\nu+2\sqrt{\mu}}{\sqrt{2t}} \right) + \calo \left( \sqrt{ \frac{2 \log ( \calo(Tk)/\delta) }{t}} ) \right) +  \xi_1. \right)
\end{equation}
When $m$ is large enough, we have $\xi_1  = \calo(\frac{1}{\sqrt{T}})$. Therefore, in round $T$, we have
\begin{equation}
 R_T \leq  \calo \left(\frac{6L\nu + 4\sqrt{\mu})}{\sqrt{2T}} \right) +  \calo\left( \sqrt{ \frac{2 \log ( \calo(Tk)/\delta) }{T}} \right).
\end{equation}

Finally, the regret of $T$ rounds is 
\begin{equation}
\begin{aligned}
\mathbf{R}_T = & \sum_{t=1}^T R_t \\
\leq & \sum_{t=1}^T  2 \left( \underbrace{  \calo \left( \frac{3L\nu + 2\sqrt{\mu}}{\sqrt{2t}} \right) + \sqrt{ \frac{2 \log ( \calo(Tk)/\delta) }{t}}}_{I_1} +  \underbrace{\xi_1}_{I_2} \right) \\
\leq & 2 \left( \underbrace{ (2\sqrt{T} -1) \left[  \calo \left( \frac{3L\nu +2\sqrt{\mu}}{\sqrt{2}} \right) + \sqrt{2 \log ( \calo(Tk)/\delta)} + \underbrace{\calo (1)}_{I_2} \right] }_{I_1}        \right)
\end{aligned}
\end{equation}
where $I_1$ is due to $\sum_{t=1}^T \frac{1}{\sqrt{t}}  \leq  \int_{1}^T   \frac{1}{\sqrt{t}}  \ dx  +1 =  2 \sqrt{T} -1$  and $I_2$ is because of the choice of $m$. 
The proof is complete.
\end{proof}

\subsection{Proof of Theorem \ref{theo2}}
\begin{proof}
Given $\bar{\calt}$,
suppose $T > \bar{\calt}$,  we have
\begin{equation}
\begin{aligned}
\mathbf{R}_T = & \sum_{t=1}^T R_t \\
 = & \sum_{t=1}^{\bar{\calt}} R_t + \sum_{t = \bar{\calt} + 1}^T R_t \\
 =  &  \underbrace{ \sum_{t=1}^{\bar{\calt}}  \underset{\bx_t, y_t }{\bbe} [  L(\by_{t, \hi}, \by_{t, y_t}) -  L(\by_{t, i^\ast}, \by_{t, y_t})]   }_{I_1}  +   \underbrace{  \sum_{t = \bar{\calt} + 1}^T  \underset{\bx_t, y_t }{\bbe} [  L(\by_{t, \hi}, \by_{t, y_t}) -  L(\by_{t, i^\ast}, \by_{t, y_t})]   }_{I_2}\\
\end{aligned}
\end{equation}
For $I_1$, based on Eq.(\ref{eq:regretoft}), for any $\mu \in (0, 1), t \in [\bar{\calt}]$, we have
\begin{equation}
\begin{aligned}
I_1 & \leq  \sum_{t=1}^{\bar{\calt}}  2 \left(\calo \left(\frac{3L\nu +2 \sqrt{\mu}}{\sqrt{2t}} \right) + \sqrt{ \frac{2 \log ( \calo(Tk)/\delta) }{t}} +  \xi_1. \right) \\
& \overset{E_1}{\leq} (2\sqrt{\bar{\calt}} -1) \left[ \calo \left( \frac{6 L\nu + 4 \sqrt{\mu})}{\sqrt{2}} \right) + \sqrt{2 \log ( \calo(Tk)/\delta)}  \right]      
\end{aligned}
\end{equation}
where $E_1$ is because of $\sum_{t=1}^T \frac{1}{\sqrt{t}}  \leq  \int_{1}^T   \frac{1}{\sqrt{t}}  \ dx  +1 =  2 \sqrt{T} -1$  and the choice of $m$.
It is straight forward to show that $R_T$ also satisfies this upper bound when $T \leq \bar{\calt}$.
For $I_2$, based on the Lemma \ref{lemma:queryroundbound},  we have $\bbe_{\bx_t \sim \cald_\calx } [ h(\bx_{t, \hi}) - h(\bx_{t, i^\ast}))] = 0$ when $t \geq \bar{\calt}$. 
This implies
\begin{equation} \label{eq:regreofll}
\bbe_{(\bx_t, y_t) \sim \cald } [  L(\by_{t, i^\ast}, \by_{t, y_t})  - L(\by_{t, \hi}, \by_{t, y_t})] = 0.
\end{equation}
Therefore, we have $I_2 = 0$.
Putting them together, we have
\begin{equation}
\mathbf{R}_T  \leq (2\sqrt{\bar{\calt}} -1) \left[ \calo \left(  \frac{6L\nu + 4\sqrt{\mu}}{\sqrt{2}} \right) + \sqrt{2 \log ( \calo(Tk)/\delta)} \right].
\end{equation}
According to   Eq.(\ref{eq:regretoft}) and Eq.(\ref{eq:regreofll}), we have 
\begin{equation}
\begin{cases}
R_T \leq  \calo \left(\frac{6L\nu + 4\sqrt{\mu} }{\sqrt{2T}} \right) + 2 \sqrt{ \frac{2 \log ( \calo(Tk)/\delta) }{T}}, \ \ \text{if} \ \ T \leq \bar{\calt};   \\
R_T = 0,  \text{else} .
\end{cases}
\end{equation}
Then, replace $\bar{\calt}$ and the proof is complete.
\end{proof}

\subsection{ Main Lemmas}

\begin{lemma} \label{lemma:roundsbound} 
When $t > \bar{\mathcal{T}} =  \frac{12 (\gamma +1)^2 \cdot \left[  2 \mu + 9 L^2 \nu^2 C_1^2 + 2\log(C_2 T k / \delta) \right] }{\epsilon^2}$, it has $2 (\gamma + 1) \bbeta_t \leq \epsilon$.
\end{lemma}

\proof
To achieve $2 (\gamma + 1) \bbeta_t \leq \epsilon$, there exist constants $C_1, C_2$, such that 
\[
\begin{aligned}
\sqrt{ \frac{2 \mu }{t}} +  \left(\frac{3 C_1 L \nu}{\sqrt{2t}} \right) + \sqrt{ \frac{2 \log ( C_2 T k/\delta) }{t}} & \leq \frac{\epsilon}{2 (\gamma +1)} \\
\left( \sqrt{ \frac{2 \mu }{t}} +  \left(\frac{3 C_1 L \nu}{\sqrt{2t}} \right) + \sqrt{ \frac{2 \log ( C_2 T k/\delta) }{t}}\right)^2 & \leq \left(  \frac{\epsilon}{2 (\gamma +1)}\right)^2 \\
3 \left(  \left( \sqrt{ \frac{2 \mu }{t}} \right)^2 +  \left(\frac{3 C_1 L \nu}{\sqrt{2t}} \right)^2 +  \left ( \sqrt{ \frac{2 \log ( C_2 T k/\delta) }{t}}\right)^2   \right)  & \leq \left(  \frac{\epsilon}{2 (\gamma +1)}\right)^2
\end{aligned}
\]
By calculations, we have 
\[
t \geq \frac{12 (\gamma +1)^2 \cdot \left[  2 \mu + 9 L^2 \nu^2 C_1^2 + 2\log(C_2 T k / \delta) \right] }{\epsilon^2}.
\]
The proof is completed.

\qed

\begin{lemma} \label{lemma:queryroundbound}
For any $\delta \in (0, 1)$, $\gamma \geq 1$, suppose $T \geq \bar{\mathcal{T}}$. Then, with probability at least $ 1 - \delta$, these exist constants $C_1, C_2$, such that the following two event $\mathcal{E}_1, \mathcal{E}_2$ happens
\begin{equation}
\mathcal{E}_1 =  \left \{ t  \geq \bar{\calt},   \underset{\bx_t \sim \cald_\calx}{\bbe}[h(\bx_{t, i^\ast}) - h(\bx_{t, \hi})] = 0 \right\},
\end{equation}
\begin{equation}
\mathcal{E}_2 =  \left \{ t  \geq \bar{\calt},   \underset{\bx_t \sim \cald_\calx}{\bbe}[f(\bx_{t, i^\ast}; \btheta_{t-1}) - f(\bx_{t, \hi}; \btheta_{t-1})] = 0 \right\}.
\end{equation}
\end{lemma}

\begin{proof}
According to Lemma \ref{lemma:thetajbound} and Jensen's inequality, for any $i \in [k]$, with probability at least $1 - \delta$,  we have
\begin{equation}
\begin{aligned}
\underset{\bx_t \sim \cald_\calx }{\bbe} \left[ \min \left \{ |f(\bx_{t,i}; \btheta_{t-1}) - h(\bx_{t,i})|, 1 \right \}  \right] 
&  \leq  \underset{(\bx_t, y_t) \sim \cald}{\bbe} \left[  \min \left \{ | f(\bx_{t,i}; \btheta_{t-1}) - r_{t,j}|, 1 \right \} \right]\\
& \leq \sqrt{\frac{2\mu}{t}} + \calo \left(\frac{3L\nu}{\sqrt{2t}} \right) + \sqrt{ \frac{2 \log ( \calo(1)/\delta) }{t}} + 2\xi_1
\end{aligned}
\end{equation}
In round $t$,  define the event
\begin{equation}\label{eq:917}
\widehat{\mathcal{E}}_0 = \left\{ \tau \in [t], i \in [k],     \underset{\bx_\tau \sim \cald_\calx }{\bbe} \left[\min\{ |f(\bx_{\tau,i}; \btheta_{\tau-1}) - h(\bx_{\tau,i})|,  1 \} \right] \leq  \bbeta_{\tau} \right\}
\end{equation}
Then, applying the union bound over $T$ and $k$, then, with probability at least $1 - \delta$, $\mathcal{E}$ happens, where
\begin{equation}
\bbeta_{\tau} = \sqrt{ \frac{2\mu}{\tau}} + \calo \left(\frac{3L\nu}{\sqrt{2\tau}} \right) + \sqrt{ \frac{2 \log ( \calo(T k)/\delta) }{\tau}}.
\end{equation}
where we merge $\xi_1$ into $\calo \left(\frac{3L}{\sqrt{2\tau}} \right)$ as a result of choice of $m$.
Next, define the event
\begin{equation}
\widehat{\mathcal{E}}_1 =  \left \{ t  \geq \bar{\mathcal{T}}, \underset{\bx_t \sim \cald_\calx}{\bbe}[ f(\bx_{t, i^\ast}; \btheta_{t-1}) - f(\bx_{t, i^\circ}; \btheta_{t-1}) ]< 2\gamma \bbeta_{t}  \right\}.
\end{equation}

When $\widehat{\mathcal{E}}_0$ happens with probability at least $1 - \delta$, based on the fact $ h(\cdot) \in [0,1]$,  we have
\begin{equation}  \label{eq:920}
\begin{cases}
 \underset{\bx_t \sim \cald_\calx}{\bbe}[f(\bx_{t, i^\ast}; \btheta_{t-1} )] - \min \{  \bbeta_t, 1 \}  \leq  \underset{\bx_t \sim \cald_\calx}{\bbe}[h(\bx_{t, i^\ast})] \leq  \underset{\bx_t \sim \cald_\calx}{\bbe}[f(\bx_{t, i^\ast}; \btheta_{t-1} )] + \min \{   \bbeta_t, 1\}\\
   \underset{\bx_t \sim \cald_\calx}{\bbe}[f(\bx_{t, i^\circ}; \btheta_{t-1} )] - \min \{ \bbeta_t, 1\}  \leq  \underset{\bx_t \sim \cald_\calx}{\bbe}[h(\bx_{t, i^\circ})] \leq   \underset{\bx_t \sim \cald_\calx}{\bbe} [f(\bx_{t, i^\circ}; \btheta_{t-1} )] +  \min \{  \bbeta_t, 1 \}
\end{cases}
\end{equation}
Then, based on Lemma \ref{lemma:roundsbound}, with probability at least $1-\delta$, when $t > \bar{\calt}$, $2(\gamma+1) \bbeta_t \leq \epsilon \Rightarrow  \bbeta_t < 1$. This implies

\begin{equation} \label{eq:event1}
\begin{cases}
 \underset{\bx_t \sim \cald_\calx}{\bbe}[f(\bx_{t, i^\ast}; \btheta_{t-1} )] -   \bbeta_t  \leq  \underset{\bx_t \sim \cald_\calx}{\bbe}[h(\bx_{t, i^\ast})] \leq  \underset{\bx_t \sim \cald_\calx}{\bbe}[f(\bx_{t, i^\ast}; \btheta_{t-1} )] +   \bbeta_t \\
   \underset{\bx_t \sim \cald_\calx}{\bbe}[f(\bx_{t, i^\circ}; \btheta_{t-1} )] - \bbeta_t  \leq  \underset{\bx_t \sim \cald_\calx}{\bbe}[h(\bx_{t, i^\circ})] \leq   \underset{\bx_t \sim \cald_\calx}{\bbe} [f(\bx_{t, i^\circ}; \btheta_{t-1} )] +    \bbeta_t 
\end{cases}
\end{equation}
Therefore, we have
\begin{equation}
\begin{aligned}
 \underset{\bx_t \sim \cald_\calx}{\bbe}[h(\bx_{t, i^\ast}) -  h(\bx_{t, i^\circ})] & \leq  \underset{\bx_t \sim \cald_\calx}{\bbe}[f(\bx_{t, i^\ast}; \btheta_{t-1} )] + \bbeta_t - \left(  \underset{\bx_t \sim \cald_\calx}{\bbe}[ f(\bx_{t, i^\circ}; \btheta_{t-1} ) ] - \bbeta_t \right)\\
&\leq   \underset{\bx_t \sim \cald_\calx}{\bbe}[f(\bx_{t, i^\ast}; \btheta_{t-1} ) - f(\bx_{t, i^\circ}; \btheta_{t-1} )]  + 2 \bbeta_t.
\end{aligned}
\end{equation}
Suppose $\widehat{\mathcal{E}}_1$ happens, we have
\begin{equation}
 \underset{\bx_t \sim \cald_\calx}{\bbe}[h(\bx_{t, i^\ast}) -  h(\bx_{t, i^\circ})]  \leq  2(\gamma+1) \bbeta_t. 
\end{equation}
Then, based on Lemma \ref{lemma:roundsbound}, when $t > \bar{\calt}$, $2(\gamma+1) \bbeta_t \leq \epsilon$. Therefore, we have
\begin{equation}
 \underset{\bx_t \sim \cald_\calx}{\bbe}[h(\bx_{t, i^\ast}) -  h(\bx_{t, i^\circ})]  \leq 2(\gamma+1) \bbeta_t \leq \epsilon.
\end{equation}
This contradicts Assumption \ref{assum:margin}, i.e.,  $h(\bx_{t, i^\ast}) -  h(\bx_{t, i^\circ})\geq \epsilon$. Hence, $\widehat{\mathcal{E}}_1 $ will not happen. Accordingly, with probability at least $1 -\delta$, the following event will happen
\begin{equation}
\widehat{\mathcal{E}}_2 =  \left \{ t  \geq \bar{\calt},  \underset{\bx_t \sim \cald_\calx}{\bbe}[ f(\bx_{t, i^\ast}; \btheta_{t-1}) - f(\bx_{t, i^\circ}; \btheta_{t-1})] \geq 2\gamma \bbeta_{t}  \right\}. 
\end{equation}
Therefore, we have $\bbe[f(\bx_{t, i^\ast}; \btheta_{t-1})]  > \bbe[f(\bx_{t, i^\circ}; \btheta_{t-1})]$.
Recall that $i^\ast = \arg \max_{i \in [k]} h(\bx_{t, i})$ and $\hi  = \arg \max_{i \in [k]} f(\bx_{t, i}; \btheta_{t-1})$.
As 
\begin{equation}
\begin{aligned}
\forall i \in ( [k] \setminus \{\hi \} ), f(\bx_{t,i}; \btheta_{t-1})   \leq  f(\bx_{t, i^\circ}; \btheta_{t-1}) \\
\Rightarrow \forall  i \in ([k] \setminus \{ \hi \}), \underset{\bx_t \sim \cald_\calx}{\bbe}[f(\bx_{t,i}; \btheta_{t-1}) ]  \leq  \underset{\bx_t \sim \cald_\calx}{\bbe}[ f(\bx_{t, i^\circ}; \btheta_{t-1})]
\end{aligned}
\end{equation}
 we have
 \begin{equation}
 \forall i \in ([k] \setminus \{ \hi \}),  \underset{\bx_t \sim \cald_\calx}{\bbe}[f(\bx_{t, i^\ast}; \btheta_{t-1})]  > \underset{\bx_t \sim \cald_\calx}{\bbe} [f(\bx_{t,i}; \btheta_{t-1})].
 \end{equation}
Based on the definition of $\hi$, we have 
\begin{equation}
\begin{aligned}
& \underset{\bx_t \sim \cald_\calx}{\bbe} [f (\bx_{t, i^\ast}; \btheta_{t-1}) ] = \underset{\bx_t \sim \cald_\calx}{\bbe} [f(\bx_{t, \hi}; \btheta_{t-1})] = \underset{\bx_t \sim \cald_\calx}{\bbe} [ \underset{ i\in [k] }{\max} f(\bx_{t,i}; \btheta_{t-1})]. 
\end{aligned}
\end{equation}
This indicates $\mathcal{E}_2$ happens with probability at least $1 -\delta$.

Therefore, based on $\widehat{\mathcal{E}}_2$, the following inferred event $\widehat{\mathcal{E}}_3$ happens with probability at least $1 -\delta$:
\begin{equation}
\widehat{\mathcal{E}}_3 =  \left \{ t  \geq \bar{\calt},  \underset{\bx_t \sim \cald_\calx}{\bbe}[ f(\bx_{t, \hi}; \btheta_{t-1}) - f(\bx_{t, i^\circ}; \btheta_{t-1})] \geq 2\gamma \bbeta_{t}  \right\}. 
\end{equation}
Then, based on Eq. \ref{eq:event1}, we have
 \begin{equation} \label{eq:hlowerbound}
 \begin{aligned}
 \bbe[h(\bx_{t, \hi}) -  h(\bx_{t, i^\circ})] &  \geq \bbe[f(\bx_{t, \hi}; \btheta_{t-1} )] - \bbeta_t  - \left(  \bbe[f(\bx_{t, i^\circ}; \btheta_{t-1} )] + \bbeta_t \right) \\
 & =  \bbe[f(\bx_{t, \hi}; \btheta_{t-1} ) -  f(\bx_{t, i^\circ}; \btheta_{t-1})] - 2 \bbeta_t\\
 & \overset{E_1}{\geq} 2 (\gamma -1)\bbeta_t \\
 & \geq 0 
 \end{aligned}
 \end{equation}
 where $E_1$ is because $\widehat{\mathcal{E}}_3$ happened with probability at least $1-\delta$. Therefore, we have 
 \begin{equation}
 \underset{\bx_t \sim \cald_\calx}{\bbe}[h(\bx_{t, \hi})] - \underset{\bx_{t} \sim \cald_\calx }{\bbe}[h(\bx_{t, i^\circ})] > 0.
 \end{equation}
 Similarly, we can prove that
 \begin{equation}
 \Rightarrow \forall i \in ([k] \setminus \{\hi \}),  \underset{\bx_t \sim \cald_\calx}{\bbe}[h(\bx_{t, \hi})] - \underset{\bx_{t} \sim \cald_\calx }{\bbe}[h(\bx_{t, i})] > 0.
 \end{equation}
 Then, based on the definition of $\bx_{t, i^\ast}$, we have 
 \begin{equation}
  \underset{\bx_t \sim \cald_\calx}{\bbe}[h(\bx_{t, \hi})] =  \underset{\bx_t \sim \cald_\calx}{\bbe}[h(\bx_{t, i^\ast})] =   \underset{\bx_t \sim \cald_\calx}{\bbe}[\max_{i \in [k]} h(\bx_{t,i})].
 \end{equation}
Thus, the event $\mathcal{E}_1$ happens with probability at least $1 - \delta$.
\end{proof}

\begin{lemma} \label{lemma:thetajbound}
For any $\delta \in (0, 1), \nu > 0 $, suppose $m$ satisfies the conditions in Theorem \ref{theo1}. Then,
with probability at least $1-\delta$, given any fixed index $i \in [k]$,  it holds that 
\begin{equation}
\begin{aligned}
 &\underset{(\bx_t, y_t) \sim \cald }{\bbe} \left[   \min \left \{ \left| f_1(\bx_{t,i}; \btheta^{1}_{t-1}) + f_2(\phi(\bx_{t, i}); \btheta^2_{t-1})   - r^1_{t,i}  \right|, 1  \right \} \right] \\
 \leq & \sqrt{\frac{2 \mu}{t}} + \calo \left(\frac{3L\nu}{\sqrt{2t}} \right) + \sqrt{ \frac{2 \log ( \calo(1)/\delta) }{t}} + 2\xi_1.
\end{aligned}
\end{equation}

\end{lemma}

\proof

Given any $i \in [k]$, let $i$ be a fixed index, i.e., suppose there exist a policy $\Omega_i$ which always select the $i$-th context $(\bx_{t,i}, r^1_{t,i})$ for every round $t \in [T]$. Then, in round $t$, we have the collected data by $\Omega_i$: $\mathcal{H}_{t-1}^{1,i} = \{ \bx_{\tau,i}, r^1_{\tau,i}\}_{\tau= 1}^{t-1}$. Then, let $ \btheta^{1, i}_{t - 1}, \btheta^{2, i}_{t - 1} $ represent the parameters trained only on $\mathcal{H}_{t-1}^i$ using Algorithm \ref{alg:BGD}, satisfying $ \|\btheta^{1, i}_{t - 1} - \btheta^1_0 \|_2 \leq  \calo(\frac{\nu}{\sqrt{m}})  $ and   $ \|\btheta^{2, i}_{t - 1} - \btheta^2_0 \|_2 \leq  \calo(\frac{\nu}{\sqrt{m}})  $ . Note that $ \btheta^{1, i}_{t - 1}, \btheta^{2, i}_{t - 1}$ are  uniformly drawn from $\{  \hbtheta^{1, i}_{\tau - 1},  \hbtheta^{2, i}_{\tau - 1}\}_{\tau =0}^{t-1}$ and these parameters are unknown but introduced for the sake of analysis.
Then, for $\tau \in [t]$,  we define
\begin{equation}
\begin{aligned}
V_{\tau} = & \underset{ (\bx_\tau, y_\tau) \sim \cald }{\bbe} \left[ \min \{  |f_1(\bx_{\tau, i}; \hbtheta^{1, i}_{\tau - 1}) + f_2(\phi(\bx_{\tau, i}); \hbtheta^{2, i}_{\tau - 1}) - r^1_{\tau, i}|, 1 \} \right]  \\
& - \min \{ |   f_1(\bx_{\tau, i}; \hbtheta^{1, i}_{\tau - 1}) + f_2(\phi(\bx_{\tau, i}); \hbtheta^{2, i}_{\tau - 1}) - r^1_{\tau, i} |, 1 \}
\end{aligned}
\end{equation}
Then, we have
\begin{equation}
\begin{aligned}
\bbe[V_{\tau}| F_{\tau - 1}]  =&  \underset{ (\bx_\tau, y_\tau) \sim \cald }{\bbe} \left[ \min \{  |f_1(\bx_{\tau, i}; \hbtheta^{1, i}_{\tau - 1}) + f_2(\phi(\bx_{\tau, i}); \hbtheta^{2, i}_{\tau - 1}) - r^1_{\tau, i}|, 1 \} \right]  \\
& -   \underset{ (\bx_\tau, y_\tau) \sim \cald }{\bbe} \left[ \min \{  |f_1(\bx_{\tau, i}; \hbtheta^{1, i}_{\tau - 1}) + f_2(\phi(\bx_{\tau, i}); \hbtheta^{2, i}_{\tau - 1}) - r^1_{\tau, i}|, 1 \} \right]  \\
= & 0
\end{aligned}
\end{equation}
where $F_{\tau - 1}$ denotes the $\sigma$-algebra generated by the history $\mathcal{H}_{\tau -1}^i$. Therefore, $\{V_{\tau}\}_{\tau =1}^t$ are the martingale difference sequence.

Then, using the similar proof method in Lemma \ref{lemma:thetabound}, we have
\begin{equation}\label{eq:940}
\begin{aligned}
  & \underset{ (\bx_t, y_t) \sim \cald }{\bbe} \left[   \min \left \{ \left| f_1(\bx_{t,i}; \btheta^{1, i}_{t-1}) + f_2(\phi(\bx_{t, i}); \btheta^{2, i}_{t-1})   - r_{t,i}^1 \right|, 1  \right \} \mid  \mathcal{H}_{t-1}^i \right] \\
\leq & \sqrt{\frac{2\mu}{t}} + \calo \left(\frac{3L\nu}{\sqrt{2t}} \right) + \sqrt{ \frac{2 \log ( \calo(1)/\delta) }{t}}.
\end{aligned}
\end{equation}
Let $f(\bx; \btheta_{t-1}) =  f_1(\bx; \btheta^{1}_{t-1}) + f_2(\phi(\bx); \btheta^{2}_{t-1})$ and $f(\bx; \btheta_{t-1}^i) =  f_1(\bx; \btheta^{1, i}_{t-1}) + f_2(\phi(\bx); \btheta^{2, i}_{t-1})$.

Then, given $i \in [k]$, we have
\begin{equation}
\begin{aligned}
& \underset{ (\bx_t, y_t) \sim \cald }{\bbe} \left[  \min \left \{ |f(\bx_{t,i}; \btheta_{t-1}) - r_{t, i}^1 |, 1\right\}  \right] \\
= & \underset{ (\bx_t, y_t) \sim \cald }{\bbe} \left[ \min \left \{  |f(\bx_{t,i}; \btheta_{t-1}) - f(\bx_{t,i}; \btheta_{t-1}^i) + f(\bx_{t,i}; \btheta_{t-1}^i) - r^1_{t,i}|, 1 \right\}  \right] \\
\leq & \underset{ (\bx_t, y_t) \sim \cald }{\bbe}[ |f(\bx_{t,i}; \btheta_{t-1}) - f(\bx_{t,i}; \btheta_{t-1}^i)| ] + \underset{ (\bx_t, y_t) \sim \cald }{\bbe} [ \min \left \{  | f(\bx_{t,i}; \btheta_{t-1}^i) - r^1_{t,i}|, 1 \right\}]\\
\leq & \sqrt{\frac{2 \mu}{t}} + \calo \left(\frac{3L\nu}{\sqrt{2t}} \right) + \sqrt{ \frac{2 \log ( \calo(1)/\delta) }{t}}+ 2\xi_1
\end{aligned} 
\end{equation}
where the last inequality is the application of Lemma \ref{lemma:difference} and Eq. (\ref{eq:940}). The proof is complete.

\begin{lemma}[Label Complexity Analysis]
For any $\delta \in (0, 1), \gamma \geq 1$, suppose $m$ satisfies the conditions in Theorem \ref{theo1}. Then,
with probability at least $1-\delta$, we have
\begin{equation}
\mathbf{N}_T \leq  \frac{12 (\gamma +1)^2 \cdot \left[  2 \mu + 9 L^2 \nu^2 C_1^2 + 2\log(C_2 T k / \delta) \right] }{\epsilon^2}.
\end{equation}

\end{lemma}

\begin{proof}
Recall that $\bx_{t, \hi} = \max_{\bx_{t,i}, i \in [k]} f(\bx_{t,i}; \btheta_{t-1})$, and $\bx_{t, i^\circ} = \max_{\bx_{t,i}, i \in ([k] /\{ \bx_{t, \hi} \}) } f(\bx_{t,i}; \btheta_{t-1})$.
With probability at least $1-\delta$, according to Eq. (\ref{eq:917}) the event 
\[
\widehat{\mathcal{E}}_0 = \left\{ \tau \in [t], i \in [k],     \underset{\bx_\tau \sim \cald_\calx }{\bbe} \left[\min\{ |f(\bx_{\tau,i}; \btheta_{\tau-1}) - h(\bx_{\tau,i})|,  1 \} \right] \leq  \bbeta_{\tau} \right\}
\]
happens. Therefore, we have
\begin{equation}
\begin{cases}
 \underset{\bx_t \sim \cald_\calx }{\bbe}[h(\bx_{t, \hi})] - \min\{\bbeta_t, 1\}  \leq    \underset{\bx_t \sim \cald_\calx }{\bbe}[f(\bx_{t, \hi}; \btheta_{t-1} )]
 \leq \underset{\bx_t \sim \cald_\calx }{\bbe}[h(\bx_{t, \hi})] +  \min\{\bbeta_t, 1\}\\
 \underset{\bx_t \sim \cald_\calx }{\bbe}[h(\bx_{t, i^\circ})] -  \min\{\bbeta_t, 1\}  \leq  \underset{\bx_t \sim \cald_\calx }{\bbe}[f(\bx_{t, i^\circ}; \btheta_{t-1} )]  \leq  \underset{\bx_t \sim \cald_\calx }{\bbe}[h(\bx_{t, i^\circ})] +  \min\{\bbeta_t, 1\}.
\end{cases}
\end{equation}
Then, we have 
\begin{equation} \label{eq:froundbound}
\begin{cases}
 \underset{\bx_t \sim \cald_\calx }{\bbe}[ f(\bx_{t, \hi}; \btheta_{t-1} ) - f(\bx_{t, i^\circ}; \btheta_{t-1} )   ] \leq   \underset{\bx_t \sim \cald_\calx }{\bbe}[h(\bx_{t, \hi})]  -  \underset{\bx_t \sim \cald_\calx }{\bbe}[h(\bx_{t, i^\circ})] + 2  \min\{\bbeta_t, 1\} \\
 \underset{\bx_t \sim \cald_\calx }{\bbe}[ f(\bx_{t, \hi}; \btheta_{t-1} ) - f(\bx_{t, i^\circ}; \btheta_{t-1} )   ] \geq  \underset{\bx_t \sim \cald_\calx }{\bbe}[h(\bx_{t, \hi})]  -  \underset{\bx_t \sim \cald_\calx }{\bbe}[h(\bx_{t, i^\circ})] - 2  \min\{\bbeta_t, 1\}.
\end{cases}
\end{equation}
Let $\epsilon_t =  | \underset{\bx_t \sim \cald_\calx }{\bbe}[h(\bx_{t, \hi})]  -  \underset{\bx_t \sim \cald_\calx }{\bbe}[h(\bx_{t, i^\circ})]|$. Then, based on Lemma \ref{lemma:roundsbound}, when $t \geq \bar{\calt}$, we have
\begin{equation} \label{eq:945}
  2(\gamma +1) \bbeta_t \leq \epsilon_t \leq 1.
\end{equation}
For any $t \in [T]$ and $t < \bar{\calt}$, we have $ \underset{\bx_t \sim \cald_\calx }{\bbe}[\mathbf{I}_t] \leq 1$.
For the round $t > \bar{\calt}$, suppose $   \underset{\bx_t \sim \cald_\calx }{\bbe}[h(\bx_{t, \hi})]  -  \underset{\bx_t \sim \cald_\calx }{\bbe}[h(\bx_{t, i^\circ})] = - \epsilon_t$, then, we have
\begin{equation}
 \underset{\bx_t \sim \cald_\calx }{\bbe}[ f(\bx_{t, \hi}; \btheta_{t-1} ) - f(\bx_{t, i^\circ}; \btheta_{t-1} )   ] \leq  -\epsilon_t  + 2\bbeta_t \overset{E_1}{\leq} -\epsilon_t + \frac{\epsilon_t}{2}  \leq 0,
\end{equation}
where $E_1$ is because of Eq. (\ref{eq:945}) since $\gamma \geq 1$.
This contradicts the fact $ \underset{\bx_t \sim \cald_\calx }{\bbe}[ f(\bx_{t, \hi}; \btheta_{t-1} ) - f(\bx_{t, i^\circ}; \btheta_{t-1} )   ]  \geq 0$. Therefore, $ \underset{\bx_t \sim \cald_\calx }{\bbe}[h(\bx_{t, \hi})]  -  \underset{\bx_t \sim \cald_\calx }{\bbe}[h(\bx_{t, i^\circ})] = \epsilon_t$.
Then, based on Eq.(\ref{eq:froundbound}), we have 
\begin{equation} \label{eq:948}
 \underset{\bx_t \sim \cald_\calx }{\bbe}[  f(\bx_{t, \hi}; \btheta_{t-1} ) - f(\bx_{t, i^\circ}; \btheta_{t-1} ) ] \geq \epsilon_t - 2\bbeta_t  \overset{E_2}{\geq} 2 \gamma \bbeta_t,
\end{equation}
where $E_2$ is because of Eq. (\ref{eq:945}).

According to Lemma \ref{lemma:queryroundbound}, when $ t > \bar{\calt} $, $\underset{\bx_t \sim \cald_\calx}{\bbe}[f(\bx_{t, i^\ast}; \btheta_{t-1})] =  \underset{\bx_t \sim \cald_\calx}{\bbe}[f(\bx_{t, \hi}; \btheta_{t-1})]$. Then, applying Eq. (\ref{eq:948}), for the round $t > \bar{\calt}$, we have $ \underset{\bx_t \sim \cald_\calx }{\bbe}[\mathbf{I}_t] = 0$.

Then, assume $T> \bar{\calt}$, we have
\begin{equation}
\begin{aligned}
\mathbf{N}_T &= \sum_{t= 1}^T  \underset{\bx_t \sim \cald_\calx }{\bbe}  \left [      \mathbbm{1}\{  f(\bx_{t, \hi}; \btheta_{t-1} ) - f(\bx_{t, i^\circ}; \btheta_{t-1} ) < 2 \gamma \bbeta_t  \}   \right]\\
&\leq  \sum_{t= 1}^{\bar{\calt}} 1 + \sum_{t = \bar{\calt} +1}^T  \underset{\bx_t \sim \cald_\calx }{\bbe}  \left [      \mathbbm{1}\{  f(\bx_{t, \hi}; \btheta_{t-1} ) - f(\bx_{t, i^\circ}; \btheta_{t-1} ) < 2 \gamma \bbeta_t  \}   \right] \\
&=  \bar{\calt} + 0.
\end{aligned}
\end{equation}
Therefore, we have $\mathbf{N}_T \leq \bar{\calt}$.
\end{proof}

\begin{lemma} \label{lemma:correctlabel}
For any $\delta \in (0, 1), \gamma \geq 1$, suppose $m$ satisfies the conditions in Theorem \ref{theo1}. Then,
with probability at least $1-\delta$, when $\mathbf{I}_t = 0$,  we have
\[
\begin{aligned}
\underset{\bx_t \sim \cald_\calx }{\bbe}[h(\bx_{t, \hi})] &= \underset{\bx_t \sim \cald_\calx }{\bbe}[h(\bx_{t, i^\ast})],\\
 \underset{(\bx_t, y_t) \sim \cald }{\bbe}[L(\by_{t, \hi}, \by_{t,y_t})] & = \underset{(\bx_t, y_t) \sim \cald }{\bbe}[L(\by_{t, i^\ast}, \by_{t,y_t})] .
\end{aligned}
\]
\end{lemma}

\begin{proof}
As $\mathbf{I}_t = 0$, we have
\[
|f(\bx_{t, \hi}; \btheta_{t-1} ) -  f(\bx_{t, i^\circ}; \btheta_{t-1})| = f(\bx_{t, \hi}; \btheta_{t-1} ) -  f(\bx_{t, i^\circ}; \btheta_{t-1})  \geq 2 \gamma \bbeta_t
\]

When $\widehat{\mathcal{E}}_0$ (Eq. (\ref{eq:917})) happens with probability at least $1 - \delta$, based on the fact $ h(\cdot) \in [0,1]$,  we have
\begin{equation}  \label{eq:920}
\begin{cases}
 \underset{\bx_t \sim \cald_\calx}{\bbe}[f(\bx_{t, \hi}; \btheta_{t-1} )] - \min \{  \bbeta_t, 1 \}  \leq  \underset{\bx_t \sim \cald_\calx}{\bbe}[h(\bx_{t, \hi})] \leq  \underset{\bx_t \sim \cald_\calx}{\bbe}[f(\bx_{t, \hi}; \btheta_{t-1} )] + \min \{   \bbeta_t, 1\}\\
   \underset{\bx_t \sim \cald_\calx}{\bbe}[f(\bx_{t, i^\circ}; \btheta_{t-1} )] - \min \{ \bbeta_t, 1\}  \leq  \underset{\bx_t \sim \cald_\calx}{\bbe}[h(\bx_{t, i^\circ})] \leq   \underset{\bx_t \sim \cald_\calx}{\bbe} [f(\bx_{t, i^\circ}; \btheta_{t-1} )] +  \min \{  \bbeta_t, 1 \}
\end{cases}
\end{equation}

Then, with probability at least $1-\delta$,we have
\begin{equation}
\begin{aligned}
 \underset{\bx_t \sim \cald_\calx}{\bbe}[h(\bx_{t, \hi}) -  h(\bx_{t, i^\circ})] &  \geq  \underset{\bx_t \sim \cald_\calx}{\bbe}[f(\bx_{t, \hi}; \btheta_{t-1} ) -  f(\bx_{t, i^\circ}; \btheta_{t-1})] - 2 \min \{ \bbeta_t, 1 \}\\
 & \geq 2 \gamma \bbeta_t   - 2 \min \{ \bbeta_t, 1 \}  \\
 & \geq 0  
 \end{aligned}
\end{equation}
where the last inequality is because of $\gamma \geq 1$.
Then, similarly, for any $ i' \in  ([k] \setminus \{\hi, i^\circ \})$,  we have $ \underset{\bx_t \sim \cald_\calx}{\bbe}[h(\bx_{t, \hi}) - h(\bx_{t, i'})] \geq 0$. Thus, based on the definition of $h(\bx_{t, i^\ast})$, we have $ \underset{\bx_t \sim \cald_\calx}{\bbe}[h(\bx_{t, \hi})] =  \underset{\bx_t \sim \cald_\calx}{\bbe}[h(\bx_{t, i^\ast})]$. 
Because $  \underset{\bx_t \sim \cald_\calx }{\bbe}[h(\bx_{t, \hi})] =     \underset{(\bx_t, y_t) \sim \cald }{\bbe}[ 1 -  L(\by_{t, \hi}, \by_{t,y_t})]$, we have
\[
\begin{aligned}
 \underset{(\bx_t, y_t) \sim \cald }{\bbe}[ 1 -  L(\by_{t, \hi}, \by_{t,y_t})] =  \underset{(\bx_t, y_t) \sim \cald }{\bbe}[ 1 -  L(\by_{t, i^\ast}, \by_{t,y_t})] \\
 \Rightarrow   \underset{(\bx_t, y_t) \sim \cald }{\bbe}[L(\by_{t, \hi}, \by_{t,y_t})] = \underset{(\bx_t, y_t) \sim \cald }{\bbe}[L(\by_{t, i^\ast}, \by_{t,y_t})] .
\end{aligned}
\]
The proof is complete.
\end{proof}

\begin{lemma} \label{lemma:thetabound}
For any $\delta \in (0, 1), \nu > 0$, suppose $m$ satisfies the conditions in Theorem \ref{theo1}.
In round $t \in [T]$, given $(\bx_t, y_t) \sim \cald$, let
\[\hi = \arg \max_{i \in [k]} \left(   f_1(\bx_{t,\hi}; \btheta^1_{t-1}) + f_2(\phi(\bx_{t,\hi}); \btheta^2_{t-1})   \right).\]
Then, with probability at least $1-\delta$,  we have
\begin{equation}
\begin{aligned}
\underset{(\bx_{t}, y_t) \sim \cald}{\bbe} \left[   \min \left \{ \left| f_1(\bx_{t,\hi}; \btheta^1_{t-1}) + f_2(\phi(\bx_{t,\hi}); \btheta^2_{t-1})   - r^1_{t, \hi} \right|, 1  \right \} |  \calh^1_{t-1} \right] \\
\leq \calo \left(\frac{3L\nu + 2\sqrt{\mu}}{\sqrt{2t}} \right) +  2\sqrt{ \frac{2 \log ( \calo(1)/\delta) }{t}}, 
\end{aligned}
\end{equation}
where $\calh_{t-1}^1 = \{\bx_{\tau, \hi}, r^1_{\tau, \hi}\}_{\tau = 1}^{t-1}$ is historical data and the expectation is taken over $(\btheta^1_{t-1}$, $\btheta^2_{t-1})$.
\end{lemma}

\begin{proof}
This lemma is inspired by Lemma 5.1 in \cite{ban2021ee}. For any round $\tau \in [t]$, define
\begin{equation}
\begin{aligned}
V_{\tau} =& \underset{(\bx_{\tau}, y_\tau) \sim \cald }{\bbe} \left[ \min \{  |f_1(\bx_{\tau, \hi}; \hbtheta^1_{\tau - 1}) + f_2( \phi(\bx_{\tau, \hi}); \hbtheta^2_{\tau - 1}) - r^1_{\tau, \hi}|, 1 \} \right]  \\
&- \min \{ |   f_1(\bx_{\tau, \hi}; \hbtheta^1_{\tau - 1}) + f_2(\phi(\bx_{\tau, \hi}); \hbtheta^2_{\tau - 1}) - r_{\tau, \hi}^1 |, 1 \}
\end{aligned}
\end{equation}
Then, we have
\begin{equation}
\begin{aligned}
\bbe[V_{\tau}| F_{\tau - 1}]  =&\underset{(\bx_{\tau}, y_\tau) \sim \cald }{\bbe} \left[ \min \{ |f_1(\bx_{\tau, \hi}; \hbtheta^1_{\tau - 1}) + f_2( \phi(\bx_{\tau, \hi}); \hbtheta^2_{\tau - 1}) - r^1_{\tau, \hi}|, 1 \} \right] \\
& - \underset{(\bx_{\tau}, y_\tau) \sim \cald }{\bbe} \left[ \min \{   |f_1(\bx_{\tau, \hi}; \hbtheta^1_{\tau - 1}) + f_2( \phi(\bx_{\tau, \hi}); \hbtheta^2_{\tau - 1}) - r^1_{\tau, \hi}|, 1 \} \right] \\
= & 0
\end{aligned}
\end{equation}
where $F_{\tau - 1}$ denotes the $\sigma$-algebra generated by the history $\mathcal{H}^1_{\tau -1}$. Therefore, $\{V_{\tau}\}_{\tau =1}^t$ are the martingale difference sequence.

Then, applying the Hoeffding-Azuma inequality, with probability at least $1-\delta$, we have
\begin{equation}
    \bbp \left[  \frac{1}{t}  \sum_{\tau=1}^t  V_{\tau}  -   \underbrace{ \frac{1}{t} \sum_{\tau=1}^t  \bbe[ V_{\tau} | \mathbf{F}_{\tau-1} ] }_{I_1}   >  \sqrt{ \frac{2 \log (1/\delta)}{t}}  \right] \leq \delta   \\
\end{equation}    
As $I_1$ is equal to $0$, we have
\begin{equation}
\begin{aligned}
      &\frac{1}{t} \sum_{\tau=1}^t \underset{(\bx_{\tau}, y_\tau) \sim \cald }{\bbe} \left[ \min \{ |f_1(\bx_{\tau, \hi}; \hbtheta^1_{\tau - 1}) + f_2( \phi(\bx_{\tau, \hi}); \hbtheta^2_{\tau - 1}) - r^1_{\tau, \hi}|, 1 \} \right] \\
 \leq & \frac{1}{t}\sum_{\tau=1}^t \min \{ |f_1(\bx_{\tau, \hi}; \hbtheta^1_{\tau - 1}) + f_2( \phi(\bx_{\tau, \hi}); \hbtheta^2_{\tau - 1}) - r^1_{\tau, \hi}|, 1 \} +   \sqrt{ \frac{2 \log (1/\delta) }{t}}    \\
 \leq & \frac{1}{t}\sum_{\tau=1}^t \left|f_2 \left( \bx_{\tau, \hi} ; \hbtheta_{\tau-1}^2 \right) - \left(r^1_{\tau, \hi} - f_1(\bx_{\tau, \hi}; \hbtheta_{\tau-1}^1) \right)  \right|   +   \sqrt{ \frac{2 \log (1/\delta) }{t}}   ~.
    \end{aligned}  
\label{eq:pppuper}
\end{equation}

Based on the the definition of $\btheta_{t-1}^1, \btheta^2_{t-1}$, we have
\begin{equation}
\begin{aligned}
& \underset{  ( \bx_{t}, y_t) \sim \cald}{\bbe}  \underset{(\btheta^1, \btheta^2)}{\bbe} \left[ \min \{ |f_1(\bx_{t,\hi}; \btheta^1_{t - 1}) + f_2(\bx_{t,\hi}; \btheta^2_{t - 1}) - r^1_{t,\hi}|, 1\} \right]  \\
= & \frac{1}{t} \sum_{\tau=1}^t  \underset{  ( \bx_{\tau}, y_\tau) \sim \cald}{\bbe}  \left[  \min \{ \left |f_1(\bx_{\tau, \hi}; \hbtheta^1_{\tau - 1}) + f_2(\bx_{\tau,\hi}; \hbtheta^2_{\tau - 1}) - r^1_{\tau, \hi} \right|, 1 \}  \right].
\end{aligned}
\end{equation}
Therefore, putting them together, we have
\begin{equation}\label{eq9.57}
\begin{aligned}
 &  \underset{  ( \bx_{t}, y_t) \sim \cald}{\bbe}  \underset{(\btheta^1, \btheta^2)}{\bbe} \left[ |f_1(\bx_{t,\hi}; \btheta^1_{t - 1}) + f_2(\bx_{t,\hi}; \btheta^2_{t - 1}) - r^1_{t, \hi}| \right] \\
 \leq  & \underbrace{\frac{1}{t}\sum_{\tau=1}^t \left|f_2 \left( \bx_{\tau, \hi} ; \hbtheta_{\tau-1}^2 \right) - \left(r^1_{\tau, \hi} - f_1(\bx_{\tau, \hi}; \hbtheta_{\tau-1}^1) \right)  \right|}_{I_2}   +   \sqrt{ \frac{2 \log (1/\delta) }{t}}  ~.
\end{aligned}
\end{equation}

For $I_2$, based on Lemma \ref{lemma:caogeneli}, we have
\begin{equation} \label{eq9.58}
\begin{aligned}
I_2 &\leq \frac{1}{t}\sum_{\tau=1}^t \left|f_2 \left( \bx_{\tau, \hi} ; \widetilde{\btheta}^2 \right) - \left(r^1_{\tau, \hi} - f_1(\bx_{\tau, \hi}; \hbtheta_{\tau-1}^1) \right)  \right| + \calo \left(\frac{3L \nu}{\sqrt{2t}} \right) +  \sqrt{ \frac{2 \log (1/\delta) }{t}} \\
& \leq \frac{1}{t}\sqrt{t}\sqrt{
\underbrace{
\sum_{\tau =1}^t 
\left(      
f_2 \left( \bx_{\tau, \hi} ; \widetilde{\btheta}^2 \right) - \left(r^1_{\tau, \hi} - f_1(\bx_{\tau, \hi}; \hbtheta_{\tau-1}^1) \right) 
\right)^2}_{I_3} }+ \calo \left(\frac{3L \nu}{\sqrt{2t}} \right) +  \sqrt{ \frac{2 \log (1/\delta) }{t}
} \\
& \leq \sqrt{ \frac{2\mu}{t}} + \calo \left(\frac{3L \nu}{\sqrt{2t}} \right) +  \sqrt{ \frac{2 \log (1/\delta) }{t}}
\end{aligned}
\end{equation}
where $I_3$  is based on the assumption of $\mu$.

Combining above Eq. (\ref{eq9.57}) and (\ref{eq9.58})  together,  with probability at least $1 - \delta$, we have 
\begin{equation}
\begin{aligned}
\underset{  ( \bx_{t}, y_t) \sim \cald}{\bbe}  \left[   \min \left \{ \left| f_1(\bx_{t,\hi}; \btheta^1_{t-1}) + f_2(\phi(\bx_{t,\hi}); \btheta^2_{t-1})   - r^1_{t,\hi} \right|, 1  \right \} \right] \\
\leq   \calo \left(\frac{3L\nu + 2\sqrt{\mu}}{\sqrt{2t}} \right) +  2 \sqrt{ \frac{2 \log ( \calo(1)/\delta) }{t}}.
\end{aligned}
\end{equation}
where we apply union bound over $\delta$ to make above events occur concurrently.

Then, based on Lemma \ref{lemma:zhu} (2), it is sufficient to show that $\btheta^1_{t-1}, \btheta^2_{t-1} $ are close to initialization for any $t \in [T]$.
The proof is complete.
\end{proof}

\begin{lemma} \label{lemma:thetaastbound}

In round $t \in [T]$, given $(\bx_t, y_t) \sim \cald$, let  $i^\ast = \arg \max_{ i\in[k]} h(\bx_{t,i})$. 
Let $\btheta^{1,\ast}_{t-1}, \btheta^{2, \ast}_{t-1}$ are the parameters trained on $ \calh_{t-1}^\ast$ using Algorithm \ref{alg:BGD}.
For any $\nu > 0$, suppose 
 $   \underset{ \widetilde{\btheta}^{2, \ast} \in \calb(\theta^2_0, \nu) }{\inf} \frac{1}{2} \sum_{\tau =1}^t 
\left(      
f_1(\bx_{\tau, i^\ast}; \widehat{\btheta}_{\tau-1}^{1, \ast})
+ f_2 \left( \phi(\bx_{\tau, i^\ast}) ; \widetilde{\btheta}^{2, \ast} \right) - r^1_{\tau, i^\ast}  
\right)^2 \leq \mu$.
Then, with probability at least $1 - \delta$,  we have
\begin{equation}
\begin{aligned}
 \underset{  ( \bx_{t}, y_t) \sim \cald}{\bbe} \left[   \min \left \{ \left| f_1(\bx_{t, i^\ast}; \btheta^{1,\ast}_{t-1}) + f_2(\phi(\bx_{t, i^\ast}); \btheta^{2, \ast}_{t-1})   - r^1_{t, i^\ast} \right|, 1  \right \} |  \calh_{t-1}^\ast \right] \\
\leq   \calo \left(\frac{3L\nu + 2\sqrt{\mu} }{\sqrt{2t}} \right) +  2\sqrt{ \frac{2 \log ( \calo(1)/\delta) }{t}}, 
\end{aligned}
\end{equation}
where $\calh_{t-1}^\ast = \{\bx_{\tau, i^\ast}, r^1_{\tau, i^\ast} \}_{\tau = 1}^{t-1}$ is optimal data of past rounds the expectation is taken over $\btheta^{1, \ast}_{t-1}$, $\btheta^{2, \ast}_{t-1}$.
\end{lemma}

\begin{proof}
This lemma is a direct corollary of Lemma \ref{lemma:thetabound}. For any $\tau \in [t]$,
define
\begin{equation}
\begin{aligned}
V_{\tau} = &  \underset{  ( \bx_{\tau}, y_\tau) \sim \cald}{\bbe} \left[ \min \{  |f_1(\bx_{\tau, i^\ast}; \hbtheta^{1,\ast}_{\tau - 1}) + f_2(\phi(\bx_{\tau, i^\ast}); \hbtheta^{2, \ast}_{\tau - 1}) - r^1_{\tau, i^\ast}|, 1 \} \right] \\
& - \min \{ |   f_1(\bx_{\tau, i^\ast}; \hbtheta^{1,\ast}_{\tau - 1}) + f_2(\phi( \bx_{\tau, i^\ast}); \hbtheta^{2, \ast}_{\tau - 1}) - r^1_{\tau, i^\ast} |, 1 \}
\end{aligned}
\end{equation}
Then, we have
\begin{equation}
\begin{aligned}
\bbe[V_{\tau}| F_{\tau - 1}]  = &  \underset{  ( \bx_{\tau}, y_\tau) \sim \cald}{\bbe} \left[ \min \{ |f_1(\bx_{\tau, i^\ast}; \hbtheta^{1, \ast}_{\tau - 1}) + f_2(\phi(\bx_{\tau, i^\ast}); \hbtheta^{2, \ast}_{\tau - 1}) - r^1_{\tau, i^\ast}|, 1 \} \right] \\
& -   \underset{  ( \bx_{\tau}, y_\tau) \sim \cald}{\bbe} \left[ \min \{  |   f_1(\bx_{\tau, i^\ast}; \hbtheta^{1, \ast}_{\tau - 1}) + f_2(\phi(\bx_{\tau, i^\ast}); \hbtheta^{2, \ast}_{\tau - 1}) - r^1_{\tau, i^\ast} |, 1 \} \right] \\
 =&  0
\end{aligned}
\end{equation}
where $F_{\tau - 1}$ denotes the $\sigma$-algebra generated by the history $\mathcal{H}_{\tau -1}$. Therefore, $\{V_{\tau}\}_{\tau =1}^t$ are the martingale difference sequence.

Then, applying the Hoeffding-Azuma inequality, with probability at least $1-\delta$, we have
\begin{equation}
    \bbp \left[  \frac{1}{t}  \sum_{\tau=1}^t  V_{\tau}  -   \underbrace{ \frac{1}{t} \sum_{\tau=1}^t  \bbe[ V_{\tau} | \mathbf{F}_{\tau} ] }_{I_1}   >  \sqrt{ \frac{2 \log (1/\delta)}{t}}  \right] \leq \delta   \\
\end{equation}    
As $I_1$ is equal to $0$, we have
\begin{equation}
\begin{aligned}
&\underset{  ( \bx_{t}, y_t) \sim \cald}{\bbe}  \underset{(\btheta^1, \btheta^2)}{\bbe}  \left[ \min \{ |f_1(\bx_{t, i^\ast}; \btheta^1_{t - 1}) + f_2(\phi(\bx_{t, i^\ast}); \btheta^2_{t - 1}) - r^1_{t, i^\ast}|, 1\} \right]  \\
     = &\frac{1}{t} \sum_{\tau=1}^t \underset{  ( \bx_{\tau}, y_\tau) \sim \cald}{\bbe} \left[ \min \{ \left |f_1(\bx_{\tau, i^\ast}; \hbtheta^{1, \ast}_{\tau - 1}) + f_2( \phi(\bx_{\tau, i^\ast}); \hbtheta^{2,\ast}_{\tau - 1}) - r^1_{\tau, i^\ast}, 1\} \right|  \right]  \\
 \leq  & \underbrace{\frac{1}{t}\sum_{\tau=1}^t \left|f_2 \left( \phi(\bx_{\tau, i^\ast}); \widehat{\btheta}_{\tau-1}^{2, \ast} \right) - \left(r^1_{\tau, i^\ast} - f_1(\bx_{\tau, i^\ast}; \widehat{\btheta}_{\tau-1}^{1, \ast}) \right)  \right|}_{I_2}   +   \sqrt{ \frac{2 \log (1/\delta) }{t}}.
    \end{aligned}  
\label{eq:pppuper}
\end{equation}

For $I_2$, applying Lemma \ref{lemma:caogeneli}, for any $\widetilde{\btheta}^{2, \ast}$ satisfying $\| \widetilde{\btheta}^{2, \ast}  - \btheta^2_0 \|_2 \leq \calo(\frac{\nu }{\sqrt{m}}  )$, with probability at least $1 -3\delta$, we have
\begin{equation}
\begin{aligned}
I_2 &\leq \frac{1}{t}  \sum^t_{\tau = 1 } |f_1(\bx_{\tau, i^\ast}; \hbtheta_{\tau-1}^{1, \ast}) + f_2 \left( \phi(\bx_{\tau, i^\ast}) ; \widetilde{\btheta}^{2, \ast} \right) - r^1_{\tau, i^\ast}| + \calo \left(\frac{3L \nu}{\sqrt{2t}} \right) +  \sqrt{ \frac {2 \log (1/\delta)}{t} } \\
& \leq \frac{1}{t}\sqrt{t}\sqrt{
\underbrace{
\sum_{\tau =1}^t 
\left(      
f_1(\bx_{\tau, i^\ast}; \widehat{\btheta}_{\tau-1}^{1, \ast})
+ f_2 \left( \phi(\bx_{\tau, i^\ast}) ; \widetilde{\btheta}^{2, \ast} \right) - r^1_{\tau, i^\ast}  
\right)^2}_{I_3}} + \calo \left(\frac{3L\nu}{\sqrt{2t}} \right) +  \sqrt{ \frac {2 \log (1/\delta)}{t} } 
 \\
& \leq \sqrt{\frac{2 \mu}{t}} + \calo \left(\frac{3L \nu}{\sqrt{2t}} \right) +  \sqrt{ \frac {2 \log (1/\delta)}{t} } 
\end{aligned}
\end{equation}
where $I_3$ is because of the assumption of $\mu$.

Combining above inequalities together, as $\mu \in (0, 1]$, with probability at least $1 - \delta$, we have 
\begin{equation}
\begin{aligned}
 \underset{  ( \bx_{t}, y_t) \sim \cald}{\bbe} \left[   \min \left \{ \left| f_1(\bx_{t, i^\ast}; \btheta^{1, \ast}_{t-1}) + f_2(\phi(\bx_{t, i^\ast}); \btheta^{2, \ast}_{t-1})   - r_{t, i^\ast}^1 \right|, 1  \right \} \right] \\
\leq \calo \left(\frac{3L \nu + 2 \sqrt{\mu}}{\sqrt{2t}} \right) + 2\sqrt{ \frac{2 \log ( \calo(1)/\delta) }{t}}, 
\end{aligned}
\end{equation}
where we apply union bound over $\delta$ to make above events occur concurrently.
Then, based on Lemma \ref{lemma:zhu} (2), it is sufficient to show that $\btheta^{1,\ast}_{t-1}, \btheta^{2, \ast}_{t-1} $ are close to initialization for any $t \in [T]$.
\end{proof}

\begin{lemma} \label{lemma:caogeneli}
For any $\delta \in (0, 1)$, suppose $m$ satisfies the condition in Theorem \ref{theo1}.
Then, with probability at least $1 - \delta$, setting $\eta_2 =  \frac{ \kappa \nu}{ m \sqrt{t} }$ for algorithm \ref{alg:main},   for $\nu > 0$ and any $\widetilde{\btheta}^2$ satisfying $\|   \widetilde{\btheta}^2  - \btheta^2_0    \|_2 \leq  \mathcal{O}(\frac{\nu}{\sqrt{m}})$,  and $i \in [k]$, there exists a small enough constant $\kappa$, such that
\[
\begin{aligned}
\sum_{\tau=1}^t \left|   f_2 \left( \phi(\bx_{\tau, i})  ; \widehat{\btheta}_{\tau-1}^2  \right) - \left(r^1_{\tau, i} - f_1(\bx_{\tau, i}; \widehat{\btheta}_{\tau-1}^1) \right) \right| \\
\leq   \sum_{\tau=1}^t \left|   f_2  \left( \phi( \bx_{\tau,i}) ; \widetilde{\btheta}^2 \right) - \left(r^1_{\tau, i} - f_1(\bx_{\tau, i}; \widehat{\btheta}_{\tau-1}^1) \right) \right| + \calo \left(\frac{3L\nu\sqrt{t} }{\sqrt{2 }}\right) +  \sqrt{ 2 t \log (1/\delta)}. \\
\end{aligned}
\]
\end{lemma}
\begin{proof}
This is a direct application of Lemma \ref{lemma:newbsgd} by setting $\hat{\epsilon} = \frac{L\nu}{ \sqrt{2\kappa t}}$,  and, where $\kappa$ is some small enough absolute constant.
We set $L_\tau( \widehat{\btheta}_{\tau -1}^2 ) =  \left |  f_2 ( \phi(\bx_{\tau, i})  ; \widehat{\btheta}_{\tau-1}^2 ) - \left(r^1_{\tau, i} - f_1(\bx_{\tau, i}; \widehat{\btheta}_{\tau-1}^1) \right) \right| $.
 Then,  for any $\widetilde{\btheta}^2$ satisfying $\|   \widetilde{\btheta}^2  - \btheta^2_0    \|_2 \leq  \mathcal{O}(\frac{\nu}{\sqrt{m} })$, there exist a small enough absolute constant $\kappa$, such that 
\begin{equation}
\sum_{\tau=1}^t  L_\tau( \widehat{\btheta}_{\tau -1}^2 ) \leq \sum_{\tau=1}^t L_\tau( \widetilde{\btheta}^2) + \calo(3 t \hat{\epsilon}) +  \sqrt{ 2 t \log (1/\delta)}.
\end{equation}
Then, replacing $\hat{\epsilon}$ completes the proof.
\end{proof}

\begin{lemma} \label{lemma:newbsgd}

With probability at least $1 -\delta$ over the randomness of $\btheta_0$, given the convex loss $L$ satisfying $L' \leq \calo(1)$, for any $\hat{\epsilon}, \nu>0$ and $\widetilde{\btheta}$ satisfying $ \|\hbtheta - \btheta_0\|_2 \leq \calo(\frac{\nu}{\sqrt{m}})$, Algorithm \ref{alg:main} with $\eta = \frac{\kappa \hat{\epsilon}}{Lm}$ and $t = \frac{L^2 \nu^2}{2 \kappa \hat{\epsilon}^2}$ for some small enough constant $\kappa$ has the following bound:
\begin{equation}
\sum_{\tau = 1}^t \min\{ L_{(\bx_\tau, r_\tau)}(\hbtheta_{\tau-1}) -  L_{(\bx_\tau, r)}(\widetilde{\btheta}), 1 \} \leq \calo(3 t \hat{\epsilon})  +    \sqrt{ 2 t \log (1/\delta)}.
\end{equation}
\end{lemma}

\newcommand{\hbh}{\widehat{\mathcal{H}}}

\begin{proof}
Define $\mathcal{B}(\btheta_0, \nu) = \{\btheta \in \bbr^p: \|\btheta - \btheta_0\|_2 \leq \calo(\nu/\sqrt{m}) \}$ and   $ L_{(\bx; r)}(\btheta) =  | r - f(\bx; \btheta)|$.
First, we need to show $\hbtheta_1, \dots, \hbtheta_t$ also are in $\mathcal{B}(\btheta_0, w)$, where. 
According to Lemma \ref{lemma:difference}, when  $\btheta \in \mathcal{B}(\btheta_0, w)$, we have
\begin{equation}
\begin{aligned}
\|\tri_{\btheta} f(\bx; \btheta)\|_2 &\leq \calo(L), \ \
\|\tri_{\btheta} L_{(\bx; r)}(\btheta)\|_2 & \leq  \sqrt{\sum_{l=1}^L \| \calo (  \tri_{\bw_l} f(\bx; \btheta) )\|_2^2} \leq \calo(L).
\end{aligned}
\end{equation}
The proof follows a simple induction. Suppose that $\btheta_0, \hbtheta_1, \dots, \hbtheta_t \in \mathcal{B}(\btheta_0, w)$, by triangle inequality, we have
\begin{equation}
\|\hbtheta_t -\btheta_0\|_2  \leq \sum_{\tau = 0}^{t-1} \|\hbtheta_{\tau+1} - \hbtheta_{\tau}\|_2 \leq  \frac{1}{| \hbh_t |} \sum_{\tau = 0}^{t-1} \sum_{(\bx, r) \in \hbh_t} \|\tri_{\btheta} L(\bx; r) \|_2 \leq \calo( L \eta t).
\end{equation}
Because $\eta = \calo(\frac{1}{m})$, we have $ \|\hbtheta_t -\btheta_0\|_2 \leq  \calo(\nu/\sqrt{m})$.
In round $\tau \in [t]$, recall that $|\hbh_\tau| = b$ and $\hbh_\tau \subset \calh_\tau $.
Given the context $\bx$ and its reward $r$,  we have the fact
\begin{equation} \label{eq:expectl}
\begin{aligned}
\tau \underset{\hbh_\tau}{\bbe} \left[ \frac{1}{b} \sum_{(\bx, r) \in \hbh_\tau} \tri_{\hbtheta} L_{(\bx, r)}(\hbtheta^{t-1}) \right]
& \overset{E_1}{=} \tau \underset{(\bx, r) \sim \calh_\tau}{\bbe}  \left[     \tri_{\hbtheta} L_{(\bx, r)}(\hbtheta^{t-1})\right]\\ 
&  = \tau \sum_{ (\bx, r) \in \calh_\tau} \frac{1}{|\calh_\tau|}    \tri_{\hbtheta} L_{(\bx, r)}(\hbtheta^{t-1}) \\
& \overset{E_2}{=}    \sum_{ (\bx, r) \in \calh_\tau}   \tri_{\hbtheta} L_{(\bx, r)}(\hbtheta^{t-1}),
\end{aligned}
\end{equation}
where $E_1$ is because $\hbh_\tau$ is uniformly drawn from $\calh_\tau$ and $E_2$ is duo to $|\calh_\tau| = \tau$. Then, based on Lemma \ref{lemma:gu2}, for any $\epsilon > 0$, we have
\begin{equation}
\begin{aligned}
\underset{(\bx, r) \sim \calh_\tau}{\bbe}[L_{(\bx, r)}(\hbtheta_{\tau-1}) -  L_{(\bx, r)}(\widetilde{\btheta})] & \leq \langle \underset{ (\bx, r) \sim \calh_\tau}{\bbe}[ \tri_{\hbtheta} L_\tau(\hbtheta_{\tau-1})], \hbtheta_{\tau-1} - \widetilde{\btheta} \rangle + \hat{\epsilon} \\
    & \leq \left \langle  \underset{\hbh_\tau}{\bbe} \left[ \frac{1}{b} \sum_{(\bx, r) \in \hbh_\tau} \tri_{\hbtheta} L_{(\bx, r)}(\hbtheta^{t-1}) \right], \hbtheta_{\tau-1} - \widetilde{\btheta}  \right\rangle +  \hat{\epsilon} \\
& =  \frac{  \langle \hbtheta_{\tau-1} -  \underset{\hbh_\tau}{\bbe}[  \hbtheta_{\tau}],    \hbtheta_{\tau-1} - \widetilde{\btheta} \rangle  }{ \eta}      +  \hat{\epsilon}.
\end{aligned}
\end{equation}
Based on the fact $2\langle \mathbf{A}, \mathbf{B} \rangle = \|\mathbf{A} \|_2^2  +  \|\mathbf{B} \|_2^2 - \|\mathbf{A}, \mathbf{B}\|_2^2$, we have
\begin{equation}
\begin{aligned}
\underset{(\bx, r) \sim \calh_\tau}{\bbe}[L_{(\bx, r)}(\hbtheta_{\tau-1}) -  L_{(\bx, r)}(\widetilde{\btheta})]
& \leq  \frac{ \|   \hbtheta_{\tau-1} -  \underset{\hbh_\tau}{\bbe}[  \hbtheta_{\tau}]  \|_2^2  + \|\hbtheta_{\tau-1} - \widetilde{\btheta} \|_2^2 - \| \underset{\hbh_\tau}{\bbe}[  \hbtheta_{\tau}] - \widetilde{\btheta}   \|_2^2  }{ 2 \eta }  + \hat{\epsilon} \\
& \overset{E_3}{\leq}  \frac{\|\hbtheta_{\tau-1} - \widetilde{\btheta} \|_2^2 - \| \underset{\hbh_\tau}{\bbe}[  \hbtheta_{\tau}] - \widetilde{\btheta}   \|_2^2     }{2\eta} + \calo(L^2\eta) + \hat{\epsilon} 
\end{aligned}
\end{equation}
where $E_3$ is because of \ref{eq:expectl}:
\begin{equation}
\begin{aligned}
\|   \hbtheta_{\tau-1} -  \underset{\hbh_\tau}{\bbe}[  \hbtheta_{\tau}]  \|_2 &=  \eta  \left  \|  \underset{\hbh_\tau}{\bbe} \left[ \frac{1}{b} \sum_{(\bx, r) \in \hbh_\tau} \tri_{\hbtheta} L_{(\bx, r)}(\hbtheta^{t-1}) \right] \right\|_2 \\
&= \eta  \frac{1}{\tau}\|  \sum_{ (\bx, r) \in \calh_\tau}   \tri_{\hbtheta} L_{(\bx, r)}(\hbtheta^{t-1}) \|_2 \leq \calo(\eta L).
\end{aligned}
\end{equation}
Therefore, we have
\begin{equation} \label{eq:elupper}
\begin{aligned}
\sum_{\tau =1}^t  \underset{(\bx, r) \sim \calh_\tau}{\bbe}[L_{(\bx, r)}(\hbtheta_{\tau-1}) -  L_{(\bx, r)}(\widetilde{\btheta})]
& \leq   \frac{\|\hbtheta_{0} - \widetilde{\btheta} \|_2^2 - \| \underset{\hbh_\tau}{\bbe}[  \hbtheta_{t}] - \widetilde{\btheta}   \|_2^2     }{2\eta} + \calo(tL^2\eta) + t \hat{\epsilon} \\
& \leq  \frac{\|\hbtheta_{0} - \widetilde{\btheta} \|_2^2  }{2\eta} + \calo(tL^2\eta) + t \hat{\epsilon}\\
& \leq \frac{LR^2}{2\eta m} + \calo(tL^2\eta) + t \hat{\epsilon}.
 \end{aligned} 
\end{equation}
Then, for $\tau \in [t]$, define
\begin{equation}
V_\tau =\min\{ L_{(\bx, r)}(\hbtheta_{\tau-1}) -  L_{(\bx, r)}(\widetilde{\btheta}), 1 \}  - \underset{(\bx, r) \sim \calh_\tau}{\bbe}[ \min\{ L_{(\bx, r)}(\hbtheta_{\tau-1}) -  L_{(\bx, r)}(\widetilde{\btheta}) , 1 \}].  
\end{equation}
Then, we have
\begin{equation}
\begin{aligned}
\bbe [V_\tau |\mathcal{F}_{\tau -1}] = & \underset{(\bx, r) \sim \calh_\tau}{\bbe} [ \min\{ L_{(\bx, r)}(\hbtheta_{\tau-1}),  -  L_{(\bx, r)}(\widetilde{\btheta}),  1 \}]  \\
& -  \underset{(\bx, r) \sim \calh_\tau}{\bbe}[ \min\{ L_{(\bx, r)}(\hbtheta_{\tau-1}) -  L_{(\bx, r)}(\widetilde{\btheta}), 1 \}] = 0,
\end{aligned}
\end{equation}
where  where $F_{\tau - 1}$ denotes the $\sigma$-algebra generated by the history $\mathcal{H}_{\tau -1}$.
Therefore, $\{V_0 , \dots, V_t\}$ is the  martingale  difference sequence. Then,  applying the Hoeffding-Azuma inequality, with probability at least $1- \delta$, we have
\begin{equation}
\begin{aligned}
& \sum_{\tau = 1}^t \min\{ L_{(\bx, r)}(\hbtheta_{\tau-1}) -  L_{(\bx_\tau, r)}(\widetilde{\btheta}), 1 \} \\
 \leq &  \sum_{\tau = 1}^t  \underset{(\bx, r) \sim \calh_\tau}{\bbe}[ \min\{ L_{(\bx, r)}(\hbtheta_{\tau-1}) -  L_{(\bx, r)}(\widetilde{\btheta}) , 1 \}] +  t  \sqrt{ \frac{2 \log (1/\delta) }{t}} \\
 \overset{E_4}{\leq }  & \frac{LR^2}{2\eta m} + \calo(tL^2\eta) + t \hat{\epsilon} +  \sqrt{ 2 t \log (1/\delta)} \\
  \overset{E_5}{\leq} & \calo(3 t \hat{\epsilon})  +    \sqrt{ 2 t \log (1/\delta)} 
\end{aligned}
\end{equation}
where $ E_4$ be because of \ref{eq:elupper} and $E_5$ is by placing the parameter choice $\eta = \frac{\kappa \hat{\epsilon}}{Lm}$ and $t = \frac{L^2 \nu^2}{2 \kappa \hat{\epsilon}}$. The proof is completed.
\end{proof}

\subsection{Ancillary Lemmas}

\begin{lemma} [Theorem 5, \cite{allen2019convergence}] \label{allenzhu:t5}
For any $\delta \in (0, 1)$, if $w$ satisfies that
\begin{equation}
\calo ( m^{-3/2}L^{-3/2} \max\{\log^{-3/2}m, \log^{3/2}(Tn/\delta) \}) \leq w \leq  \calo( L^{-9/2} \log^{-3}m ),
\end{equation}
then, with probability at least $1 - \delta$, for all $\|\btheta - \btheta_0\|_2 \leq w$, we have
\begin{equation}
\|\tri_{\btheta} f(\bx; \btheta) - \tri_{\btheta_0}    f(\bx; \btheta_0)   \|_2 \leq \calo(\sqrt{\log m} w^{1/3} L^3) \|\tri_{\btheta_0}f(\bx; \btheta_0)\|_2.
\end{equation}
\end{lemma}

\begin{lemma} [Lemma 4.1, \citep{cao2019generalization}]  \label{lemma:gu} 
For any $\delta \in (0, 1)$, if $w$ satisfies 
\[
\mathcal{O}( m^{-3/2} L^{-3/2} [ \log (tnL^2/\delta) ]^{3/2}  )   \leq w \leq  \mathcal{O} (L^{-6}[\log m]^{-3/2} ),
\]
then,
with probability at least $1- \delta$ over randomness of $\btheta_0$, for any $ t \in [T], \|\bx \|_2 =1 $, and $\btheta, \btheta'$
satisfying $\| \btheta - \btheta_0 \|_2 \leq w$ and $\| \btheta' - \btheta_0 \|_2 \leq w$
, it holds uniformly that
\begin{equation}
| f(\bx_i; \btheta) - f(\bx_i; \btheta') -  \langle  \triangledown_{\btheta'}f(\bx_i; \btheta'), \btheta - \btheta'    \rangle    | \leq \mathcal{O} (w^{1/3}L^2 \sqrt{m \log(m)}) \|\btheta - \btheta'\|_2.
\end{equation}
\end{lemma}

\begin{lemma} [Lemma 4.2, \citep{cao2019generalization}]  \label{lemma:gu2} 
For any $\delta \in (0, 1), \hat{\epsilon} >0$, if $w$ satisfies 
\[
\mathcal{O}( m^{-3/2} L^{-3/2} [ \log (tnL^2/\delta) ]^{3/2}  )   \leq w \leq  \kappa  L^{-6} m^{-3/8} [\log m]^{-3/2} \hat{\epsilon}^{3/4},
\]
then,
with probability at least $1- \delta$ over randomness of $\btheta^{(0)}$, for any $\hat{\epsilon} >0, i \in [n]$, and $\btheta, \widetilde{\btheta}$
satisfying $\| \btheta - \btheta^{(0)} \|_2 \leq w$ and $\| \widetilde{\btheta} - \btheta^{(0)} \|_2 \leq w$
, it holds uniformly that
\begin{equation}
L_{(\bx, r)}(\hbtheta_{\tau-1}) -   L_{(\bx, r)}(\widetilde{\btheta})  \leq
\langle \tri_{\hbtheta} L_{\bx, r}(\hbtheta_{\tau-1}), \hbtheta_{\tau-1} - \widetilde{\btheta} \rangle + \hat{\epsilon} 
\end{equation}
\end{lemma}

\begin{lemma} \label{lemma:zhu}
Given a constant $0 < \hat{\epsilon} < 1$, suppose $m$ satisfies the conditions in Lemma \ref{lemma:1}, the learning rate $\eta = \Omega(\frac{\rho}{ \text{poly} (t ,n, L) m} ) $, the number of iterations $K = \Omega (\frac{\text{poly}(t, n, L)}{\rho^2} \cdot \log \hat{\epsilon}^{-1})$. Then, with probability at least $1 - \delta$,  starting from random initialization $\btheta_0$, 
\begin{enumerate}[label=(\arabic*)]
    \item (Theorem 1 in \citep{allen2019convergence}) 
    In round $t \in [T]$, given the collected data $\{\bx_\tau, r_\tau\}_{i=\tau}^{t}$, the loss function is defined as:
    $\mathcal{L}(\btheta) = \frac{1}{2} \sum_{\tau = 1}^{t}   \left( f(\bx_{\tau}; \btheta) - r_\tau \right)^2$.
    Then,  there exists  $\widetilde{\btheta}$  satisfying $\|\widetilde{\btheta} - \btheta_0\|_2 \leq \mathcal{O}\left( \frac{ t^3 }{\rho \sqrt{m}} \log m  \right)$, such that
    $ \mathcal{L}(\widetilde{\btheta}) \leq \hat{\epsilon} $ in $K = \Omega (\frac{\text{poly}(t, n, L)}{\rho^2} \cdot \log \hat{\epsilon}^{-1})$ iterations;\\
    \item For any $t \in [T]$, it holds uniformly that
    $
    \|\btheta_{t-1} - \btheta_0\|_2 \leq \mathcal{O}\left( \frac{ t^3 }{\rho \sqrt{m}} \log m  \right)
    $;
    \item (Lemma C.4 in \citep{ban2021ee}) Following the initialization, given $\|\bx\|_2 =1$, it holds that
    \[
     \|   \triangledown_{\btheta_0 }f(\bx; \btheta_0)  \|_2 \leq \mathcal{O} (L),   \    \   \
      |  f(\bx; \btheta_0)  | \leq \mathcal{O} (1).
    \]
\end{enumerate}
\end{lemma}

\begin{proof}
(2) is a corollary of Theorem 1 in \citep{allen2019convergence}.
Suppose $\btheta_{\tau} =  \btheta_{\tau-1} - \frac{\eta}{b} \sum_{(\bx, r) \in \widehat{\calh}_{\tau}} \tri_{\btheta}  \call[ (\bx, r); \btheta_{\tau-1}]$. The proof is based on the following induction. Let $w = \calo \left( \frac{t^3}{\delta \sqrt{m}} \log m \right)$. Then, based on the Theorem 1 in \citep{allen2019convergence}, we have 
\[
 \call[ (\bx, r); \btheta_{\tau}] = \left(1 - \Omega \left( \frac{\eta \rho m}{ t^2} \right)\right)   \call[ (\bx, r); \btheta_{\tau-1}].
\]
Then, we have
\[
\begin{aligned}
\|   \btheta_{t} -   \btheta_{t-1} \|_2 & \leq \sum_{\tau =1} \|    \frac{\eta}{b} \sum_{(\bx, r) \in \widehat{\calh}_{\tau}} \tri_{\btheta}  \call[ (\bx, r); \btheta_{\tau-1}] \| \leq 
\calo(\eta \sqrt{t m})   \sum_{\tau =1}^t \sqrt{   \call[ (\bx, r); \btheta_{\tau-1}] } \\
& \leq  \calo(\eta \sqrt{t m}) \cdot \Omega( \frac{t^2}{\eta \rho m} ) \cdot \calo (  \sqrt{t \log^2 m}     ) \leq \calo \left( \frac{t^3}{\rho \sqrt{m}} \log m \right). 
\end{aligned}
\]
\end{proof}

\begin{lemma}[Lemma C.2 \cite{ban2021ee}] \label{lemma:difference}
For any $\delta \in (0, 1), \nu > 0$, suppose $m$ satisfies the conditions in Theorem \ref{theo1}. Then, with probability at least $1 - \delta$, in each round $t \in [T]$, for any $\bx$ satisfying $\|\bx\|_2 = 1$, $ \btheta^{1, \ast}_{t-1}, \btheta^{1}_{t-1}$ satisfying $ \|   \btheta^{1, \ast}_{t-1} - \btheta^{1}_{t-1} \|_2 \leq \calo \left( \frac{\nu}{\sqrt{m}} \right)$,  and $ \btheta^{2, \ast}_{t-1}, \btheta^{2}_{t-1}$ satisfying $ \|   \btheta^{2, \ast}_{t-1} - \btheta^{2}_{t-1} \|_2 \leq \calo \left( \frac{\nu}{\sqrt{m}} \right)$,
we have 
\begin{equation}
\begin{aligned}
 (1) \qquad  & |f_1(\bx; \btheta^{1, \ast}_{t-1}) - f_1(\bx; \btheta^1_{t-1})| \\
 \leq &  \calo \left(\frac{\nu L}{\sqrt{m}} \right) +  \mathcal{O} \left(  \frac{ L^2  \sqrt{\log m }  \nu^{4/3}}{  m^{1/6}}\right) := \xi_1;
 \end{aligned}
\end{equation}
\begin{equation}
\begin{aligned}
(2) \qquad & \left| f_2 \left( \phi(\bx)  ; \btheta_{t-1}^{2, \ast} \right) - f_2 \left( \phi(\bx); \btheta_{t-1}^{2} \right) \right|  \\
\leq &  \left(\frac{\nu L}{\sqrt{m}} \right) +  \mathcal{O} \left(  \frac{ L^2  \sqrt{\log m }  \nu^{4/3}}{  m^{1/6}}\right);
\end{aligned}
\end{equation}
\begin{equation}
\begin{aligned}
(3)  \qquad \qquad & \|\triangledown_{\btheta^1_{t-1}}f_1(\bx; \btheta^1_{t-1})\|_2,  \|\triangledown_{\btheta^2_{t-1}}f_2 \left( \phi(\bx)  ; \btheta_{t-1}^{2} \right) \|_2 \\
\leq & \calo(L)~.
\end{aligned}
\end{equation}
\end{lemma}

\section{Proof of Lemma \ref{lemma:1}} \label{sec:prooflemma1}

\begin{algorithm}[h]
\renewcommand{\algorithmicrequire}{\textbf{Input:}}
\renewcommand{\algorithmicensure}{\textbf{Output:}}
\caption{ Batch-GD-Warm-Start ( $f_1$, $f_2$, $\, \mathcal{H}_t^1, \mathcal{H}_t^2$) }\label{alg:BGD2}
\begin{algorithmic}[1] 
\State Define $\call_1[(\bx, r^1); \btheta^1] =  (r^1 - f_1(\bx; \btheta^1))^2/2$
\State Uniformly draw a set $\widehat{\mathcal{H}}^1_t  \subset \mathcal{H}_t^1, s.t.,  |\widehat{\mathcal{H}}^1_t| = b$
\State   $\hbtheta^1_t = \hbtheta^1_{t-1} -  \frac{\eta_1}{b} \underset{(\bx, r^1)\in  \widehat{\mathcal{H}}^1_t}{\sum}  \tri_{\btheta^1}  \call_1 [(\bx, r^1); \hbtheta_{t-1}^1]$
\State  Define $\call_2[(\phi(\bx), r^2); \btheta^2] =  (r^2 - f_2(\phi(\bx); \btheta^2))^2/2$
\State Uniformly draw a set $\widehat{\mathcal{H}}^2_t  \subset \mathcal{H}_t^2, s.t., |\widehat{\mathcal{H}}^2_t| = b$
\State $\hbtheta^2_t = \hbtheta^2_{t-1} -  \frac{\eta_2}{b} \underset{(\phi(\bx), r^1)\in  \widehat{\mathcal{H}}^2_t}{\sum}  \tri_{\btheta^2}  \call_2 [(\phi(\bx), r^2); \hbtheta_{t-1}^2]$
\State \textbf{Return} $(\hbtheta^1_t, \hbtheta^2_t)$
\end{algorithmic}
\end{algorithm}

\begin{proof}
For any $t \in [T] \wedge (\mathbf{I}_t = 1)$, the regret of one round can be bounded as:
\begin{equation}
\begin{aligned} 
& R_t | (\mathbf{I}_t = 1) \\
= & \underset{ y_t \sim \cald_{\caly|\bx_t}}{\bbe}[\call(\by_{t, \hi}, \by_{t, y_t}) | \bx_t ]    -  \underset{ y_t \sim \cald_{\caly|\bx_t}}{\bbe}[\call(\by_{t, i^\ast}, \by_{t, y_t}) | \bx_t ] \\
 = & 1 -  \underset{ y_t \sim \cald_{\caly|\bx_t}}{\bbe}[\call(\by_{t, i^\ast}, \by_{t, y_t}) | \bx_t ]  - ( 1 -  \underset{ y_t \sim \cald_{\caly|\bx_t}}{\bbe}[\call(\by_{t, \hi}, \by_{t, y_t}) | \bx_t ] ) \\
 = & \underset{y_t \sim \cald_{\caly|\bx_t} }{\bbe} [ r_{t, i^\ast}^1 -  r_{t, \hi}^1] \\
 = & \underset{y_t \sim \cald_{\caly|\bx_t} }{\bbe} [\min\{r_{t, i^\ast}^1 -  r_{t, \hi}^1, 1\} ]\\
 = & \underset{y_t \sim \cald_{\caly|\bx_t} }{\bbe} [\min\{r_{t, i^\ast}^1   - f(\bx_{t, \hi};  \btheta_{t-1})  + f(\bx_{t, \hi};  \btheta_{t-1})   -  r_{t, \hi}^1, 1\} ]\\
 \overset{E_1}{\leq} & \underset{y_t \sim \cald_{\caly|\bx_t} }{\bbe} [\min\{r_{t, i^\ast}^1 -   f(\bx_{t, i^\ast};  \btheta_{t-1})  + f(\bx_{t, \hi};  \btheta_{t-1})  -   r_{t, \hi}^1, 1\} ]\\
 = & \underset{y_t \sim \cald_{\caly|\bx_t} }{\bbe} [\min\{r_{t, i^\ast}^1  - f(\bx_{t, i^\ast}; \btheta_{t-1})  + f(\bx_{t, \hi}; \btheta_{t-1})  -   r_{t, \hi}^1, 1\} ]\\
 = & \underset{y_t \sim \cald_{\caly|\bx_t} }{\bbe} [\min\{r_{t, i^\ast}^1 - f(\bx_{t, i^\ast}; \btheta_{t-1}^\ast) + f(\bx_{t, i^\ast}; \btheta_{t-1}^\ast) - f(\bx_{t, i^\ast}; \btheta_{t-1})  + f(\bx_{t, \hi}; \btheta_{t-1})  -   r_{t, \hi}^1, 1\} ] \\
\leq &   \underset{y_t \sim \cald_{\caly|\bx_t} }{\bbe} [\min\{r_{t, i^\ast}^1 - f(\bx_{t, i^\ast}; \btheta_{t-1}^\ast), 1\} ] + 
 f(\bx_{t, i^\ast}; \btheta_{t-1}^\ast)  - f(\bx_{t, i^\ast}; \btheta_{t-1}) \\
& + \underset{y_t \sim \cald_{\caly|\bx_t} }{\bbe} [\min\{f(\bx_{t, \hi}; \btheta_{t-1})  -  r_{t, \hi}^1, 1\} ]   \\
\leq & \underset{y_t \sim \cald_{\caly|\bx_t} }{\bbe} [\min\{ |r_{t, i^\ast}^1 - f(\bx_{t, i^\ast}; \btheta_{t-1}^\ast) |, 1\} ] + 
| f(\bx_{t, i^\ast}; \btheta_{t-1}^\ast)  - f(\bx_{t, i^\ast}; \btheta_{t-1}) |\\
& +  \underset{y_t \sim \cald_{\caly|\bx_t} }{\bbe} [\min\{ |f(\bx_{t, \hi}; \btheta_{t-1})  -  r_{t, \hi}^1 |, 1\} ]   \\
\end{aligned}
\end{equation}
where $E_1$ is because of $  f(\bx_{t, i^\ast}; \btheta_{t-1})  \leq   f(\bx_{t, \hi}; \btheta_{t-1})$. 
For any $t \in [T] \wedge  (\mathbf{I}_t = 0)$, we have $R_t| (\mathbf{I}_t = 0)  =  \underset{\bx_t, y_t }{\bbe} [  L(\by_{t, \hi}, \by_{t, y_t}) -  L(\by_{t, i^\ast}, \by_{t, y_t})] = 0$ based on Lemma \ref{lemma:correctlabel}. 
Therefore, we have
\begin{equation}
\begin{aligned}
\mathbf{R}_T = & \sum_{t=1}^T R_t \\
\leq & \underbrace{\sum_{t=1}^T \underset{y_t \sim \cald_{\caly|\bx_t} }{\bbe} [\min\{ |r_{t, i^\ast}^1 - f(\bx_{t, i^\ast}; \btheta_{t-1}^\ast) |, 1\} ] }_{I_1} 
+ \underbrace{\sum_{t=1}^T | f(\bx_{t, i^\ast}; \btheta_{t-1}^\ast)  - f(\bx_{t, i^\ast}; \btheta_{t-1}) |}_{I_2} \\
& + \underbrace{\sum_{t=1}^T  \underset{y_t \sim \cald_{\caly|\bx_t} }{\bbe} [\min\{ |f(\bx_{t, \hi}; \btheta_{t-1})  -  r_{t, \hi}^1 |, 1\} ]}_{I_3} \\
\leq & 2 \left( 2 \sqrt{t \mu}  + \calo \left(\frac{3}{\sqrt{2} } L\nu\sqrt{T} \right) + \sqrt{2 T \log ( \calo(T)/\delta) } \right) + 2 T \xi_1  \\
\overset{E_2}{\leq} & \calo(1) + \calo \left( \frac{6L\nu + 4 \sqrt{\mu}}{\sqrt{2}}  \right)  \sqrt{T} + 2\sqrt{2 T \log ( \calo(T)/\delta) } 
\end{aligned}
\end{equation}
where $I_1$  is because of Lemma \ref{lemma:newthetaboundast}, $I_3$ is due to Lemma \ref{lemma:newthetabound}, $I_2$ is the application of Lemma \ref{lemma:difference} and $E_2$ is the result of choice of $m$.
\end{proof}

\begin{lemma} \label{lemma:newthetabound}
For any $\delta \in (0, 1), \nu > 0$, $\gamma \geq 1$, suppose $m$ satisfies the conditions in Lemma \ref{lemma:1}.
In round $t \in [T]$, given $(\bx_t, y_t) \sim \cald$, let
\[\hi = \arg \max_{i \in [k]} \left(   f_1(\bx_{t,\hi}; \btheta^1_{t-1}) + f_2(\phi(\bx_{t,\hi}); \btheta^2_{t-1})   \right).\]
Then, with probability at least $1-\delta$, we have
\begin{equation}
\begin{aligned}
\frac{1}{t} \sum_{\tau=1}^t \underset{y_\tau \sim \cald_{\caly|\bx_\tau} }{\bbe} \left[   \min \left \{ \left| f_1(\bx_{\tau, \hi}; \btheta^1_{\tau-1}) + f_2(\phi(\bx_{\tau, \hi}); \btheta^2_{\tau-1})   - r_{\tau, \hi}^1 \right|, 1  \right \}  \right] \\
\leq \sqrt{\frac{2 \mu}{t}} + \calo \left(\frac{3L\nu}{\sqrt{2t}} \right) + \sqrt{ \frac{2 \log ( \calo(1)/\delta) }{t}}.
\end{aligned}
\end{equation}
\end{lemma}

\begin{proof}
For any $\tau \in [t]$, define
\begin{equation}
\begin{aligned}
V_{\tau} =&\underset{y_\tau \sim \cald_{\caly|\bx_\tau} }{\bbe} \left[ \min \{  |f_1(\bx_{\tau, \hi}; \btheta^1_{\tau - 1}) + f_2(\bx_{\tau, \hi}; \btheta^2_{\tau - 1}) - r_{\tau, \hi}^1|, 1 \} \right]  \\
& - \min \{ |   f_1(\bx_{\tau, \hi}; \btheta^1_{\tau - 1}) + f_2(\bx_{\tau, \hi}; \btheta^2_{\tau - 1}) - r_{\tau, \hi}^1 |, 1 \}
\end{aligned}
\end{equation}
Then, we have
\begin{equation}
\begin{aligned}
\bbe[V_{\tau}| F_{\tau - 1}]  =&\underset{y_\tau \sim \cald_{\caly|\bx_\tau} }{\bbe} \left[ \min \{ |f_1(\bx_{\tau, \hi}; \btheta^1_{\tau - 1}) + f_2(\bx_{\tau, \hi}; \btheta^2_{\tau - 1}) - r_{\tau, \hi}^1|, 1 \} \right] \\
& - \underset{y_\tau \sim \cald_{\caly|\bx_\tau} }{\bbe} \left[ \min \{  |   f_1(\bx_{\tau, \hi}; \btheta^1_{\tau - 1}) + f_2(\bx_{\tau, \hi}; \btheta^2_{\tau - 1}) - r_{\tau, \hi}^1 |, 1 \} \right] \\
= & 0
\end{aligned}
\end{equation}
where $F_{\tau - 1}$ denotes the $\sigma$-algebra generated by the history $\mathcal{H}_{\tau -1}$. Therefore, $\{V_{\tau}\}_{\tau =1}^t$ are the martingale difference sequence.

Applying the Hoeffding-Azuma inequality, with probability at least $1-\delta$, we have
\begin{equation}
    \bbp \left[  \frac{1}{t}  \sum_{\tau=1}^t  V_{\tau}  -   \underbrace{ \frac{1}{t} \sum_{\tau=1}^t \underset{y_\tau \sim \cald_{\caly|\bx_\tau} }{\bbe}[ V_{\tau} | \mathbf{F}_{\tau} ] }_{I_1}   >  \sqrt{ \frac{2 \log (1/\delta)}{t}}  \right] \leq \delta   \\
\end{equation}    
As $I_1$ is equal to $0$, we have
\begin{equation}
\begin{aligned}
      &\frac{1}{t} \sum_{\tau=1}^t \underset{y_\tau \sim \cald_{\caly|\bx_\tau} }{\bbe} \left[ \min \{ \left |f_1(\bx_{\tau, \hi}; \btheta^1_{\tau - 1}) + f_2(\bx_{\tau, \hi}; \btheta^2_{\tau - 1}) - r_{\tau, \hi}^1) \right|, 1\}   \right]  \\
 \leq  &  \underbrace{ \frac{1}{t}\sum_{\tau=1}^t \min \{  \left|f_2 \left( \bx_{\tau, \hi} ; \btheta_{\tau-1}^2 \right) - \left(r_{\tau, \hi} - f_1(\bx_{\tau, \hi}; \btheta_{\tau-1}^1) \right)  \right|, 1 \} }_{I_3}  +   \sqrt{ \frac{2 \log (1/\delta) }{t}}   ~.
    \end{aligned}  
\label{eq:pppuper}
\end{equation}

For $I_3$, based on Lemma \ref{lemma:caogeneli}, for any $\widetilde{\btheta}^2$ satisfying $\| \widetilde{\btheta}^2  - \btheta^2_0 \|_2 \leq \calo(\frac{\nu}{\sqrt{m}})$, with probability at least $1 -\delta$, we have
\begin{equation}
\begin{aligned}
I_3 &\leq \frac{1}{t}  \sum^t_{\tau = 1 } \min \{ |f_1(\bx_{\tau, \hi}; \btheta_{\tau-1}^1) + f_2 \left( \bx_{\tau, \hi} ; \widetilde{\btheta}^2 \right) - r_{\tau, \hi}^1|, 1 \} + \calo \left (\frac{3L \nu}{\sqrt{2t}} \right) \\
& \leq \frac{1}{t}\sqrt{t}\sqrt{
\underbrace{
\sum_{\tau =1}^t 
\left(      
f_1(\bx_{\tau, \hi}; \btheta_{\tau-1}^1)
+ f_2 \left( \bx_{\tau, \hi} ; \widetilde{\btheta}^2 \right) - r_{\tau, \hi}^1  
\right)^2}_{I_4}} + \calo \left (\frac{3L \nu}{\sqrt{2t}} \right)  +   \sqrt{ \frac{2 \log (1/\delta)}{t}} 
 \\
& \leq \sqrt{\frac{2  \mu } {t}} + \calo \left (\frac{3L \nu}{\sqrt{2t}} \right) +    \sqrt{ \frac{2 \log (1/\delta) }{t}}.
\end{aligned}
\end{equation}
where $I_4$ is by the definition of $\mu$.

Combining above inequalities together, with probability at least $1 - \delta$, we have 
\begin{equation}
\begin{aligned}
\frac{1}{t} \sum_{\tau=1}^t \underset{y_\tau \sim \cald_{\caly|\bx_\tau} }{\bbe} \left[   \min \left \{ \left| f_1(\bx_{\tau, \hi}; \btheta^1_{\tau-1}) + f_2(\phi(\bx_{\tau, \hi}); \btheta^2_{\tau-1})   - r_{\tau, \hi}^1 \right|, 1  \right \}  \right] \\
\leq \sqrt{ \frac{2\mu }{t}} + \calo \left(\frac{3L \nu}{\sqrt{2t}} \right) + 2 \sqrt{ \frac{2 \log ( \calo(1)/\delta) }{t}}, 
\end{aligned}
\end{equation}
where we applied union bound over $\delta$ to make above events occur concurrently.
\end{proof}

\begin{lemma} \label{lemma:newthetaboundast}
For any $\delta \in (0, 1), \nu > 0$, $\gamma \geq 1$, suppose $m$ satisfies the conditions in Lemma \ref{lemma:1}.
In round $t \in [T]$, given $(\bx_t, y_t) \sim \cald$, let  $i^\ast = \arg \max_{ i\in[k]} h(\bx_{t,i})$. Then, with probability at least $1 - \delta$,  there exists $\btheta^{1,\ast}_{t-1}, \btheta^{2, \ast}_{t-1}$,  such that
\begin{equation}
\begin{aligned}
\frac{1}{t} \sum_{\tau=1}^t  \underset{y_\tau \sim \cald_{\caly|\bx_\tau} }{\bbe} \left[   \min \left \{ \left| f_1(\bx_{\tau, i^\ast}; \btheta^1_{\tau-1}) + f_2(\phi(\bx_{\tau, i^\ast}); \btheta^2_{\tau-1})   - r_{\tau, i^\ast}^1 \right|, 1  \right \}  \right] \\
\leq \sqrt{ \frac{2\mu}{t}}+ \calo \left(\frac{3L \nu}{\sqrt{2t}} \right) +  2 \sqrt{ \frac{2 \log ( \calo(1)/\delta) }{t}}, 
\end{aligned}
\end{equation}
where $\calh_{t-1} = \{\bx_{\tau, i^\ast}, r_{\tau, i^\ast}^1 \}_{\tau = 1}^{t-1}$ is historical data.
\end{lemma}

\begin{proof}
For any $\tau \in [t]$, define
\begin{equation}
\begin{aligned}
V_{\tau} =& \underset{y_\tau \sim \cald_{\caly|\bx_\tau} }{\bbe} \left[ \min \{  |f_1(\bx_{\tau, i^\ast}; \btheta^1_{\tau - 1}) + f_2(\bx_{\tau, i^\ast}; \btheta^2_{\tau - 1}) - r_{\tau, i^\ast}^1)|, 1 \} \right]  \\
& - \min \{ |   f_1(\bx_{\tau, i^\ast}; \btheta^1_{\tau - 1}) + f_2(\bx_{\tau, i^\ast}; \btheta^2_{\tau - 1}) - r_{\tau, i^\ast}^1 |, 1 \}
\end{aligned}
\end{equation}
Then, we have
\begin{equation}
\begin{aligned}
\bbe[V_{\tau}| F_{\tau - 1}]  =& \underset{y_\tau \sim \cald_{\caly|\bx_\tau} }{\bbe} \left[ \min \{ |f_1(\bx_{\tau, i^\ast}; \btheta^1_{\tau - 1}) + f_2(\bx_{\tau, i^\ast}; \btheta^2_{\tau - 1}) - r_{\tau, i^\ast}^1)|, 1 \} \right] \\
& -  \underset{y_\tau \sim \cald_{\caly|\bx_\tau} }{\bbe} \left[ \min \{  |   f_1(\bx_{\tau, i^\ast}; \btheta^1_{\tau - 1}) + f_2(\bx_{\tau, i^\ast}; \btheta^2_{\tau - 1}) - r_{\tau, i^\ast}^1 |, 1 \} \right] \\
= & 0
\end{aligned}
\end{equation}
where $F_{\tau - 1}$ denotes the $\sigma$-algebra generated by the history $\mathcal{H}_{\tau -1}$. Therefore, $\{V_{\tau}\}_{\tau =1}^t$ are the martingale difference sequence.

Applying the Hoeffding-Azuma inequality, with probability at least $1-\delta$, we have
\begin{equation}
    \bbp \left[  \frac{1}{t}  \sum_{\tau=1}^t  V_{\tau}  -   \underbrace{ \frac{1}{t} \sum_{\tau=1}^t  \underset{y_\tau \sim \cald_{\caly|\bx_\tau} }{\bbe}[ V_{\tau} | \mathbf{F}_{\tau} ] }_{I_1}   >  \sqrt{ \frac{2 \log (1/\delta)}{t}}  \right] \leq \delta   \\
\end{equation}    
As $I_1$ is equal to $0$, we have
\begin{equation}
\begin{aligned}
      &\frac{1}{t} \sum_{\tau=1}^t  \underset{y_\tau \sim \cald_{\caly|\bx_\tau} }{\bbe} \left[ \min \{ \left |f_1(\bx_{\tau, i^\ast}; \btheta^1_{\tau - 1}) + f_2(\bx_{\tau, i^\ast}; \btheta^2_{\tau - 1}) - r_{\tau, i^\ast}^1) \right|, 1\}   \right]  \\
 \leq  &  \underbrace{ \frac{1}{t}\sum_{\tau=1}^t \min \{  \left|f_2 \left( \bx_{\tau, i^\ast} ; \btheta_{\tau-1}^2 \right) - \left(r_{\tau, i^\ast} - f_1(\bx_{\tau, i^\ast}; \btheta_{\tau-1}^1) \right)  \right|, 1 \} }_{I_3}  +   \sqrt{ \frac{2 \log (1/\delta) }{t}}   ~.
    \end{aligned}  
\label{eq:pppuper}
\end{equation}

For $I_3$, based on Lemma\ref{lemma:caogeneli}, for any $\widetilde{\btheta}^2$ satisfying $\| \widetilde{\btheta}^2  - \btheta^2_0 \|_2 \leq \calo(\frac{\nu}{\sqrt{m}})$, with probability at least $1 -3\delta$, we have
\begin{equation}
\begin{aligned}
I_3 &\leq \frac{1}{t}  \sum^t_{\tau = 1 } \min \{ |f_1(\bx_{\tau, i^\ast}; \btheta_{\tau-1}^1) + f_2 \left( \bx_{\tau, i^\ast} ; \widetilde{\btheta}^2 \right) - r_{\tau, i^\ast}^1|, 1 \} + \calo \left (\frac{3L \nu}{\sqrt{2t}} \right) \\
& \leq \frac{1}{t}\sqrt{t}\sqrt{
\underbrace{
\sum_{\tau =1}^t 
\left(      
f_1(\bx_{\tau, i^\ast}; \btheta_{\tau-1}^1)
+ f_2 \left( \bx_{\tau, i^\ast} ; \widetilde{\btheta}^2 \right) - r_{\tau, i^\ast}^1  
\right)^2}_{I_4}} + \calo \left (\frac{3L \nu}{\sqrt{2t}} \right) +   \sqrt{ \frac{2 \log (1/\delta) }{t}}
 \\
& \leq \sqrt{\frac{2\mu} {t}} + \calo \left (\frac{3L \nu}{\sqrt{2t}} \right) +   \sqrt{ \frac{2 \log (1/\delta) }{t}}.
\end{aligned}
\end{equation}
where $I_4$ is by the assumption of $\mu$.

Combining above inequalities together, with probability at least $1 - \delta$, we have 
\begin{equation}
\begin{aligned}
\frac{1}{t} \sum_{\tau=1}^t  \underset{y_\tau \sim \cald_{\caly|\bx_\tau} }{\bbe} \left[   \min \left \{ \left| f_1(\bx_{\tau, i^\ast}; \btheta^1_{\tau-1}) + f_2(\phi(\bx_{\tau, i^\ast}); \btheta^2_{\tau-1})   - r_{\tau, i^\ast}^1 \right|, 1  \right \}  \right] \\
\leq \sqrt{ \frac{2\mu }{t}} + \calo \left(\frac{3L \nu}{\sqrt{2t}} \right) + 2\sqrt{ \frac{2 \log ( \calo(1)/\delta) }{t}}, 
\end{aligned}
\end{equation}
where we applied union bound over $\delta$ to make above events occur concurrently.
\end{proof}

\section{Bounds for Effective Dimension $\widetilde{d}$} \label{sub:effective}

\newcommand{\bsigma}{\boldsymbol{\Sigma}}

 Let $\{\bx_{t, \hi}\}_{t=1}^T$ be the selected contexts in $T$ rounds, then we have the following definition of NTK.

\begin{definition} [ NTK \cite{ntk2018neural, arora2019exact}] Let $\mathcal{N}$ denote the normal distribution.
Define 
\[
\begin{aligned}
&\mathbf{H}_{i,j}^0 = \bsigma^{0}_{i,j} =  \langle \bx_i, \bx_j\rangle,   \ \ 
\mathbf{N}^{l}_{i,j} =
\begin{pmatrix}
\bsigma^{l}_{i,i} & \bsigma^{l}_{i,j} \\
\bsigma^{l}_{j,i} &  \bsigma^{l}_{j,j} 
\end{pmatrix} \\
&   \bsigma^{l}_{i,j} = 2 \mathbb{E}_{a, b \sim  \mathcal{N}(\mathbf{0}, \mathbf{N}_{i,j}^{l-1})}[ \sigma(a) \sigma(b)] \\
& \mathbf{H}_{i,j}^l = 2 \mathbf{H}_{i,j}^{l-1} \mathbb{E}_{a, b \sim \mathcal{N}(\mathbf{0}, \mathbf{N}_{i,j}^{l-1})}[ \sigma'(a) \sigma'(b)]  + \bsigma^{l}_{i,j}.
\end{aligned}
\]
Then, over the contexts $\{\bx_{t, \hi}\}_{t=1}^T$, the Neural Tangent Kernel (NTK) is defined as $ \mathbf{H} =  (\mathbf{H}^L + \bsigma^{L})/2$.
\end{definition}

Then, we define the following gram matrix $\mathbf{G}$. Let $g(x; \btheta_0) = \tri_{\btheta}f(x; \btheta_0) \in \bbr^p $ and  $G = [g(\bx_{1, \hi}; \btheta_0)/\sqrt{m}, \dots, g(\bx_{T, \hi}; \btheta_0)/\sqrt{m}] \in \bbr^{p \times T}$ where $p = m + mkd+m^2(L-1)$. Therefore, we have $ \mathbf{G} = G^\top G$. Based on Theorem 3.1 in \cite{arora2019exact}, when $m \geq \calo(T^4 k^6 \log (2Tk/\delta) / \lambda_0^4)$ where $\lambda_0$ is the smallest eigenvalue of $\mathbf{H}$, with probability at least $1-\delta$, we have
\begin{equation}
\|\mathbf{G}  - \mathbf{H}\| \leq \frac{\lambda_0}{2}.
\end{equation}
Then, we have the following bound:
\begin{equation}
\begin{aligned}
\log \det( \mathbf{I} + \mathbf{H}) & = \log \det \left(\mathbf{I} + \mathbf{G } + (\mathbf{H} - \mathbf{G})    \right) \\
& \leq   \log \det (\mathbf{I} + \mathbf{G}) + \langle (\mathbf{I} + \mathbf{G})^{-1}, (\mathbf{H} -\mathbf{G})   \rangle\\
& \leq  \log \det (\mathbf{I} + \mathbf{G})  + \|  (\mathbf{I} + \mathbf{G})^{-1}  \|_F \| \mathbf{H}- \mathbf{G}\|_F\\
& \leq \log \det (\mathbf{I} + \mathbf{G})  + \sqrt{T} \| \mathbf{H}- \mathbf{G}\|_F\\
& \leq \log \det (\mathbf{I} + \mathbf{G})  + 1\\
\end{aligned}
\end{equation}
where the first inequality is because of the  concavity of $\log \det (\cdot)$ and the third inequality is by  Lemma B.1 in \cite{zhou2020neural} with the choice of $m$.
Then, the effective dimension $\widetilde{d}$ can be bounded by:
\begin{equation}
\begin{aligned}
\widetilde{d}  = & \frac{\log \det (\mathbf{I} + \mathbf{H})}{\log (1 + T)} \\
&\leq \frac{ \log \det (\mathbf{I} + \mathbf{G})  + 1}{  \log (1 + T)  } \\
& \overset{E_1}{ =} \frac{ \log \det (\mathbf{I} + G G^\top)  + 1}{  \log (1 + T)  } \\
& \overset{E_2}{\leq} p \cdot \frac{ \log \| \mathbf{I} +  G G^\top \|_2 } { \log (1 + T) } + \frac{1}{ \log (1 + T) } \\
& \overset{E_3}{\leq} p  + \frac{1}{ \log (1 + T) }
\end{aligned}
\end{equation}
where $E_1$ is because of $\det (\mathbf{I} + G^\top G) =  \det (\mathbf{I} + GG^\top)$ and $E_2$ is due to $\det (  G G^\top) = \|G G^\top \|^p_2$ ($G G^\top \in \bbr^{p \times p}$) and $E_3$ is according to
\[
\| \mathbf{I} +  G G^\top \|_2 \leq 1 + \|  G G^\top\|_2 \leq 1 +\sum_{t=1}^T \|g(\bx_{t, \hi}; \btheta_0)  g(\bx_{t, \hi}; \btheta_0)^\top/m\|_2 \leq  1 +T,
\]
where the last inequality is as the result of $\| g(\bx_{t, \hi}; \btheta_0)/\sqrt{m}\|_2 \leq 1$ (Lemma B.3 in \cite{cao2019generalization}).
Therefore, we have
\begin{equation}
\widetilde{d}  \leq   p  + \frac{1}{ \log (1 + T) }  \ \  \text{and} \ \  p= m + mkd+m^2(L-1).
\end{equation}
\clearpage

\section{Further Details in Experiments} \label{append:exp}

In this section, we report the specific configurations in the experiments, the sensitivity study of the core hyperparameter $\gamma$ for \sysn, and the ablation study for label budget. Table \ref{stat} exhibits the details of using datasets.

\begin{table}[h]
\centering
\begin{tabular}{c|ccc}
\hline
Dataset  & Features & Samples & Classes \\ \hline
Phishing     & 68  & 11,055 &    2    \\
IJCNN & 22  & 12,000   &    2   \\
Letter   & 784 & 12,000  &    26     \\
Fashion   & 784  & 12,000  &   10  \\ 
MNIST    & 784 & 12,000 & 10     \\
CIFAR-10   & 3,072 & 12,000 & 10      \\ \hline
\end{tabular}
\caption{Statistics of the datasets used in our experiments. We conduct experiments on binary classification tasks. For Letter, the binary task is to separate `A-M' versus `N-Z'. For Fashion, the binary task is to separate `T-shirt' versus `Trouser' images. For MNIST, the binary task is to separate odd and even digits. For CIFAR-10, the binary task is to separate `horse' and `ship' images.}
\label{stat}
\end{table}

\para{Implementation Details}. 
We use PyTorch as our backend, and all experiments were conducted on a server with NVIDIA Tesla V100 SXM2 GPU. 
The classification model in all methods is the same 2-layer fully-connected network with 100-width for the fair comparison. We use Adam optimizer to train the classification model with the fixed learning rate is $0.001$, and the batch size is $64$, since these are model-agnostic hyperparameters.
As NeuAL-NTK-F and NeuAL-NTK-D only work on the binary classification problem, we transformed the $k$-class classification problem into the binary classification problem when $k > 2 $. In detail, given $k$ class, we regard the  $k/2$ classes as one class and remaining classes as another class.
For Random algorithm, the query probability $p$ is set as $0.1$.
To find the best performance of each method, we conduct the grid search over all hyperparameters. 
In Margin algorithm, the query threshold is searched over $\{0.3, 0.5, 0.7, 0.9, 0.95\}$ for all datasets.
For NeuAL-NTK-F and NeuAL-NTK-D, there is also an exploration parameter $\gamma$ to determine the query aggressiveness and we conduct the grid search over $\{0.1, 0.3, 0.5, 0.8, 1.0\}$ for it.
For ALPS, following the method in \cite{desalvo2021online}, we form the hypothesis class by generating 20 hypotheses on the $3 \%$ of total data samples (as same as the query budget) with different random seeds, and we conduct the grid search  $\{0.1,0.25, 0.5,0.75,0.9\}$ over the two slack terms in ALPS. We have tried to generate more hypotheses in the experiments, but the performance of ALPS does not improve accordingly.
For \sysn, the only hyperparameter  $\gamma$ is searched over $\{1, 2, 5, 6, 7, 10\}$ for all datasets ($c_1$ and $c_2$ in $\bbeta_t$ is set as $1$).
The confidence level $\delta$ is set as $0.1$ for all the needed methods.
In the end, we report the average results of 5 runs for all methods.

\begin{figure*}[htbp]
\centering
\subfigure[Phishing]{
\includegraphics[width=0.32\textwidth]{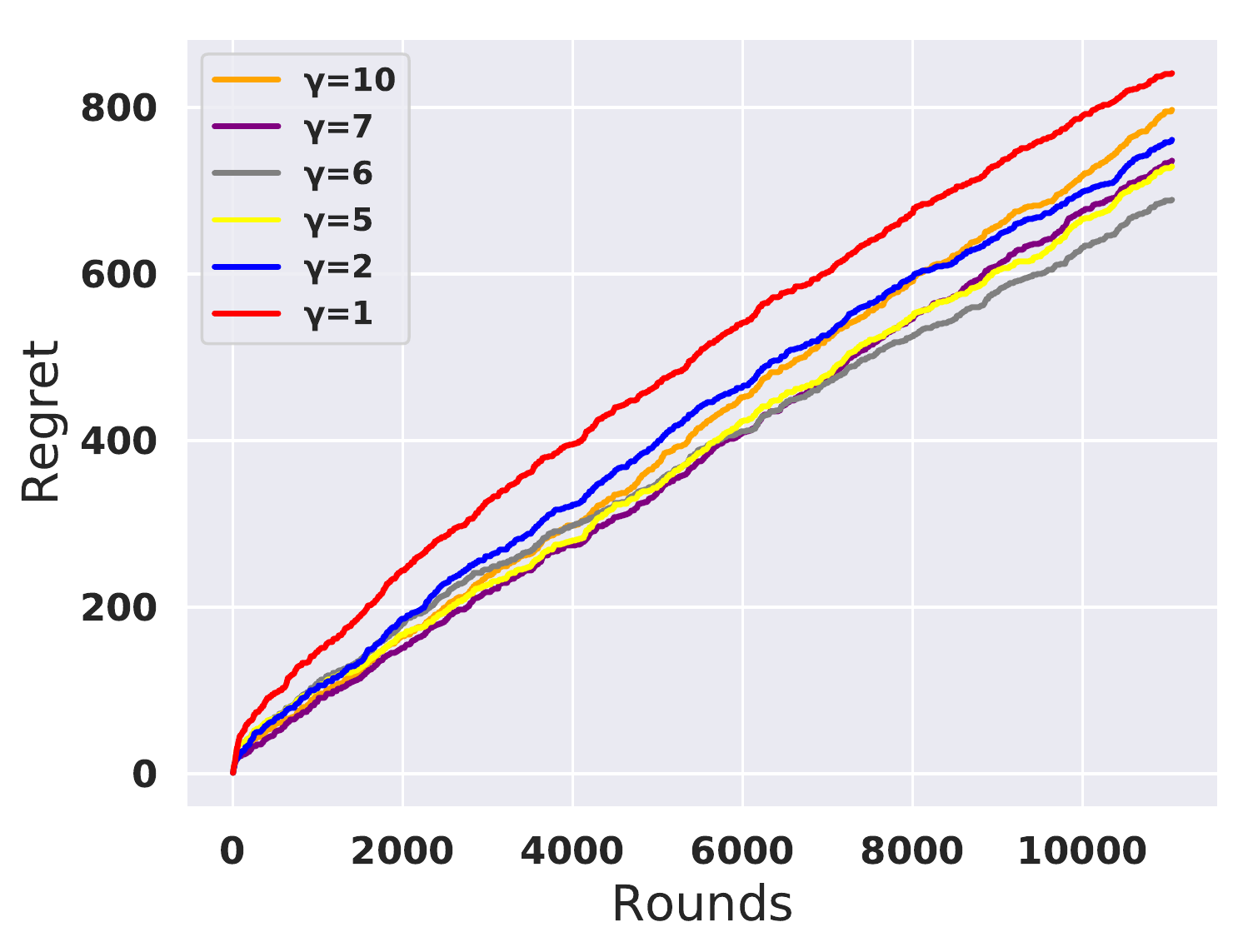}}
\subfigure[MNIST]{
\includegraphics[width=0.32\textwidth]{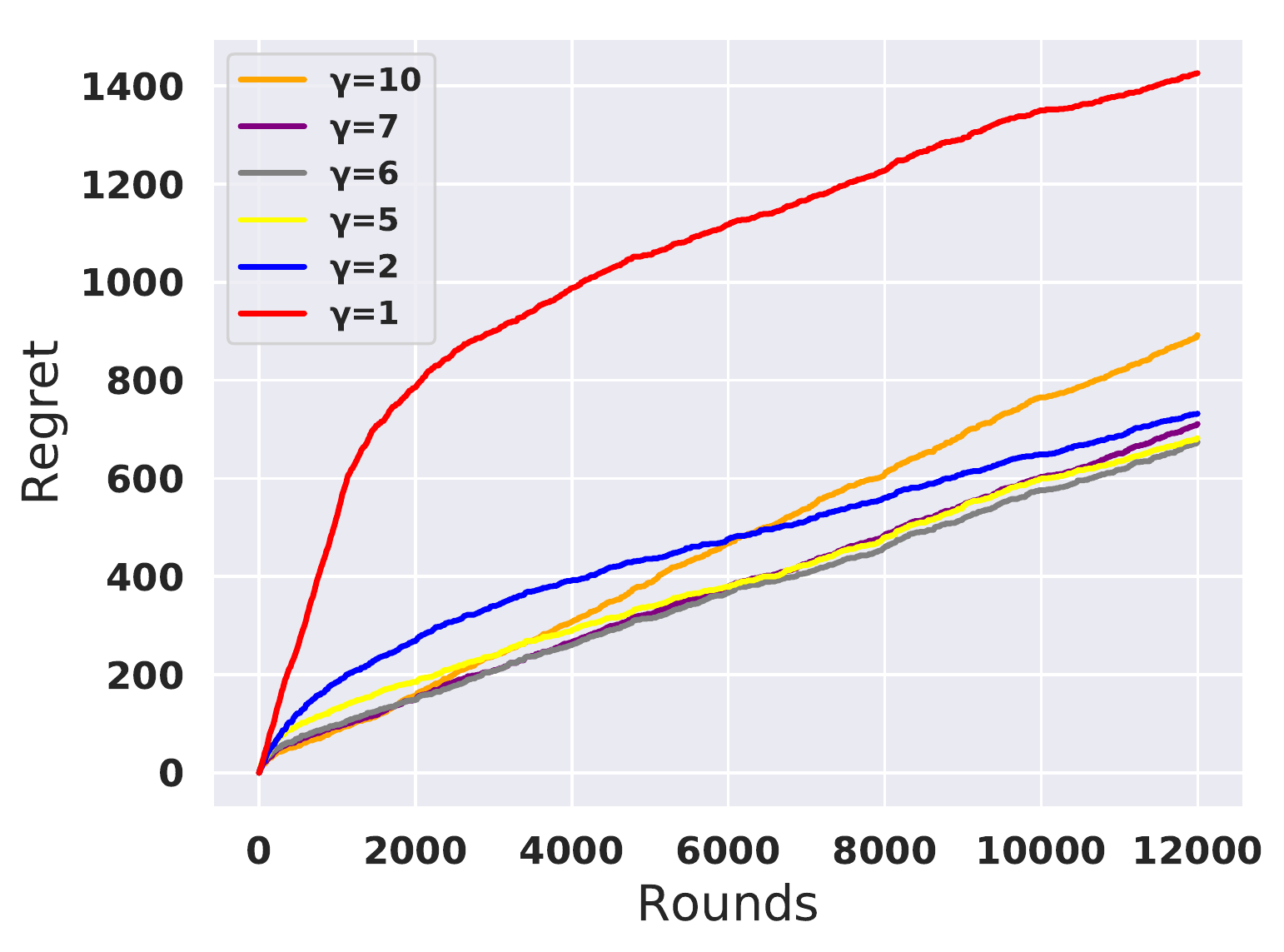}}
\subfigure[CIFAR-10]{
\includegraphics[width=0.32\textwidth]{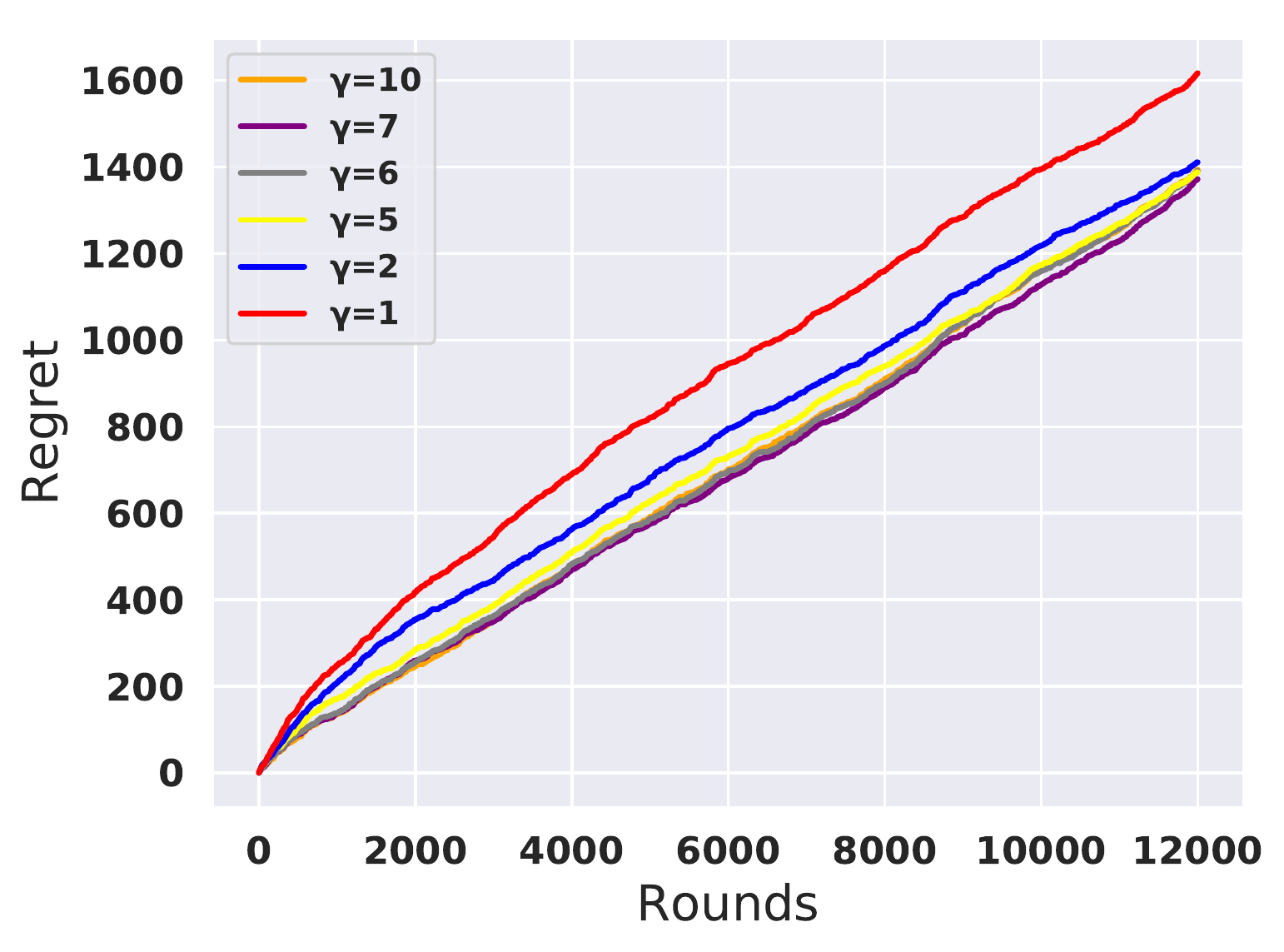}}
\caption{Parameter sensitivity on three datasets.}
\label{hyper}
\end{figure*}
\para{Sensitivity study for $\gamma$}.
As $\gamma$ is closely related to the query threshold of \sysn, we test the sensitivity of \sysn with regard to $\gamma$. Based on our analysis, it is required that $\gamma \geq 1$. When $\gamma$ is the smallest number (e.g., $\gamma =1$), \sysn queries the labels only if the difference between the top two classes is very small (i.e., the confidence level is very low). In this manner, \sysn will save more query budget but take more risks on many instances, incurring more regret. This explains why the red line ($\gamma = 1$) is above all other lines. In contrast, if $\gamma$ is a large number, \sysn will be more aggressive in making queries and thus obtain satisfactory performance. However, if $\gamma$ is too large, \sysn tends to query on these instances even when our model is very confident to the predictions, wasting the query budget that could have been used on these uncertain instances. Therefore, we expect that $\gamma$ is neither too small nor too large. The experiments verify our assumption, when $\gamma$ is $6$ or $7$, \sysn almost achieves the best performance throughout all datasets and configurations.

\para{Ablation study for label budget}. 
To examine the final performance of each algorithm, we conduct new experiments with different percentages of label budget: 3\%, 10\%, 20\%, 50\%. 
After $T$ rounds, we evaluate the latest model on the test (unseen) data to calculate the accuracy, which evaluates the population accuracy. 
For all the datasets, $T$ is set as $10000$, except that $T = 2000$ for Phishing because Phishing has fewer data instances.
Table \ref{acc3} - \ref{acc50} reports the results. 
To sum up, I-NeurAL still achieves the best accuracy with different label budget. With a small amount of label budget (3\%, 10\%), I-NeurAL can make smart decisions to query labels on these instances with big uncertainty and leverage the full feedback to exploit the past knowledge, which enable I-NeurAL to outperform all the baselines.  With the larger label budget (20\%, 50 \%),   all methods have enough labels to train. Thus, the advantages of I-NeurAL is less significant and the gap between I-NeurAL and baselines is decreasing. Nevertheless, I-NeurAl still has the best performance benefiting from smart query choices.

\begin{table}[h]
\centering
\begin{tabular}{c|cccccc}
\hline
            & Phishing     & IJCNN        & Letter          & Fashion       & MNIST        & CIFAR-10      \\ \hline
Random        & 91.75\%         & \underline{93.80\%}          & 71.60\%          & 95.70\%          & 87.90\%         & 86.40\%          \\
Margin      & 93.46\%          & 92.95\%          &      73.55\%     & \underline{98.15\%}         & \underline{90.25\%}       & \underline{88.25\%}          \\
NeuAL-NTK-F & 54.69\%         & 75.15\%         & 48.05\%         & 51.30\%        & 51.10\%         & 71.00\%          \\
NeuAL-NTK-D & \underline{92.89\%}          & 93.65\%          & \underline{73.80\%}          & 97.70\%          & 90.15\%         & 84.05\%         \\
ALPS        & 91.47\%          & 93.25\%          & 71.45\%         & 95.70\%          & 86.95\%         & 85.40\%          \\ \hline
I-NeurAL    & $\mathbf{94.22\%}$ & $\mathbf{95.75\%}$ & $\mathbf{77.45\%}$ & $\mathbf{99.15\%}$ & $\mathbf{94.45\%}$ & $\mathbf{89.00\%}$ \\ \hline
\end{tabular}
\caption{Test Accuracy with \textbf{3\%} budget.}
\label{acc3}
\end{table}

\begin{table}[h]
\centering
\begin{tabular}{c|cccccc}
\hline
            & Phishing     & IJCNN        & Letter          & Fashion       & MNIST        & CIFAR-10      \\ \hline
Random        & 93.93\%         & 96.70\%          & 79.70\%          & 97.30\%          & 90.90\%         & 89.00\%          \\
Margin      & \underline{94.98\%}          & \underline{97.10\%}          &      \underline{81.50}\%     & 98.60\%         & 94.40\%       & \underline{89.45}\%          \\
NeuAL-NTK-F & 54.69\%         & 87.65\%         & 48.05\%         & 51.30\%        & 51.10\%         & 70.95\%          \\
NeuAL-NTK-D & 92.99\%          & 96.90\%          & 80.55\%          & \underline{98.70}\%          & \underline{94.85\%}         & 89.05\%         \\
ALPS        & 92.89\%          & 96.20\%          & 78.05\%         & 97.50\%          & 93.00\%         & 89.35\%          \\ \hline
I-NeurAL    & $\mathbf{95.64\%}$ & $\mathbf{97.90\%}$ & $\mathbf{83.95\%}$ & $\mathbf{99.30\%}$ & $\mathbf{97.20\%}$ & $\mathbf{91.75\%}$ \\ \hline
\end{tabular}
\caption{Test Accuracy with \textbf{10\%} label budget. }
\label{acc10}
\end{table}

\begin{table}[h]
\centering
\begin{tabular}{c|cccccc}
\hline
            & Phishing     & IJCNN        & Letter          & Fashion       & MNIST        & CIFAR-10      \\ \hline
Random        & 93.93\%         & 96.70\%          & 81.90\%          & 98.15\%          & 93.00\%         & 89.50\%          \\
Margin      & \underline{95.17\%}         & \underline{98.15\%}          &      82.45\%     & 98.90\%         & 95.05\%       & 89.75\%          \\
NeuAL-NTK-F & 54.69\%         & 89.95\%         & 48.05\%         & 51.30\%        & 51.10\%         & 72.15\%          \\
NeuAL-NTK-D & 94.98\%          & 97.75\%          & 82.35\%          & \underline{99.30\%}          & \underline{96.15}\%         & \underline{90.90\%}         \\
ALPS        & 94.41\%          & 97.05\%          & \underline{83.20\%}         & 98.30\%          & 94.45\%         & 88.95\%          \\ \hline
I-NeurAL    & $\mathbf{95.64\%}$ & $\mathbf{98.35\%}$ & $\mathbf{84.65\%}$ & $\mathbf{99.30\%}$ & $\mathbf{97.95\%}$ & $\mathbf{91.80\%}$ \\ \hline
\end{tabular}
\caption{Test Accuracy with \textbf{20\%} label budget. }
\label{acc20}
\end{table}

\begin{table}[h]
\centering
\begin{tabular}{c|cccccc}
\hline
            & Phishing     & IJCNN        & Letter          & Fashion       & MNIST        & CIFAR-10      \\ \hline
Random        & 94.98\%         & 97.75\%          & 86.05\%          & 98.95\%          & 96.40\%         & 90.75\%          \\
Margin      & 95.73\%         & \underline{98.40\%}          &      86.35\%     & 99.05\%         & 96.20\%       & \underline{91.25\%}          \\
NeuAL-NTK-F & 54.69\%         & 90.85\%         & 48.05\%         & 51.30\%        & 51.10\%         & 72.35\%          \\
NeuAL-NTK-D & \underline{95.83\%}          & 98.00\%          & 83.35\%          & \underline{99.30\%}          & \underline{97.15\%}         & 90.55\%         \\
ALPS        & 94.50\%          & 97.80\%          & \underline{87.10\%}         & 99.05\%          & 96.70\%         & 90.85\%          \\ \hline
I-NeurAL    & $\mathbf{96.02\%}$ & $\mathbf{98.75\%}$ & $86.05\% $ & $\mathbf{99.35\%}$ & $\mathbf{97.80\%}$ & $\mathbf{92.30\%}$ \\ \hline
\end{tabular}
\caption{Test Accuracy with \textbf{50\%} label budget. }
\label{acc50}
\end{table}

\clearpage

\end{document}